\documentclass[11pt]{article}

\makeatletter
\@ifundefined{ifquadrevision}{\newif\ifquadrevision\quadrevisionfalse}{}
\@ifundefined{ifquadclean}{\newif\ifquadclean\quadcleanfalse}{}
\makeatother

\usepackage[final]{acl}

\usepackage{times}
\usepackage{latexsym}

\usepackage[T1]{fontenc}

\usepackage[utf8]{inputenc}

\usepackage{microtype}

\usepackage{inconsolata}

\usepackage{graphicx}

\usepackage{microtype}
\usepackage{hyperref}
\usepackage{url}
\usepackage{booktabs}

\usepackage{times}
\usepackage{latexsym}
\usepackage{ragged2e}
\usepackage[T1]{fontenc}

\usepackage[utf8]{inputenc}

\usepackage{microtype}

\usepackage{inconsolata}

\usepackage{graphicx}

\usepackage{algorithm}
\usepackage{graphicx}
\usepackage{amsmath}
\usepackage{algpseudocode}
\usepackage{xcolor}
\usepackage{amssymb}
\usepackage{multirow}
\usepackage{booktabs}
\usepackage{array} 
\usepackage{listings}
\usepackage{wrapfig}
\usepackage{multicol} 
\usepackage{soul}
\usepackage[normalem]{ulem}

\newcommand{\best}[1]{\uline{\textbf{#1}}}   
\newcommand{\secondbest}[1]{\textbf{#1}}     

\usepackage{placeins} 
\usepackage{etoolbox} 

\usepackage{dblfloatfix} 

\usepackage{wrapfig}
\usepackage{enumitem}

\setlist[itemize]{itemsep=0pt, topsep=2pt, parsep=0pt, partopsep=0pt}
\setlist[enumerate]{itemsep=0pt, topsep=2pt, parsep=0pt, partopsep=0pt}

\makeatletter
\newif\ifAppendixFloatsPending
\newcommand{\Appendix@checkpending}{%
  \begingroup
  \let\@elt\relax
  \xdef\Appendix@pending{\@currlist\@deferlist\@dbldeferlist}%
  \endgroup
  \ifx\Appendix@pending\@empty\AppendixFloatsPendingfalse\else\AppendixFloatsPendingtrue\fi}
\newcommand{\AppendixNewPage}{\par\newpage\if@twocolumn\if@firstcolumn\else\mbox{}\newpage\fi\fi}
\newcommand{\AppendixBeforeBox}{\par\Appendix@checkpending
  \ifAppendixFloatsPending\AppendixNewPage\fi}
\newcommand{\AppendixFlushFloats}{\par\Appendix@checkpending
  \ifAppendixFloatsPending\clearpage
    \global\@colht\textheight\global\@colroom\textheight\global\vsize\textheight
  \fi}
\makeatother

\makeatletter
\@ifundefined{fullwidth}{%
  \newenvironment{fullwidth}{%
    \par
    \noindent
    \ignorespaces
  }{%
    \par
    \ignorespaces
  }%
}{%
  \renewenvironment{fullwidth}{%
    \par
    \noindent
    \ignorespaces
  }{%
    \par
    \ignorespaces
  }%
}
\makeatother

\usepackage[most]{tcolorbox}
\tcbuselibrary{breakable,skins}
\tcbuselibrary{listings}

\tcbset{
  AppendixWide/.style={
    width=\linewidth,
    boxsep=1mm,
    left=1mm,right=1mm,top=1mm,bottom=1mm
  },
  AppendixColoredBoxBase/.style={
    AppendixWide,
    enhanced,
    breakable,
    sharp corners,
    boxrule=0.8pt,
    before skip=10pt,
    after skip=10pt,
    halign title=left,
    before upper=\justifying\setlength{\parindent}{0pt},
    coltitle=white,
    fonttitle=\bfseries,
    lefttitle=1mm,
    righttitle=1mm,
    toptitle=0.6mm,
    bottomtitle=0.6mm
  },
  AppendixColoredListingBase/.style={
    AppendixWide,
    enhanced,
    breakable,
    sharp corners,
    boxrule=0.8pt,
    before skip=10pt,
    after skip=10pt,
    halign title=left,
    coltitle=white,
    fonttitle=\bfseries,
    lefttitle=1mm,
    righttitle=1mm,
    toptitle=0.6mm,
    bottomtitle=0.6mm,
    listing engine=listings,
    listing options={
      basicstyle=\rmfamily\normalsize, 
      breaklines=true,
      breakatwhitespace=false,         
      breakautoindent=true,
      breakindent=0pt,
      prebreak=\mbox{\textcolor{gray}{$\hookleftarrow$}}, 
      postbreak=\mbox{\textcolor{gray}{$\hookrightarrow$}\space},
      columns=fullflexible,
      keepspaces=true,
      showstringspaces=false,
      numbers=none,
      frame=none,
      mathescape=false
    }
  }
}

\makeatletter

\@ifundefined{algobox}{%
  \newtcolorbox{algobox}[2][]{%
    AppendixColoredBoxBase,
    colframe=black,
    colback=black!2,
    colbacktitle=black,
    title=#2,
    #1
  }%
}{%
  \renewtcolorbox{algobox}[2][]{%
    AppendixColoredBoxBase,
    colframe=black,
    colback=black!2,
    colbacktitle=black,
    title=#2,
    #1
  }%
}

\@ifundefined{questionnairebox}{%
  \newtcolorbox{questionnairebox}[2][]{%
    AppendixColoredBoxBase,
    colframe=teal!65!black,
    colback=teal!3,
    colbacktitle=teal!65!black,
    title=#2,
    #1
  }%
}{%
  \renewtcolorbox{questionnairebox}[2][]{%
    AppendixColoredBoxBase,
    colframe=teal!65!black,
    colback=teal!3,
    colbacktitle=teal!65!black,
    title=#2,
    #1
  }%
}

\@ifundefined{appendixquestionbox}{%
  \newtcolorbox{appendixquestionbox}[2][]{%
    AppendixColoredBoxBase,
    colframe=cyan!65!black,
    colback=cyan!3,
    colbacktitle=cyan!65!black,
    title=#2,
    #1
  }%
}{%
  \renewtcolorbox{appendixquestionbox}[2][]{%
    AppendixColoredBoxBase,
    colframe=cyan!65!black,
    colback=cyan!3,
    colbacktitle=cyan!65!black,
    title=#2,
    #1
  }%
}

\@ifundefined{minedresponsesbox}{%
  \newtcolorbox{minedresponsesbox}[2][]{%
    AppendixColoredBoxBase,
    colframe=brown!70!black,
    colback=brown!3,
    colbacktitle=brown!70!black,
    title=#2,
    #1
  }%
}{%
  \renewtcolorbox{minedresponsesbox}[2][]{%
    AppendixColoredBoxBase,
    colframe=brown!70!black,
    colback=brown!3,
    colbacktitle=brown!70!black,
    title=#2,
    #1
  }%
}

\@ifundefined{systempromptbox}{%
  \newtcolorbox{systempromptbox}[2][]{%
    AppendixColoredBoxBase,
    colframe=blue!60!black,
    colback=blue!3,
    colbacktitle=blue!60!black,
    title=#2,
    #1
  }%
}{%
  \renewtcolorbox{systempromptbox}[2][]{%
    AppendixColoredBoxBase,
    colframe=blue!60!black,
    colback=blue!3,
    colbacktitle=blue!60!black,
    title=#2,
    #1
  }%
}

\@ifundefined{userpromptbox}{%
  \newtcolorbox{userpromptbox}[2][]{%
    AppendixColoredBoxBase,
    colframe=green!45!black,
    colback=green!3,
    colbacktitle=green!45!black,
    title=#2,
    #1
  }%
}{%
  \renewtcolorbox{userpromptbox}[2][]{%
    AppendixColoredBoxBase,
    colframe=green!45!black,
    colback=green!3,
    colbacktitle=green!45!black,
    title=#2,
    #1
  }%
}

\@ifundefined{instructionbox}{%
  \newtcolorbox{instructionbox}[2][]{%
    AppendixColoredBoxBase,
    colframe=purple!60!black,
    colback=purple!3,
    colbacktitle=purple!60!black,
    title=#2,
    #1
  }%
}{%
  \renewtcolorbox{instructionbox}[2][]{%
    AppendixColoredBoxBase,
    colframe=purple!60!black,
    colback=purple!3,
    colbacktitle=purple!60!black,
    title=#2,
    #1
  }%
}

\@ifundefined{workshopbox}{%
  \newtcolorbox{workshopbox}[2][]{%
    AppendixColoredBoxBase,
    colframe=red!65!black,
    colback=red!3,
    colbacktitle=red!65!black,
    title=#2,
    #1
  }%
}{%
  \renewtcolorbox{workshopbox}[2][]{%
    AppendixColoredBoxBase,
    colframe=red!65!black,
    colback=red!3,
    colbacktitle=red!65!black,
    title=#2,
    #1
  }%
}

\@ifundefined{dataexamplebox}{%
  \newtcolorbox{dataexamplebox}[2][]{%
    AppendixColoredBoxBase,
    colframe=orange!70!black,
    colback=orange!4,
    colbacktitle=orange!70!black,
    title=#2,
    #1
  }%
}{%
  \renewtcolorbox{dataexamplebox}[2][]{%
    AppendixColoredBoxBase,
    colframe=orange!70!black,
    colback=orange!4,
    colbacktitle=orange!70!black,
    title=#2,
    #1
  }%
}

\makeatother

\newtcblisting{systempromptbox}[2][]{%
  AppendixColoredListingBase,
  colframe=blue!60!black,
  colback=blue!3,
  colbacktitle=blue!60!black,
  title=#2,
  listing only,
  #1
}

\newtcblisting{userpromptbox}[2][]{%
  AppendixColoredListingBase,
  colframe=green!45!black,
  colback=green!3,
  colbacktitle=green!45!black,
  title=#2,
  listing only,
  #1
}

\newtcblisting{instructionbox}[2][]{%
  AppendixColoredListingBase,
  colframe=purple!60!black,
  colback=purple!3,
  colbacktitle=purple!60!black,
  title=#2,
  listing only,
  #1
}

\newtcblisting{workshopbox}[2][]{%
  AppendixColoredListingBase,
  colframe=red!65!black,
  colback=red!3,
  colbacktitle=red!65!black,
  title=#2,
  listing only,
  #1
}

\newtcblisting{dataexamplebox}[2][]{%
  AppendixColoredListingBase,
  colframe=orange!70!black,
  colback=orange!4,
  colbacktitle=orange!70!black,
  title=#2,
  listing only,
  #1
}

\newcommand{\ScaleEq}[2][0.99]{%
  \resizebox{#1\linewidth}{!}{\mbox{$\displaystyle #2$}}%
}

\newcommand{\eqtight}{\setlength{\abovedisplayskip}{3pt plus 1pt minus 1pt}\setlength{\belowdisplayskip}{3pt plus 1pt minus 1pt}\setlength{\abovedisplayshortskip}{0pt plus 1pt}\setlength{\belowdisplayshortskip}{3pt plus 1pt minus 1pt}}

\newcommand{\safeincludegraphics}[2][]{%
  \IfFileExists{#2}{\includegraphics[#1]{#2}}{\fbox{Missing file: #2}}%
}

\definecolor{ColorA}{HTML}{B0B0B0}  
\definecolor{ColorB}{HTML}{6A8ACF}  
\definecolor{ColorC}{HTML}{8FD68F}  
\definecolor{ColorD}{HTML}{C999D1}  
\definecolor{ColorE}{HTML}{F7D78A}  

\newcommand{\ie}{{\it i.e.,\ }}
\newcommand{\tool}{{\it QuAD }}
\newcommand{\toole}{{\it QuAD}}

\newcommand{\lp}[1]{\textcolor[rgb]{0, 0, 0}{#1}}
\newcommand{\qq}[1]{\textcolor[rgb]{0.4, 0.4, 1.0}{Qiqi: #1}}

\definecolor{rtcolor}{rgb}{0.0, 0.42, 0.85}

\ifquadclean
  \renewcommand{\lp}[1]{#1}
  
  \renewcommand{\qq}[1]{}

\fi
\ifquadrevision
  \newcommand{\ws}[1]{#1}
  \newcommand{\resrev}[1]{#1}
  \newcommand{\clsecref}[2]{#2 (Sec.~\ref{#1})}
\else
  \newcommand{\ws}[1]{#1}
  \newcommand{\resrev}[1]{#1}
  \newcommand{\clsecref}[2]{#2}
\fi

\usepackage{lineno}
\usepackage{pdfpages}

\definecolor{darkblue}{rgb}{0, 0, 0.5}
\hypersetup{colorlinks=true, citecolor=darkblue, linkcolor=darkblue, urlcolor=darkblue}

\title{\toole: Quantitative Assessment via Distribution-Alignment for Qualitative Data Quality Enabled by a Content-Aware Multi-Agent System}

\author{
  Wenshuo Zhang$^{1}$\thanks{First author: \texttt{wzhangeb@connect.ust.hk}},
  Leixian Shen$^{1}$, Qiqi Zhou$^{1}$, Shuchang Xu$^{1}$, Feiyang Hao$^{2}$ \\
  {\bf Huamin Qu$^{1}$, Haobo Li$^{1}$, Zhao Chen$^{3}$,
  Lin-Ping Yuan$^{1}$\thanks{Corresponding author: \texttt{yuanlp@ust.hk}}} \\
  $^{1}$The Hong Kong University of Science and Technology \\
  $^{2}$Shanghai Jiao Tong University \\
  $^{3}$The Hong Kong University of Science and Technology (Guangzhou)
}

\begin{document}
\ifquadrevision
  \onecolumn
  \nolinenumbers
  \hypertarget{package-contents}{}
\pdfbookmark[0]{Part I: Package Contents}{bookmark-package-contents}

\begin{center}
{\Large Complete Two-Round Revision Package}\\[4pt]
{\large Current Submission 5875; Previous Submission 3322}
\end{center}

\section*{Part I: Package Contents}

This document preserves the two revision rounds as separate, self-contained records.

\begin{enumerate}
  \item \textbf{\hyperlink{current-cover}{Part II: Current-Round Revision}}
  \begin{enumerate}
    \item \hyperlink{current-cover}{Current-round cover letter};
    \item \hyperlink{current-comments}{Complete current-round reviewer and Area Chair comments};
    \item \hyperlink{current-manuscript}{Current revised manuscript}, with only current-round changes highlighted in
    \textcolor[rgb]{1.0,0.4,0.4}{pink}.
  \end{enumerate}

  \item \textbf{\hyperlink{previous-cover}{Part III: Previous-Round Revision}}
  \begin{enumerate}
    \item \hyperlink{previous-cover}{Previous-round cover letter};
    \item \hyperlink{previous-comments}{Complete previous-round reviewer and Area Chair comments};
    \item \hyperlink{previous-manuscript}{Archived previous-round revised manuscript}, with previous-round changes
    highlighted in \textcolor[rgb]{0.1,0.35,0.9}{blue}.
  \end{enumerate}
\end{enumerate}

\paragraph{Color separation.}
Previous-round revisions are rendered as regular black text in the current manuscript,
so the two colors identify changes from their respective revision rounds without overlap.

  \clearpage
  \hypertarget{current-cover}{}
  \pdfbookmark[0]{Part II: Current-Round Revision}{bookmark-current-round}
  \begingroup
  \renewcommand{\ws}[1]{#1}
  \renewcommand{\resrev}[1]{#1}
  \renewcommand{\lp}[1]{#1}
  \begin{center}
{\Large \lp{Reviewer-Wise Cover Letter}}\\[4pt]
{\large \lp{Round-Two Revision of Submission 5875}}
\end{center}

\lp{We thank the Area Chair and reviewers for their careful assessment. The prior revision established QuAD as a pre-analysis assessment and response-triage aid that complements researcher-led qualitative coding. Building on that scope, this round makes its role as an iterative QDA checkpoint explicit and adds external-label validation, repeated-run reliability, pipeline baselines, measured inference cost, and a three-evaluator report assessment. To improve readability, we consolidate the appendix into five top-level sections, retain headline findings in the main text, and map each Method, Experiment, and Case Study claim to the corresponding appendix protocol or table.}

\section*{\lp{Response to Area Chair 3kpK}}

\paragraph{\lp{AC-1: Positioning QuAD within QDA practice.}}
\begin{itemize}
  \item \lp{\textbf{Concern.} The Area Chair requested a clearer bridge between QuAD and established qualitative-research practice. In particular, the review asked whether a quantitative assessment changes the iterative process of data collection, and how QuAD supports that process while qualitative researchers retain control over sampling, coding, interpretation, and theory building.}
  \item \lp{\textbf{Response.} The prior revision established QuAD as pre-analysis support, with coding and interpretation retained by researchers. This revision makes the operational relationship to iterative QDA explicit: QuAD can assess the responses currently available in a pilot corpus, be re-run as data accumulate, and provide updated signals when the sample, study scope, or analytical lens changes. Researchers use these signals alongside close reading and retain all sampling, coding, saturation, and interpretive decisions.}
  \item \lp{\textbf{Where changed.} The workflow clarification appears in the \clsecref{Sec:qha_discussion}{Supporting Iterative Qualitative Data Analysis} appendix, together with a table that maps signal combinations to next analytical steps. Supporting context is provided in the \clsecref{Sec:background_implementation_appendix}{Background and Reproducibility Materials} appendix, specifically \clsecref{sec:appendix_related_work}{Extended Background}, and the workflow is instantiated in the \clsecref{Sec.Exp_case_study}{main-text Case Study}.}
\end{itemize}

\paragraph{\lp{AC-2: External and human validation.}}
\begin{itemize}
  \item \lp{\textbf{Concern.} Beyond automated and synthetic-data evaluation, the Area Chair requested an external human-subject assessment involving qualitative researchers. The purpose is to determine whether QuAD's lens-level conclusions and prioritized responses agree with independent qualitative judgments and provide useful support for subsequent coding.}
  \item \lp{\textbf{Response.} We provide two complementary forms of external evidence. First, pre-existing human-coded labels independently test the lens-level conclusion, response distances, and retrieval neighborhoods: grade has weak label association (\(V=0.130\)); same- and different-label mean distances are \(14.61\) and \(30.38\); and top-3/top-5 purity reaches \(0.742/0.694\), compared with \(0.489/0.489\) for random retrieval. Second, three evaluators recruited through a WeChat advertisement, mostly with HCI or human-centered research backgrounds, first viewed the raw responses grouped by grade without access to the QuAD report, then reviewed the report and rated its correctness, interpretability, and usefulness for subsequent coding on seven-point scales. Median ratings were \(6\) for all three statements (correctness range \(4\)--\(7\); interpretability and usefulness range \(6\)--\(7\)), and no rating fell below the scale midpoint.}
  \item \lp{\textbf{Where changed.} The headline evidence appears in the \clsecref{Sec.Exp_case_study}{main-text Case Study}. Full details are in the \clsecref{Sec:case_study_full_appendix}{Real-World Case Study and External Validation} appendix, specifically \clsecref{Sec:external_validation_appendix}{External Validation against Human-Coded Labels} and \clsecref{Sec:expert_validation_appendix}{Independent Evaluator Validation}.}
\end{itemize}

\paragraph{\lp{AC-3: Main throughline and presentation.}}
\begin{itemize}
  \item \lp{\textbf{Concern.} As the revision accumulated additional material, the paper's main research question, design, experiments, and findings became difficult to follow. The Area Chair noted that essential evidence was being carried by a lengthy appendix and requested a clearer main throughline with more readable access to supporting results.}
  \item \lp{\textbf{Response.} We now state the headline findings in the main text and link each methodological or robustness claim directly to its supporting protocol or table. In the Experiment section, the former standalone ablation study is merged into Robustness Checks, which now presents four labeled checks---adaptive-anchor ablation, dissimilarity baselines, cross-model and repeated-run stability, and reconstruction sensitivity---each stating its headline result and appendix location. In the Method and Case Study, appendix pointers now sit at the sentence they support rather than interrupting the argument, and repeated setup details are stated once, so that each subsection reads as a single line of reasoning. The Case Study follows one sequence: setup and external criterion, dataset-level verdict with the weak-versus-strong lens contrast, response-level validation, and the evaluator study. We also reorganize the appendix from nine top-level sections into five coherent parts---responsible use and limitations, supplementary evaluation, real-world validation, experimental materials, and background and reproducibility materials---so that readers can locate the relevant evidence more efficiently.} \lp{The abstract, introduction, and conclusion now present the quantitative fitness-for-purpose assessment as the main contribution, with response prioritization as an additional output of the same analysis.}
  \item \lp{\textbf{Where changed.} The revised throughline and mappings appear in \clsecref{Sec:intro_system}{Content-Aware Dissimilarity Extraction}, \clsecref{Sec:Framework_Intro}{Data-Distribution-Driven Qualitative Dataset Assessment}, and \clsecref{sec:experiment}{Experiment}, specifically \clsecref{Sec.Exp_Additional_Checks}{Robustness Checks}, and \clsecref{Sec.Exp_case_study}{Case Study}, with supporting evidence organized in the top-level \clsecref{Sec:revision_support}{Supplementary Evaluation, Robustness, and Deployment}, \clsecref{Sec:case_study_full_appendix}{Real-World Case Study and External Validation}, and \clsecref{Sec:experimental_materials_appendix}{Experimental Materials} appendices.} \lp{The framing is stated in the \clsecref{sec:introduction}{Introduction} and the Conclusion.}
\end{itemize}

\section*{\lp{Response to Reviewer 5R54}}

\paragraph{\lp{5R54-1: Inference scalability and deployment cost.}}
\begin{itemize}
  \item \lp{\textbf{Concern.} The reviewer questioned whether repeated LLM judgments remain computationally practical as the number of responses grows. The review also requested efficient embedding-distance baselines to determine whether generic semantic representations could provide comparable dissimilarity estimates at lower inference cost.}
  \item \lp{\textbf{Response.} We evaluate both questions directly. Pairwise judgment requires \(O(N^2)\) LLM comparisons, whereas scalar projection scores each response once with \(O(N)\) LLM calls and computes pairwise differences locally. Scalar scoring of all 122 food-choice responses used \(56{,}727\) API tokens at an estimated cost of \$0.0060 with Qwen-Flash. Three embedding families provide efficient baselines, with Spearman correlations of \(0.594\)--\(0.621\) in the pairwise setting and \(0.614\)--\(0.698\) under scalar projection; task-conditioned GLM and DeepSeek-v3 reach \(0.830/0.726\) and \(0.797/0.827\), respectively. The reported cost boundary covers scoring calls and excludes instruction generation, retries, and local computation.}
  \item \lp{\textbf{Where changed.} The embedding comparison is summarized under ``Dissimilarity baselines'' in \clsecref{Sec.Exp_Additional_Checks}{Robustness Checks}, and the cost summary at the start of \clsecref{sec:experiment}{Experiment}. Full results are organized in the \clsecref{Sec:revision_support}{Supplementary Evaluation, Robustness, and Deployment} appendix, specifically \clsecref{Sec:additional_baselines_appendix}{Baseline and Clustering Robustness} and \clsecref{Sec:scalability_appendix}{Scalability and Deployment Cost}.}
\end{itemize}

\paragraph{\lp{5R54-2: Purpose and robustness of clustering evaluation.}}
\begin{itemize}
  \item \lp{\textbf{Concern.} The purpose and setup of the clustering evaluation were unclear. Specifically, the review left open whether Adaptive Anchor Manifold Maintenance was itself a clustering method, whether clustering occurred before or after dissimilarity extraction, and which variables were held fixed when comparing clustering algorithms.}
  \item \lp{\textbf{Response.} Adaptive anchors maintain semantic references during LLM-based dissimilarity extraction; fixed-\(K\) clustering is a subsequent reconstruction step. The sensitivity check therefore holds the synthetic data, QuAD dissimilarity matrix, and researcher-specified \(K\) fixed while varying only K-medoids, agglomerative, spectral, and K-means reconstruction, with and without ensembling. The K-means ensemble is strongest overall (ARI \(0.675\), NMI \(0.743\), purity \(0.850\)), while the alternatives recover comparable ground-truth structure.} \lp{We also check the two-dimensional MDS step that precedes clustering: on the case-study matrix it preserves the input distances closely (Kruskal stress-1 \(0.046\), Pearson \(r=0.997\)).}
  \item \lp{\textbf{Where changed.} The distinction and headline finding appear under ``Reconstruction sensitivity'' in \clsecref{Sec.Exp_Additional_Checks}{Robustness Checks}. The complete protocol and table are in the \clsecref{Sec:revision_support}{Supplementary Evaluation, Robustness, and Deployment} appendix, specifically \clsecref{Sec:additional_baselines_appendix}{Baseline and Clustering Robustness}.}
\end{itemize}

\paragraph{\lp{5R54-3: Experimental narrative and appendix presentation.}}
\begin{itemize}
  \item \lp{\textbf{Concern.} The reviewer found the Experiments and Results difficult to navigate because several tables were presented without sufficient introduction or interpretation. The review also identified broken or cramped example boxes in the appendix and requested a clearly identifiable location for the limitations.}
  \item \lp{\textbf{Response.} The Experiment section now introduces each supporting appendix block at the point of use and states the corresponding headline result before directing readers to the full table. In particular, the ablation study and the baseline, reliability, and clustering checks are consolidated into one Robustness Checks subsection with four labeled parts, each giving its key numbers and the appendix subsection with the full protocol; the ablation setting and variant definitions move to the ablation appendix. Within the appendix, every result table is now referenced and introduced in the subsection that discusses it. Prompt and data examples use full-width containers, the three detailed algorithms are typeset as page-wide boxes that no longer break across columns, and obsolete hard-coded appendix numbering has been removed. Limitations now appear as a dedicated section after the Conclusion and before the references, followed by an Ethics Statement.}
  \item \lp{\textbf{Where changed.} Narrative and table mappings appear in \clsecref{sec:experiment}{Experiment}, specifically \clsecref{Sec.Exp_Additional_Checks}{Robustness Checks}, and the \clsecref{Sec:revision_support}{Supplementary Evaluation, Robustness, and Deployment} appendix, specifically \clsecref{Sec:ablation_appendix}{Ablation Details for Adaptive Anchor Updating}. Layout materials appear in the top-level \clsecref{Sec:experimental_materials_appendix}{Experimental Materials} and \clsecref{Sec:background_implementation_appendix}{Background and Reproducibility Materials} appendices. Limitations are in the Limitations section following the Conclusion.}
\end{itemize}

\section*{\lp{Response to Reviewer 31x4}}

\paragraph{\lp{31x4-1: Outside criteria on real qualitative material.}}
\begin{itemize}
  \item \lp{\textbf{Concern.} The main validity evidence recovered structure embedded in synthetic data but did not compare QuAD with an independent target on real qualitative material. The reviewer requested outside criteria such as expert readiness judgments, human dissimilarity, coding yield, thematic richness, or saturation to test whether the reported scores and response rankings correspond to meaningful qualitative structure.}
  \item \lp{\textbf{Response.} We use pre-existing human-coded labels as independent proxies for three requested targets. Grade--label association evaluates the lens-level conclusion over all 122 responses (\(V=0.130\)); same- versus different-label distances evaluate pair-level dissimilarity over the 119 non-unclear responses (\(14.61\) versus \(30.38\)); and top-3/top-5 neighborhood purity evaluates coding-oriented retrieval (\(0.742/0.694\), versus \(0.489/0.489\) for random retrieval). The evaluator study separately examines whether analysts judge the QuAD report consistent with their own reading and its prioritized responses useful starting points for coding. Broader effects on thematic richness and saturation require dedicated downstream studies.}
  \item \lp{\textbf{Where changed.} The summary appears in the \clsecref{Sec.Exp_case_study}{main-text Case Study}. Protocols and complete results are in the \clsecref{Sec:case_study_full_appendix}{Real-World Case Study and External Validation} appendix, specifically \clsecref{Sec:external_validation_appendix}{External Validation against Human-Coded Labels} and \clsecref{Sec:expert_validation_appendix}{Independent Evaluator Validation}.}
\end{itemize}

\paragraph{\lp{31x4-2: Repeated-run and cross-model reliability.}}
\begin{itemize}
  \item \lp{\textbf{Concern.} Because QuAD is presented as a measurement framework, the reviewer asked whether its distances and four assessment metrics remain stable across repeated runs, model families, raters, prompts, and pairwise comparisons. Without such reliability evidence, variation in the LLM judge could undermine the validity of the resulting dataset-level conclusions.}
  \item \lp{\textbf{Response.} We provide an initial fixed-prompt reliability test on a controlled synthetic slice. With inputs, prompts, and clustering randomness held fixed, Qwen-max and DeepSeek-v3 were each run independently three times. Within-model distance Spearman correlations are \(0.80\) and \(0.94\), cross-model agreement is \(0.87\), and standard deviations across GMR, DDR, DC, and DP range from \(0.007\) to \(0.050\). This experiment directly covers repeated-run and cross-model stability under the tested protocol; the evaluator study adds report-level human agreement, while prompt-perturbation and human pairwise-rating reliability remain extensions.}
  \item \lp{\textbf{Where changed.} The headline result appears under ``Cross-model and repeated-run stability'' in \clsecref{Sec.Exp_Additional_Checks}{Robustness Checks}. The complete reliability protocol and table are in the \clsecref{Sec:revision_support}{Supplementary Evaluation, Robustness, and Deployment} appendix, specifically \clsecref{Sec:reliability_appendix}{Repeated-Run and Cross-Model Reliability}; human agreement is reported in the \clsecref{Sec:case_study_full_appendix}{Real-World Case Study and External Validation} appendix, specifically \clsecref{Sec:expert_validation_appendix}{Independent Evaluator Validation}.}
\end{itemize}

\paragraph{\lp{31x4-3: Necessity of the full pipeline.}}
\begin{itemize}
  \item \lp{\textbf{Concern.} The reviewer questioned whether QuAD's multi-agent instruction generation, adaptive anchors, and reconstruction pipeline were necessary for the reported signal. Simpler alternatives—including direct LLM prompting, embedding distances, standard clustering, or distributional baselines—could potentially produce comparable results unless each stage was evaluated against an appropriate baseline and, where possible, an external target.}
  \item \lp{\textbf{Response.} We add stage-specific comparisons. Direct prompting versus QuAD instruction generation yields Spearman correlations of \(-0.036/0.725\) for Qwen-max, \(0.599/0.830\) for GLM, and \(0.331/0.726\) for DeepSeek-v3. In the adaptive-anchor ablation, the full configuration ranks first in five of ten model--regime combinations and within the top two in nine\lp{, has the highest mean Spearman correlation in both regimes (pairwise \(0.743\) vs.\ \(\le0.726\); scalar \(0.776\) vs.\ \(\le0.775\)), and has the highest worst case across the ten settings (\(0.708\) vs.\ \(\le0.659\)); single ablations can win for a specific model, but none is consistently better}. Embedding baselines and fixed-\(K\) clustering sensitivity evaluate the remaining stages. Under the human-coded labels, random, TF--IDF, and QuAD obtain mismatch AUCs of \(0.500/0.551/0.749\), top-3 purities of \(0.489/0.650/0.742\), and top-5 purities of \(0.489/0.619/0.694\). \lp{As a distributional baseline for \lp{the group-separation component of the assessment}\lp{, which also serves as a convergent-validity check of DDR}, standard association tests (\(\chi^2\), Kruskal--Wallis, PERMANOVA) applied to the same QuAD scores on nine lenses reach the same grade-level verdict as DDR (\(p\ge0.27\)) and detect the human code (\(p\le0.002\)); \lp{KW and PERMANOVA also detect current diet and sports, where both effect sizes are small (\(\varepsilon^2\le0.04\), DDR \(=0.03\), versus \(0.45\) and \(0.52\) for the human code), so these differences are detectable but explain little of the response structure. We therefore read DDR as an effect size reported alongside the tests rather than as a competing test, and recommend reporting both with the four metrics, which add cluster-level readings and response-level triage.} \lp{The appendix now also states what the profile adds beyond a test: for grade level, a non-significant test is consistent both with homogeneous responses and with varied responses organized along another axis, whereas the profile separates these cases, since high DC (\(0.90\)) with low GMR, DDR, and DP (\(0.39\), \(0.02\), \(0.43\)) shows varied responses whose themes cut across grades and points to revising the lens or coding cross-group themes.}} \lp{Beyond these tests, \lp{the four metrics are read jointly as one profile, from lens-level match and separation to cluster mixing and within-cluster spread, which indicates where close reading is needed; in the controlled distribution-shift experiment they respond in different patterns to different structural changes, and the profile} applies to the pairwise regime as well; ranking responses by distance to the group median score reproduces 23 of QuAD's 24 grade-level picks, as expected in the scalar regime\lp{; this shortcut does not exist in the pairwise regime, where no single score is available}. On the human-coded labels, generic embeddings reach slightly higher neighbour purity than QuAD, whereas QuAD has the highest mismatch AUC\lp{, so its advantage lies in criterion-specific discrimination rather than in generic neighbourhood structure}.} Together, these experiments support the complete\lp{, task-conditioned} configuration as \lp{the most reliable default}, while retaining the observed model-specific ablation variation.}
  \item \lp{\textbf{Where changed.} The direct-prompting comparison appears in the instruction-quality table of \clsecref{Sec.Exp_User_Study}{User Study for Instruction Generation}, and the main synthesis appears under ``Adaptive-anchor ablation'' and ``Dissimilarity baselines'' in \clsecref{Sec.Exp_Additional_Checks}{Robustness Checks}. Component-level comparisons are organized in the \clsecref{Sec:revision_support}{Supplementary Evaluation, Robustness, and Deployment} appendix, specifically \clsecref{Sec:ablation_appendix}{Ablation Details for Adaptive Anchor Updating} and \clsecref{Sec:pipeline_necessity_appendix}{Pipeline Component Necessity Checks}; the real-label baseline appears in the \clsecref{Sec:case_study_full_appendix}{Real-World Case Study and External Validation} appendix, specifically \clsecref{Sec:external_validation_appendix}{External Validation against Human-Coded Labels}\lp{ and \clsecref{Sec:code_then_test_appendix}{Code-then-Test Check of Group Separation}}\lp{; the positioning against both simpler alternatives is summarized in \clsecref{sec:appendix_related_work}{Extended Background}}.}
\end{itemize}

\paragraph{\lp{31x4-4: Reproducibility details.}}
\begin{itemize}
  \item \lp{\textbf{Concern.} The reviewer found that exact model identifiers, randomization controls, code or configuration settings, and parts of the synthetic ground-truth construction were distributed across the paper or underspecified. This made the experiments difficult to reconstruct even though prompts, screenshots, and several algorithmic details were available.}
  \item \lp{\textbf{Response.} We consolidate the reproducibility information currently reported in the manuscript: model and inference descriptions, fixed-\(K\) clustering protocol, prompt templates, detailed algorithms, and the synthetic-data generation process. The reliability subsection additionally states which inputs, prompts, and clustering randomness are held fixed, making the reported experimental configuration easier to audit.} \lp{The quantity table states that the cluster number is set to \(K=C\), and the prompt appendix states that the returned difference score is used as the dissimilarity.} \lp{The Method and the quantity table now also state the pairwise/scalar assignment used in all experiments: the 15-response datasets use pairwise scoring and the 90- and 122-response datasets scalar projection, an assignment reproduced by any \(N_{\text{threshold}}\in[15,90)\).}
  \item \lp{\textbf{Where changed.} Model controls and reliability details appear in the \clsecref{Sec:revision_support}{Supplementary Evaluation, Robustness, and Deployment} appendix, specifically \clsecref{Sec:reliability_appendix}{Repeated-Run and Cross-Model Reliability}. Synthetic-data construction is in \clsecref{Sec:experimental_materials_appendix}{Experimental Materials}, specifically \clsecref{Sec.Exp_Data_Gen_Sys}{Controllable Synthetic Data Generation System}; prompts and algorithms are consolidated in \clsecref{Sec:background_implementation_appendix}{Background and Reproducibility Materials}.} \lp{The pairwise/scalar assignment is stated in \clsecref{Sec:Anchor_Updating}{Adaptive Anchor Manifold Maintenance} and \clsecref{Sec:metric_parameter_clarification}{Metric and Quantity Clarification}.}
\end{itemize}

\section*{\lp{Response to Reviewer ofpg}}

\paragraph{\lp{ofpg-1: Philosophical scope and iterative QDA workflow.}}
\begin{itemize}
  \item \lp{\textbf{Concern.} The reviewer questioned how quantitative readiness signals fit interpretive QDA traditions. Qualitative researchers commonly begin analysis on a pilot subset and revise sampling, scope, and analytical direction as evidence accumulates, whereas the original presentation could be read as requiring a complete dataset before a single final assessment.}
  \item \lp{\textbf{Response.} The prior revision established QuAD as pre-analysis support under a researcher-selected lens. This revision makes its iterative use explicit: researchers can run QuAD on a pilot corpus, repeat the assessment as responses accumulate, and invoke it again after changing the sample, scope, or analytical lens. The outputs provide diagnostic checkpoints and prioritized cases, while researchers retain coding, saturation, interpretation, and theoretical decisions.}
  \item \lp{\textbf{Where changed.} The workflow is detailed in the \clsecref{Sec:qha_discussion}{Supporting Iterative Qualitative Data Analysis} appendix. Its relationship to established QDA is discussed in the \clsecref{Sec:background_implementation_appendix}{Background and Reproducibility Materials} appendix, specifically \clsecref{sec:appendix_related_work}{Extended Background}.}
\end{itemize}

\paragraph{\lp{ofpg-2: Alignment, qualitative richness, and societal risks.}}
\begin{itemize}
  \item \lp{\textbf{Concern.} The reviewer cautioned that structural alignment between metadata groups and response clusters is not equivalent to qualitative richness or validity. The review also identified four societal risks: exclusion of dialectal or cultural nuance, metric-washing that downplays meaningful outliers, cultural homogenization, and overinterpretation of alignment as data validity.}
  \item \lp{\textbf{Response.} We distinguish the signals and \lp{read them jointly as one diagnostic profile}. GMR summarizes lens--cluster alignment, DDR how well the analytical groups separate responses, DC \lp{within-cluster} dispersion, and DP analytical-group mixing; a separate response ranking identifies central and unusual cases. \lp{The Method now states the placement of each metric: general-pattern metrics (GMR, DDR) measure how far the lens explains response structure, unique-insight metrics (DP, DC) describe what it leaves within clusters, data-point metrics (GMR, DP) count individual assignments, and distribution metrics (DDR, DC) aggregate continuous dissimilarities.} Alignment and non-alignment therefore motivate distinct forms of close reading\lp{; the signal table gives reading guides for this inspection rather than automatic decision rules}. The limitations explicitly address linguistic and cultural bias, homogenization, loss of nuance, metric-washing, model drift, and the need for human inspection of instructions, anchors, representative responses, and outliers.}
  \item \lp{\textbf{Where changed.} \lp{The metric placement is in \clsecref{Sec.Problem_Formulation_Assessment}{Problem Formulation} of the assessment framework.} The case-level interpretation appears in the \clsecref{Sec.Exp_case_study}{main-text Case Study}. The conceptual boundary and safeguards are in the Limitations section following the Conclusion.}
\end{itemize}

\paragraph{\lp{ofpg-3: Novelty, deployment trade-offs, and evidence breadth.}}
\begin{itemize}
  \item \lp{\textbf{Concern.} The reviewer found it difficult to separate the paper's technical novelty from existing LLM-as-a-judge stabilization methods. The review also noted that the method was dense within the page limit, that real-world evidence came from a single corpus, and that computational and human deployment costs had not been weighed clearly against practical utility.}
  \item \lp{\textbf{Response.} We center the technical distinction on two components: a task-specific instruction-generation loop that incorporates researcher knowledge, and an adaptive anchor set that maintains a semantic reference frame across dataset-level judgments\lp{; the Method now names the three agent roles and their exchange: an Oracle (LLM in Stage 1, human expert in Stage 2) returns corrected scores and critiques, an Updater LLM rewrites the instruction, and an Evaluator LLM scores responses under the instruction and anchors}. We separate both from downstream reconstruction and connect them to direct-prompt, ablation, embedding, and clustering checks. For deployment, we distinguish the \(O(N^2)\) pairwise and \(O(N)\) scalar regimes and report 56,727 tokens and an estimated \$0.0060 for scalar scoring of 122 responses, together with the one-time human instruction-refinement step and the information loss introduced by scalar projection. To reduce density within the page limit, the Method now introduces each concept once: the fitness-for-purpose rationale is merged into the problem objective, and the pairwise and scalar regimes are described only where they are implemented. The appendix is consolidated to keep results accessible. Within the single real corpus, switching the lens from grade level to the dataset's own eating-habit-change code \lp{shifts the joint profile and} raises DDR from \(0.02\) to \(0.52\) and DP from \(0.43\) to \(0.75\), showing that the weak-lens verdict reflects the lens rather than the instrument\lp{; randomly reassigning part of the human labels lowers DDR monotonically from \(0.52\) to \(0.02\), covering the cases in between}; validation across additional real corpora\lp{, and of pairwise relations beyond a single axis,} is stated as an important extension.}
  \item \lp{\textbf{Where changed.} \lp{The agent roles are stated in \clsecref{Sec:Problem_Formulation_Dissimilarity}{Problem Formulation} of the dissimilarity extraction.} Component evidence and deployment measurements are in the \clsecref{Sec:revision_support}{Supplementary Evaluation, Robustness, and Deployment} appendix, specifically \clsecref{Sec:pipeline_necessity_appendix}{Pipeline Component Necessity Checks} and \clsecref{Sec:scalability_appendix}{Scalability and Deployment Cost}. Positioning and evidence boundaries are discussed in the Limitations section following the Conclusion.}
\end{itemize}

\paragraph{\lp{ofpg-4: Demonstrable value to QDA.}}
\begin{itemize}
  \item \lp{\textbf{Concern.} The paper demonstrated pre-analysis signals and response triage but did not establish whether using QuAD changes the quality of downstream QDA outcomes, such as the resulting codes, themes, or interpretations. The reviewer therefore asked for real-world evidence of value to qualitative analysis itself.}
  \item \lp{\textbf{Response.} The human-coded-label results show that QuAD retrieves thematically coherent neighborhoods and separates same- from different-label pairs, supporting its pre-analysis triage role. The three-evaluator study adds report-level evidence of perceived correctness and usefulness for subsequent coding: after reviewing the report, evaluators gave median ratings of \(6/7\) for correctness, interpretability, and usefulness for subsequent coding, with eight of nine ratings at \(6\) or \(7\). A controlled coding study comparing final themes and interpretations with and without QuAD remains necessary to establish downstream outcome effects.}
  \item \lp{\textbf{Where changed.} The headline evidence appears in the \clsecref{Sec.Exp_case_study}{main-text Case Study}. Full label-based and evaluator evidence is in the \clsecref{Sec:case_study_full_appendix}{Real-World Case Study and External Validation} appendix, specifically \clsecref{Sec:external_validation_appendix}{External Validation against Human-Coded Labels} and \clsecref{Sec:expert_validation_appendix}{Independent Evaluator Validation}. The evidence boundary is stated in the \clsecref{Sec:qha_discussion}{Supporting Iterative Qualitative Data Analysis} appendix.}
\end{itemize}

\lp{Together, these revisions add external-label, independent-evaluator, reliability, baseline, and deployment evidence for QuAD as a quantitative pre-analysis signal that informs researcher-led qualitative inquiry.}

  \endgroup
  \clearpage
  \hypertarget{current-comments}{}
  \pdfbookmark[1]{Current-Round Reviewer Comments}{bookmark-current-comments}
  \section*{Round-Two Official Reviewer Comments}

\section*{Area Chair 3kpK}

\paragraph{Meta Review of Submission 5875.}
This is a revised submission of QuAD, a method for the quantitative evaluation of qualitative data. QuAD evaluates the dataset- and instance-level quality, incorporating both automated and human evaluations.

The authors did a nice job addressing comments from the prior round in this revised submission. At the same time, additional revisions are still needed. As shown in the reviews, due to the nature of the contribution, care needs to be taken to very clearly articulate the contribution, address the fact that QuAD by design is fundamentally altering a key step in the data collection process. As such, it needs to be demonstrated how it will improve the existing process and validated in a pilot study/human subject experiment with qualitative researchers to assess their thoughts on QuAD. Reviewer ofpg gives a clear description of the steps needed to build this bridge between the paradigms.

In addition, a lot has been added since the last submission, with more results being added during this response period. The cost of this is that the manuscript is now becoming bloated, with a lot of key information pushed to appendices, and key results in the main paper squeezed into Figures 3 and 4, making them very difficult to read. At this point I would recommend that the authors step back, assess the key elements needed in the main manuscript, and redesign the manuscript to best present the main throughline of the research question, design, experiments, and results. Having robust appendices is fine, but they cannot be doing the heavy lifting.

With all of the baove in mind, after incorporating the author response content, the manuscript would make a valuable contribution to an ACL ARR partner venue. My score reflects the work done to date while recognizing that with additional revisions the contribution could be much more clear and as such more impactful.

\paragraph{Summary Of Reasons To Publish.}
\begin{itemize}
  \item QuAD is an interesting innovation in qualitative data assessment that can augment the pipeline of a qualitative researcher. It can be an off-the-shelf addition to a qualitative researcher's toolkit
  \item The authors did a nice job revising the manuscript based on feedback from the prior round, which is commendable
  \item The automated evaluations are sound, including the updates made during the response period
\end{itemize}

\paragraph{Summary Of Suggested Revisions.}
\begin{itemize}
  \item The bridge between qualitative researchers and novel NLP contributions needs to be carefully built. QuAD isn't (and shouldn't be) a replacement for a qualitative researcher's toolkit. Therefore, motivating and evaluating QuAD should include qualitative researchers in the loop.
  \item External evaluation (in particular from humans) will strengthen the validity
  \item Figures 3 and 4 are unreadable; they need to be reconfigured so that the content can be read at a reasonable zoom.
\end{itemize}

\section*{Reviewer 5R54}

\paragraph{Official Review of Submission 5875: Paper Summary.}
This paper proposes an automated system for pre-assessing the quality of survey datasets. To alleviate the labor-intensive nature of manual data validation, the authors employ an LLM-as-a-Judge framework to evaluate answer relevance and quality based on the query topic.

To validate their methodology, the authors first evaluated the system using a synthetic dataset with ground-truth labels, demonstrating high Pearson and Spearman correlation coefficients between the LLM-derived dissimilarity scores and the ground truth. Furthermore, they validated the framework on a real-world dataset, showing that their four proposed metrics align closely with domain expert assessments. Finally, a usability evaluation indicates that the agent-assisted system enables users to craft more effective prompts compared to unassisted, manual efforts.

\paragraph{Summary Of Strengths.}
The authors conduct thorough experiments and a analysis to validate their framework. While synthetic datasets are utilized for validation, their inclusion across four distinct domains strongly demonstrates the system's cross-domain robustness and generalizability. Furthermore, the proposed system is highly accessible and straightforward to adopt, offering immediate practical value to researchers working with survey methodologies and dataset evaluation.

\paragraph{Summary Of Weaknesses.}
While the authors demonstrate that LLM scoring is superior to standard embedding similarities, the computational cost and resource intensity of LLM inference remain a practical bottleneck for scaling to massive datasets. Additionally, the clustering evaluation is not clear.

\paragraph{Comments Suggestions And Typos.}
While Sections 1 through 3 are well-structured and clear, the Experiments and Results sections are difficult to navigate and would benefit from a smoother narrative flow.

Several tables are included in the manuscript without explicit citation or discussion in the text; all tables should be properly introduced and analyzed to guide the reader.

Please address the formatting issues in Appendices G.5, D.6.3, and H. prompts, where the example text boxes appear broken. Formatting these boxes to span across both columns would likely resolve the layout rendering errors and significantly improve readability.

Limitations are normally put after the conclusion, before the reference section.

\subsection*{Substantive Follow-Up Acknowledgment by Reviewer 5R54}
Thank you. The computational cost part is clear. The evaluation of the proposed split sounds reasonable.

\section*{Reviewer 31x4}

\paragraph{Official Review of Submission 5875: Paper Summary.}
The paper proposes QuAD, a framework for assessing whether qualitative free-text response datasets are ready for a chosen analysis lens, such as grade level, demographic group, expertise level, or other metadata. It does this by (i) using LLM-based dissimilarity scoring to compare responses, then (ii) Clustering responses into an intrinsic structure, follow by (iii) comparing that structure against a researcher-defined metadata partition. Then, it is (iv) reported on four different metrics (GMR, DDR, DP, DC). It meant to indicate whether the lens supports general patterns or reveals unique/deviant insights. Finally, the authors add a content-aware multi-agent system for prompt/instruction generation and adaptive anchors to stabilize LLM scoring. The paper validates QuAD using synthetic datasets, ablations, robustness checks, a small user study, and a food-choice case study with 122 human-written responses.

\paragraph{Summary Of Strengths.}
\begin{itemize}
  \item The paper addresses a real need, with a focus on pre-analysis triage for qualitative response datasets before researchers commit to full manual coding.
  \item Beyond the four readiness metrics, QuAD also identifies central and unusual responses within the response set. This is a meaningful strength because qualitative coders often need to inspect both typical cases, which reveal common patterns, and deviant cases, which may reveal unexpected themes or exceptions. Even if GMR/DDR/DP/DC require further validation as readiness measures, a ranked set of representative and outlier responses could still help human coders decide where to begin manual review.
  \item I would like to give credit to the authors on this. The paper is quite open about what it did. It shows the prompts, algorithm details, ablations, user-study materials, screenshots, and limitations. That is helpful. But openness is not the same as reproducibility. Some important details are still missing, including exact model versions, seeds, code or config settings, and parts of the ground-truth definition.
\end{itemize}

\paragraph{Summary Of Weaknesses.}
\begin{itemize}
  \item The biggest gap is the absence of an outside criterion. The paper does not test QuAD against expert readiness judgments, human pairwise dissimilarity, downstream coding yield, thematic richness, saturation, or any other real target. Because that target is missing, the main validation depends on synthetic responses with hidden latent scores. That is a weak substitute. It may show that the scoring pipeline can recover structure planted into synthetic data, but it does not show that QuAD measures readiness in real qualitative material.
  \item The reliability evidence is missing. QuAD is presented as a measurement framework, but the paper does not show whether its scores are stable across runs, models, raters, prompts, or pairwise comparisons. This matters because validity requires reliability.
  \item The full QuAD pipeline is not justified. At the moment, QuAD uses multi-agent prompting, adaptive anchors, and four separate metrics. But it is missing an external readiness target to test these choices against. As such, a simple baseline, for example, embedding with clustering, standard distribution metrics or maybe direct LLM scoring, may give the same results too.
\end{itemize}

\paragraph{Comments Suggestions And Typos.}
I suggest that the authors compare QuAD with expert readiness scores, human dissimilarity ratings, or coding yield. With this, it will make validation less circular, and directly quantify if the full pipeline is actually necessary in this scenario. Also, the reliability test must be separated with repeated runs, model agreement, rater agreement, or pairwise-stability checks.

\section*{Reviewer ofpg}

\paragraph{Official Review of Submission 5875: Paper Summary.}
This paper introduces QuAD (Quantitative Assessment via Distribution-Alignment), a framework designed to provide scalable, dataset-level pre-analysis signals for qualitative interviews/data typically collected from semi-structured questions. QuAD looks at two distributions: the extrinsic distribution (the extent to which the dataset provides the coverage specified by the researcher) and intrinsic distribution ( a cluster distribution calculated by the algorithm).

QuAD is trying to provide a dataset-level assessment of the utility of that dataset to the researcher. QuAD does not undertake coding of data per se, rather it assesses the potential of coding in a particular direction. It serves as a data 'weather report' and seeks to provide a measure of data saturation.

QuAD operates through a three-stage pipeline designed to extract task-conditioned semantic dissimilarity and map it onto a four-quadrant assessment framework. Rather than compare two responses to each other (cosine similarity) QuAD effectively assesses the similarity of the responses to the research question. QuAD then calculates two distributions (intrinsic and extrinsic) and then computes a four quadrant assessment.

QuAD is designed to produced a content relevant, research question informed assessment of the potential of the data to be effectively analysed (separately).

THe authors present validation against both a synthetic and a real world dataset.

\paragraph{Summary Of Strengths.}
\begin{itemize}
  \item Undertaking a pre-assessment of an entire qualitative dataset is a novel approach to QDA. QUAD is positioned as augmenting human QDA and not replacing it, but yielding a assessment of the potential of the whole dataset to address the research questions.
  \item For NLP the paper introduces the Adaptive Anchor Manifold which is a new method for stabilising LLM as judge evaluations for open-text/s.
  \item For QDA it offers a scalable mechanism to quantify data saturation and fitness for (analytical) purpose.
  \item In practice it does offer a method of alerting to outliers at an early stage.
  \item It could be used, for example, by a supervisor to check the suitability of data prior to investing human capital in QDA.
\end{itemize}

\paragraph{Summary Of Weaknesses.}
Philosophical Comments: Oddly this paper applies some ideas from quantitative research into QDA, and many QDA researchers would find that unacceptable. The paper measures qualitative research readiness with quantitative signals. THe paper also assumes (wrongly) that whole datasets are available before analysis commences, and that simply is not the typical analytical approach where we would start pilot analyses once two or three interviews (for example) were available. Another feature of QD collection is that we do change the sample and even scope based on the analysis of early ideas as the data collection proceeds.

The algorithm is relatively complex and the paper length constraints its presentation - the 8 page paper becomes 42 pages with the inclusion of the additional materials.

the core idea seems to be to map qualitative dataset utility onto the alignment between metadata from the researcher and semantic clusters of responses. THere is a risk that the authors see structural coherence as the same as qualitative richness, but that would be a conflation.

2). it is difficult to judge the novelty here against other LLM-as-judge stabilisation methods - this is an area the authors wmay wish to pursue.

3). the evidence of utility is confined to a single synthetic dataset and a single real-world case. Competng papers which can demonstrate robustness across corpora will rate more highly.

4). it would have been good to see a discussion of the cost of deployment (computational and human) versus the perceived utility of QuAD in real world application.

value to QDA - the paper doesn;t really explore whether QuAD actually improves QDA - are QDA results better with QuAD??

\paragraph{Comments Suggestions And Typos.}
It is always a challenge to find computational novelty at the same time as improving QDA. For many QDA practitioners, the inclusion of machine learning/LLMs - even in plain vanilla forms - is a philosophical challenge as these designs appear to favour quantitative measures over qualitative methods.

The true benefit is the triaging and alignment of the broad dataset with the research questions.

Real world results with improved QDA outcomes would provide convincing evidence of the utility and desirability of QuAD.

\paragraph{Limitations And Societal Impact.}
THere are four main limitations which the authors should consider:

1). data alignment can be confused with data validity

2). LLM biases always risk exclusion of dialects/cultural nuances

3). QuAD might inadvertently participate in metric-washing whereby meaningful outliers are downplayed

4). cultural homogenisation - when detail and nuance are being sought

None are fatal, but show an awareness of QDA debates.

  \clearpage
  \hypertarget{current-manuscript}{}
  \pdfbookmark[1]{Current-Round Revised Manuscript}{bookmark-current-manuscript}
  \linenumbers
\fi
\maketitle
\begin{abstract}
\ws{Assessing whether qualitative responses collected from semi-structured questions are ready for a particular analysis goal is a persistent challenge. Manual qualitative analysis is interpretive and labor-intensive, while existing automated methods often lack a dataset-level perspective for deciding whether a response set supports a researcher-selected analytical lens. We introduce \toole, a framework for quantitative pre-analysis assessment of qualitative response datasets. 
\tool compares a reconstructed intrinsic cluster distribution reconstructed from responses with an extrinsic, researcher-defined analytical lens derived from metadata. \tool is structured into a four-quadrant assessment \lp{whose metrics are read jointly, from lens-level relations to within-cluster variation, to provide} fitness-for-purpose signals for revealing general patterns and unique insights. Because this assessment relies on \lp{lens-conditioned} inter-response dissimilarity, we propose a content-aware multi-agent system with two-stage instruction generation and adaptive anchors to improve scoring consistency.}
\ws{We validate \tool through controlled experiments on synthetic data, including robustness and baseline comparisons, and a real-world social survey case study. 
\lp{The case study shows how \tool quantitatively assesses whether a selected analysis axis is promising and also surfaces representative or unusual responses for human interpretation.}}
\end{abstract}

\section{Introduction}
\label{sec:introduction}
\begin{figure}[t]
  \centering
  \includegraphics[width=0.80\linewidth]{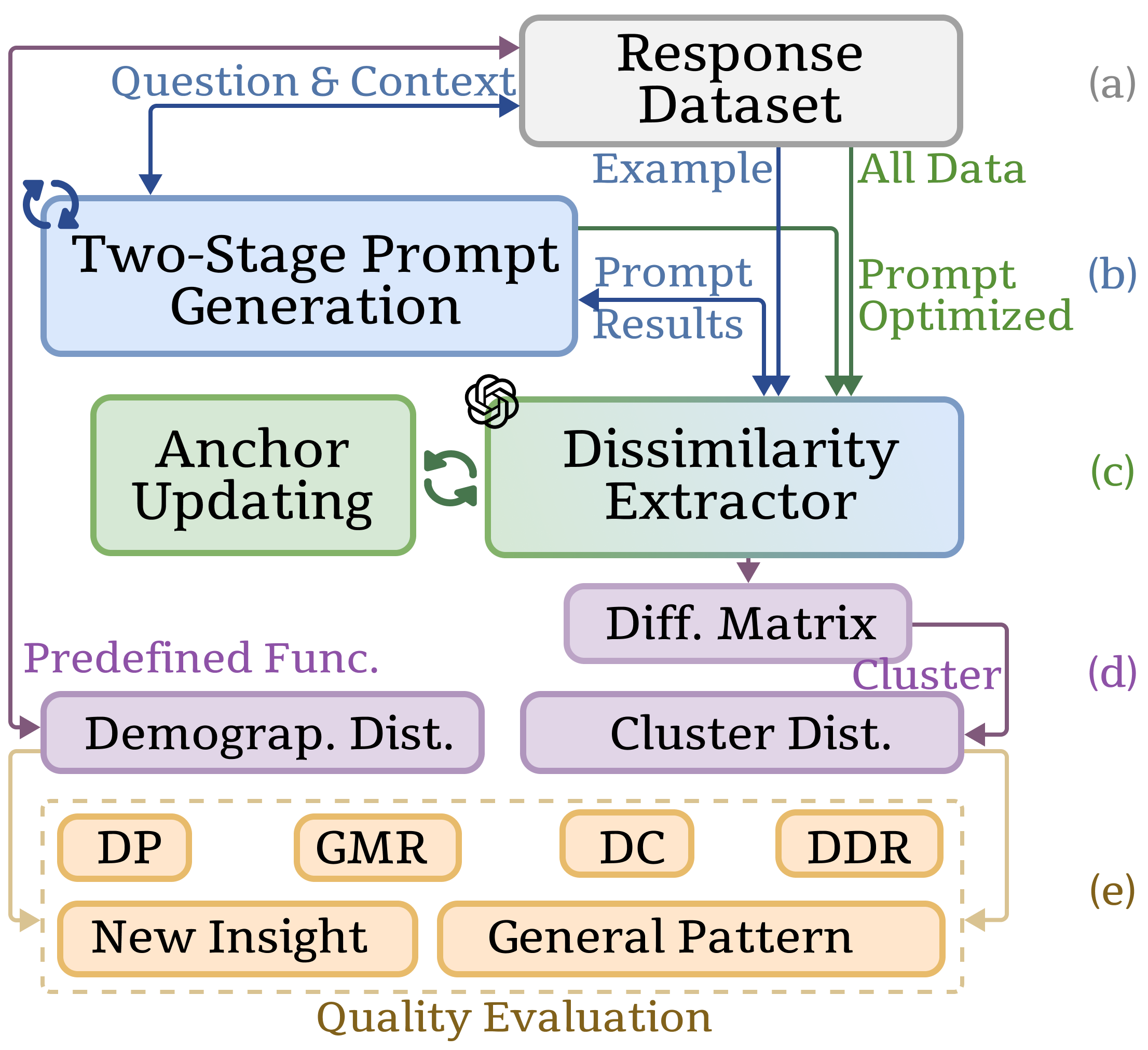}
  \caption{\ws{\lp{Overview} of \toole. \tool takes a response dataset (A), uses a content-aware multi-agent system with iterative prompt generation (B) and adaptive anchor updating (C) to extract dissimilarity, reconstructs cluster and metadata distributions (D), and performs four-quadrant fitness-for-purpose assessment (E).}}
  \label{fig:overview}
  \vspace{-12pt}
\end{figure}

Semi-structured questions play a crucial role in studies involving human participants. They enable researchers to gather detailed qualitative responses, uncovering \textit{general patterns} and \textit{unique insights} across demographic groups (e.g., age, gender).
For example, in a study on student well-being, a question like ``How have your eating habits changed since starting college?'' might reveal general patterns, such as freshmen reporting poorer diets, as well as unique insights from unexpected cases, such as freshmen who are student-athletes maintaining healthy eating habits due to their training \lp{(a reason read by the researcher)}.
\ws{Traditionally, such responses are analyzed through iterative procedures such as familiarization, open coding, theme grouping, interpretation, and validation, with attention to codebooks, intercoder reliability, data saturation, and the richness or thickness of the evidence~\citep{denzin2011sage, glaser2017discovery, campbell2013coding, ames2024assessing, naeem2024demystification}. These methods are essential for interpretation and theory building, but usually \lp{follow} data collection and need substantial human effort.}

\ws{Pre-analysis assessment of a qualitative response dataset is crucial before conducting full coding or interpretation. Without it, researchers risk investing effort in an analysis lens that the collected responses do not support, or overlooking cross-group patterns and unusual cases that deserve closer inspection.}
Existing evaluation methods are insufficient. Manual approaches do not scale and lack quantitative outputs to guide data improvement~\citep{mannually_eval_hard_mattimoe2021approaches, campbell2013coding}. 
Most automated methods evaluate responses in isolation~\citep{chen2023exploringuselargelanguage, swayamdipta2020dataset}, overlooking inter-response relations and dataset structure crucial for comparative knowledge discovery~\citep{fayyad1996data, glaser2017discovery}.

\ws{Rather than focusing on individual responses, we propose \toole, a quantitative pre-analysis assessment method that evaluates a response dataset as a whole, estimating whether a qualitative dataset is fit for a selected analysis goal by comparing its internal response structure with a researcher-defined analytical lens.}
\lp{The lens may be derived from demographic metadata such as age, gender, grade level, or expertise, and acts as an external hypothesis rather than ground truth for qualitative insight. Strong alignment between response structure and the lens suggests that the dataset supports group-level general patterns along this lens. Divergence is also informative: it may signal cross-group consensus, latent factors beyond the selected metadata, or unusual responses worth closer qualitative interpretation.}
\ws{The core idea of \tool is to quantitatively (1) measuring the alignment and divergence between a metadata-derived analytical distribution and a cluster distribution empirically derived from responses, and (2) examining variation within the cluster distribution itself. \lp{It also locates} representative responses in dense regions and candidate unusual ones in outlying or cross-lens regions.}

\ws{More specifically, \tool consists of two components. First, a \textbf{four-quadrant assessment framework} operationalizes the comparison between the cluster and metadata-derived analytical distributions (Sec. \ref{Sec:Framework_Intro}). This framework yields four \lp{complementary metrics, read jointly as one profile,} that characterize a dataset's utility for discovering \textit{general patterns} versus \textit{unique insights} at both the \textit{distribution} and \textit{data-point} levels. The four quadrants rely on the distributions reconstructed via metadata and \lp{inter-response dissimilarities}.}
Therefore, \tool integrates a \textbf{content-aware multi-agent system} to compute \lp{task-conditioned} inter-response dissimilarity (Sec. \ref{Sec:intro_system}). 
To be more reliable and domain-aware, it enhances the LLM-as-a-Judge paradigm~\citep{zheng2023judging} with two innovations.
First, a novel interactive instruction generation system enables efficient, domain-specific instruction generation (Sec.~\ref{Sec:Instruction_Refinement}), moving beyond manual or black-box methods~\citep{zhou2022large, shah2024prompt, pryzant2023automatic, chen2023instructzeroefficientinstructionoptimization}. Second, a dynamic anchor updating algorithm (Sec.~\ref{Sec:Anchor_Updating}) maintains a stable semantic reference frame to ensure consistency and mitigate scoring drift, improving upon static example-based approaches~\citep{yang2024largelanguagemodelsoptimizers, dong2024surveyincontextlearning}.

We validate \tool through controlled experiments on synthetic data and a case study on a real-world qualitative social survey of student food choices. \ws{The results evaluate both the technical quality of dissimilarity extraction and the practical role of \tool as a pre-analysis aid for qualitative research, rather than as a replacement for full thematic coding.} In summary, our contributions are:
\begin{itemize}[leftmargin=*, labelindent=0pt, itemindent=0pt]
    \item \ws{We propose a four-quadrant quantitative assessment framework for dissimilarity collected from semi-structured questions, based on comparing data-driven response structure with a researcher-defined analytical lens\lp{, whose four metrics are read jointly} \lp{to quantify analysis readiness and also prioritize key responses}.}
    \item We design a content-aware multi-agent system to extract inter-response dissimilarity, featuring a human-in-the-loop iterative process for instruction generation and an adaptive anchor algorithm for evaluation consistency.
    \item \ws{We conduct controlled experiments, including robustness checks, and a real-world study with existing thematic labels, validating \lp{\tool and its components as a pre-analysis assessment}.}
\end{itemize}
\section{Related Work}
\label{sec:related_work}

\textbf{Quality Evaluation of Qualitative Datasets.}
\ws{Qualitative dataset quality is commonly assessed through manual and interpretive practices such as coding, theme development, intercoder reliability checks, saturation, and richness/thickness assessment~\citep{glaser2017discovery, campbell2013coding, ames2024assessing, naeem2024demystification}. These methods are central but occur after data collection and require substantial human effort. Existing automated NLP and LLM-based approaches improve individual-response assessment or qualitative coding support~\citep{LDAfailchang2009reading, chen2023exploringuselargelanguage, dunivin2024scalable, liu2024assessing}, but they provide limited dataset-level pre-analysis signals. \tool addresses this gap by measuring whether an intrinsic response structure aligns with, diverges from, or is orthogonal to a researcher-defined analytical lens.}

\textbf{LLM-based Automated Evaluation.}
\ws{LLM-as-a-Judge enables scalable evaluations~\citep{zheng2023judging, kim2023prometheus}, yet suffers from bias and inconsistency (e.g., position/verbosity bias, prompt sensitivity, and scoring drift)~\citep{ye2024justice, lee2025evaluating}. \tool improves robustness for qualitative response assessment via interactive instruction generation and adaptive anchors that stabilize task-conditioned LLM dissimilarity scoring.}
\section{Proposed Method}

Figure~\ref{fig:overview} shows the overview of \toole.
\ws{The input of \tool is a \textbf{response dataset} (Fig.~\ref{fig:overview}(A)) collecting free-text responses to a question, where each response is associated with metadata that can define a researcher-selected analytical lens.}
\lp{\tool uses two components to assess the dataset's fitness for this lens: (1) a \textbf{content-aware multi-agent system} (Fig.~\ref{fig:overview}(B)--(C)) that enhances the LLM-as-Judge paradigm to compute task-conditioned inter-response dissimilarity, and (2) a \textbf{four-quadrant assessment framework} (Fig.~\ref{fig:overview}(E)) that compares the cluster and metadata-derived distributions reconstructed from this dissimilarity.}
\lp{The pipeline runs in three steps. \tool first generates a task-specific instruction via \textbf{Two-stage Iterative Instruction Generation} (Fig.~\ref{fig:overview}(B), Sec.~\ref{Sec:Instruction_Refinement}), then computes a dissimilarity matrix with it, using \textbf{Adaptive Anchor Manifold Maintenance} (Fig.~\ref{fig:overview}(C), Sec.~\ref{Sec:Anchor_Updating}) to keep stable reference examples for the LLM evaluator. Finally, ensemble clustering of the matrix yields a cluster distribution, compared with the metadata-derived distribution (Fig.~\ref{fig:overview}(D), Sec.~\ref{Sec.data_distributions}) in the \textbf{four-quadrant assessment} (Fig.~\ref{fig:overview}(E), Sec.~\ref{Sec.Assessment_Framework}).}


\subsection{Content-Aware Dissimilarity Extraction}
\label{Sec:intro_system}
\subsubsection{Problem Formulation}
\label{Sec:Problem_Formulation_Dissimilarity}
Given a response dataset $\mathcal{D} = \{ d_i \}_{i=1}^N$ from a question with context $\mathcal{C}_{\text{xt}}$ (e.g., topic, evaluation axis), our goal is to extract a consistent inter-response dissimilarity matrix $\mathbf{D}$. 
\ws{Here, dissimilarity is a task-conditioned semantic difference: for responses $d_i$ and $d_j$, $\delta(d_i,d_j)\in[0,\text{Scale}]$ measures how different the two responses are under the current research question, analytical context $\mathcal{C}_{\text{xt}}$, and evaluation instruction, where $\text{Scale}$ is the rating range of the judge.} \lp{Depending on dataset size, $\delta$ is judged pairwise or derived from per-response scalar scores (Sec.~\ref{Sec:Anchor_Updating}).}
This requires a system that can: (i) generate \lp{a task-adapted} evaluation instruction $P^*$ for $\mathcal{C}_{\text{xt}}$, and (ii) use $P^*$ to compute dissimilarities while maintaining global consistency.

\begin{figure*}[htbp]
    \centering
    \includegraphics[width=1\linewidth]{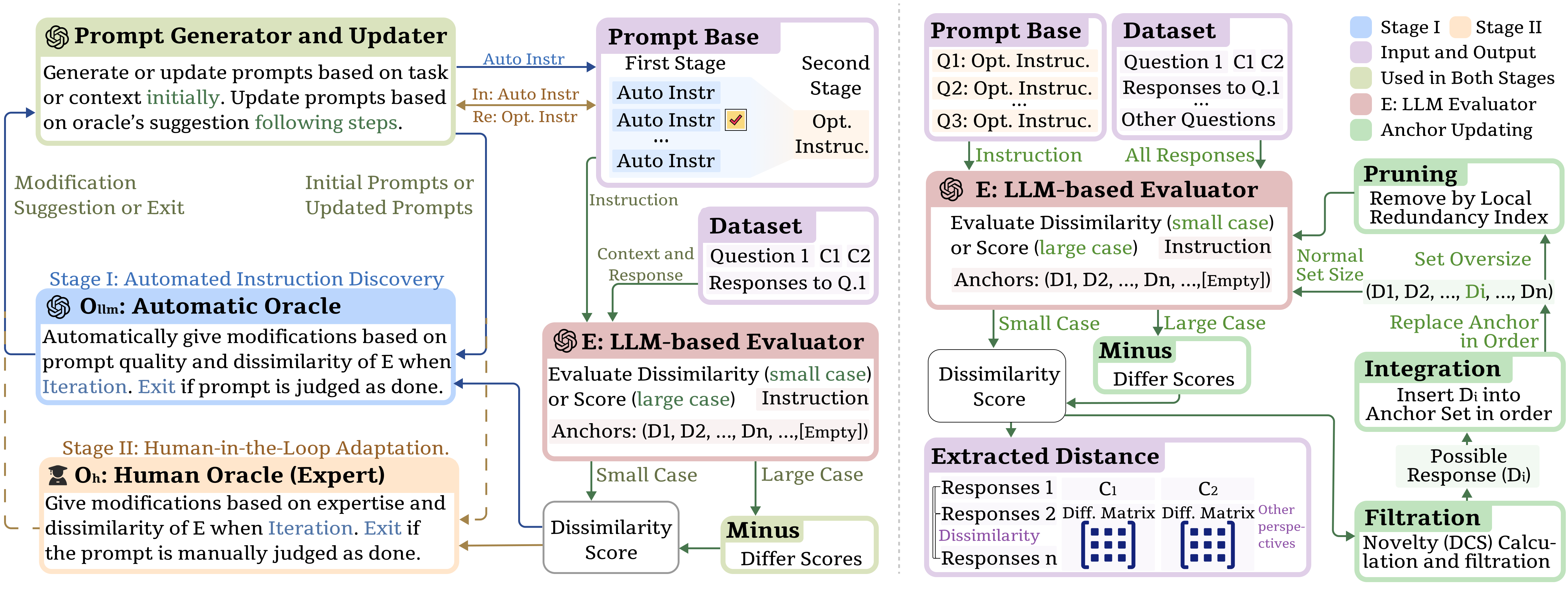}
      \vspace{-10px}
    \caption{The Content-Aware Multi-agent System for Dissimilarity Extraction. It involves two parts: the iterative instruction generation part (Sec.~\ref{Sec:Instruction_Refinement}) and the adaptive anchor manifold updating part (Sec.~\ref{Sec:Anchor_Updating}). The first part efficiently generates context-aware instructions for the second part to perform dissimilarity extraction.}
    \label{fig:Multi_agent_System}
      \vspace{-10px}
\end{figure*}

Let $\mathcal{P}$ be the space of all possible instructions. For any instruction $P \in \mathcal{P}$, an LLM-based evaluator $E$ produces a dissimilarity score $S = E(P, d_i, d_j, \mathcal{C}_{\text{xt}})$. The quality of $P$ is measured by its alignment with a reference judgment from an Oracle $O$ (a powerful LLM or human expert), providing a score $S_O$. The optimization objective is to find the instruction $P^*$ that minimizes the loss:
\eqtight\begin{equation}
\ScaleEq[0.9]{%
P^*=\arg\min_{P\in\mathcal{P}}\;\mathbb{E}_{(d_i,d_j)\sim\mathcal{D}}\!\left[\ell\!\left(E(P,d_i,d_j,\mathcal{C}_{\text{xt}}),\,O(d_i,d_j,\mathcal{C}_{\text{xt}})\right)\right]
}
\end{equation}
where $\ell(\cdot, \cdot)$ is a pseudo-loss function. \lp{This objective states the design goal, which we pursue by refining the instruction iteratively with feedback as a pseudo-gradient, without a guarantee of convergence or of a global optimum} (Sec.~\ref{Sec:Instruction_Refinement}). To enforce consistency, we introduce an anchor manifold $A$, a set of reference responses that is dynamically updated (Sec.~\ref{Sec:Anchor_Updating}). \lp{The system thus has three agent roles (Fig.~\ref{fig:Multi_agent_System}): an Oracle (an LLM in Stage 1, a human expert in Stage 2) returns corrected scores and critiques, an Updater LLM rewrites the instruction from this feedback, and an Evaluator LLM scores responses under the instruction and the current anchors, whose outputs return to the Oracle.}

\subsubsection{Iterative Instruction Generation}
\label{Sec:Instruction_Refinement}
Manually authoring context-specific instructions is labor-intensive and not scalable. We propose a two-stage framework that models instruction optimization as feedback-guided iterative refinement.

At each iteration $t$, an Oracle provides feedback on the current instruction $P^{(t)}$. This feedback, $\hat{\nabla}_{P^{(t)}}\mathcal{L}$, consists of a corrected score and a natural language critique, $(S_{O}^{(t)}, F_{\text{text}}^{(t)})$. An updater LLM, $\text{LLM}_{\text{updater}}$, uses this feedback to generate a refined instruction $P^{(t+1)}$:
\eqtight\begin{equation}
P^{(t+1)} = \text{LLM}_{\text{updater}}(P^{(t)}, \hat{\nabla}_{P^{(t)}}\mathcal{L})
\end{equation}
It treats the natural language critique as a pseudo-gradient, guiding the optimization in the high-dimensional space of instructions\lp{, following the same principle as TextGrad~\citep{yuksekgonul2024textgrad} and ProTeGi~\citep{pryzant2023automatic}}. The magnitude and direction of the ``update'' are determined by the content of the critique. \lp{Unlike these methods, \tool cannot rely on labeled data, which are unavailable at the pre-analysis stage. The feedback therefore comes first from an oracle LLM and then from domain experts, unfolding in two stages:}

\textbf{Stage 1: Automated Instruction Discovery.} It generates diverse, high-quality, general-purpose instructions with a powerful LLM as Oracle ($O \equiv O_{\text{LLM}}$). The process runs automatically, building instructions from scratch until a termination condition is met. This condition is also determined by $O_{\text{LLM}}$, acting as a strict evaluator that assesses the instruction against rigorous criteria for universality, clarity, and quality. The process halts when both the instruction and the evaluator $E$ using it score highly. This stage ends with a user selecting the most promising instruction $P_a^{*}$ from the candidates.

\textbf{Stage 2: Human-in-the-Loop Adaptation.} It specializes $P_a^{*}$ for a specific task with a human expert as Oracle ($O \equiv O_H$), who provides precise, context-aware feedback, guiding the updater LLM to \lp{refine the instruction into} a specialized instruction $P^*$. The loop ends when the expert is satisfied with the evaluator and gives no more input.

\subsubsection{Adaptive Anchor Manifold Maintenance}
\label{Sec:Anchor_Updating}
To ensure consistent dissimilarity measurements, we maintain a sorted anchor manifold $A = \{a_1, \dots, a_k\}$, where each anchor $a_j$ is a (response, score) pair and $k \le k_{\max}$\lp{; the term denotes this set of reference examples, not a geometric manifold}. It gives the LLM a stable semantic reference. \ws{This mechanism is not clustering: it assigns no cluster labels and does not partition the dataset, but maintains reference examples during scoring; clustering is applied afterward to the dissimilarity matrix.} It iteratively processes new responses and updates the manifold to maximize diversity under $k_{\max}$.

\lp{The manifold is initialized with the first response.} The update cycle involves three phases: \textbf{(1) Evaluation:} A new candidate response $d$ is scored against data from $\mathcal{D}$ under anchor guidance, and these scores are compared with the current anchors in $A$ to determine its novelty. \textbf{(2) Filtration:} Candidates that offer the most novelty are selected. Novelty is measured by the Diversity Contribution Score (DCS), defined as the average dissimilarity to existing anchors: $\text{DCS}(d) = \frac{1}{k} \sum_{j=1}^{k} \delta(d, a_j)$. \textbf{(3) Integration and Pruning:} The selected novel candidates are integrated into the manifold. If $|A| > k_{\max}$, the most redundant anchor is pruned. Redundancy is measured by the Local Redundancy Index, \lp{$\rho(a_j) = 1 - \tilde{\delta}(a_j, a_{j-1}) \cdot \tilde{\delta}(a_j, a_{j+1})$ with normalized dissimilarities $\tilde{\delta}=\delta/\text{Scale}$}, flagging anchors interpolated by their neighbors.

The implementation of the evaluation phase adapts to the dataset size $N$, but instructions for both approaches are generated through the same generation process (Sec.~\ref{Sec:Instruction_Refinement})\lp{. $N_{\text{threshold}}$ is set by the computational budget; in our experiments, the 15-response datasets use the pairwise and the 90- and 122-response datasets the scalar regime, an assignment reproduced by any $N_{\text{threshold}}\in[15,90)$}:
\begin{itemize}[leftmargin=*, labelindent=0pt, itemindent=0pt]
\item \textbf{For small datasets ($N \le N_{\text{threshold}}$):} We use a high-fidelity approach where dissimilarities $\delta(d_i, d_j)$ are computed via direct pairwise LLM calls. This has a complexity of $\mathcal{O}(N^2)$ \lp{and suits pilot subsets, where fine-grained relations between responses matter}.

\item \textbf{For large datasets ($N > N_{\text{threshold}}$):} For computational feasibility, we use a linearized $\mathcal{O}(N)$ approach. A computationally efficient LLM operator assigns a scalar score $S(d)$ to each response. All dissimilarities are then approximated as the absolute difference between these scores: $\delta(d_i, d_j) \approx |S(d_i) - S(d_j)|$. \lp{This projection suits criteria with an ordinal reading, such as improvement or decline.}
\end{itemize}
\lp{When the instruction specifies several evaluation conditions, \tool computes $\delta$, clustering, and metrics per condition and averages the metrics. Algorithms~\ref{alg:inte_framework_corrected}--\ref{alg:anchor_maintenance_corrected} in Appendix~\ref{Sec:background_implementation_appendix} give the full procedures.}

\subsection{Data-Distribution-Driven Qualitative Dataset Assessment}
\label{Sec:Framework_Intro}
\subsubsection{Problem Formulation}
\label{Sec.Problem_Formulation_Assessment}
\textbf{Notation and Definitions:} 
We formalize dataset quality assessment as comparing two distributions of the data. Let the dataset be $\mathcal{D} = \{ d_i \}_{i=1}^N$. Each $d_i$ has (i) \ws{an \textbf{extrinsic analytical label} $y_i \in \{1, \dots, C\}$, derived from metadata $\mathcal{I}_i$ via a researcher-defined mapping $y_i = f(\mathcal{I}_i)$, which defines an external analytical partition $\mathcal{P} = \{ C_j \}_{j=1}^C$;} and (ii) an \textbf{intrinsic cluster label} $\hat{y}_i \in \{1, \dots, K\}$, obtained by clustering a dissimilarity matrix $\mathbf{D} \in \mathbb{R}^{N \times N}$, which defines the cluster partition $\hat{\mathcal{P}} = \{ \hat{C}_k \}_{k=1}^K$ and induces the \emph{cluster distribution}.

\begin{figure}[htbp]
    \centering
    \includegraphics[width=\linewidth]{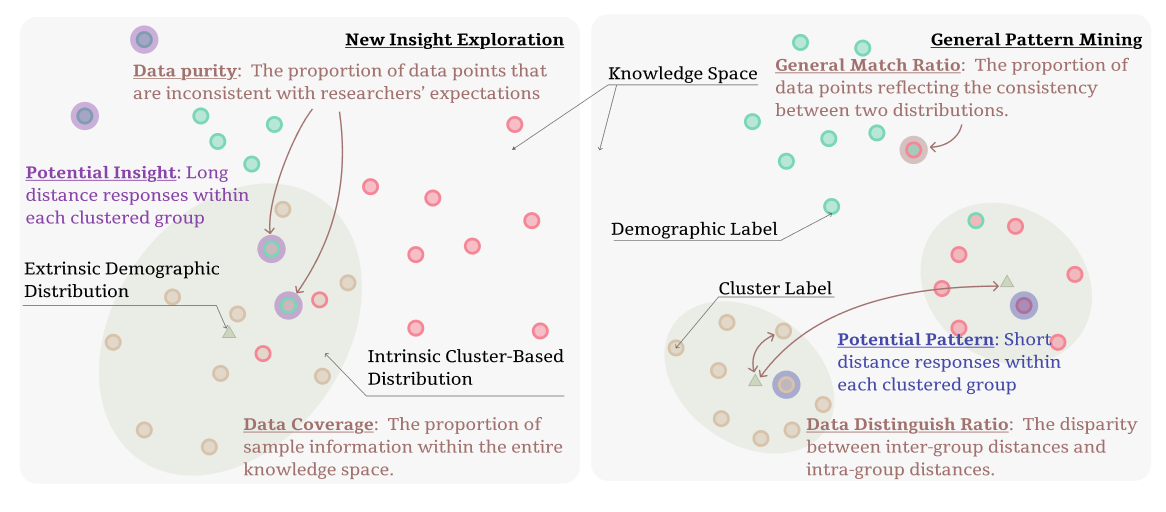}
    \vspace{-10pt}
    \caption{The Four-Quadrant Assessment Framework by research goal (General Patterns/Unique Insights) and analysis level (Distribution/Data Point).}
    \label{fig:Assessment_Framework}
    \vspace{4pt}
\end{figure}

\textbf{Objective:} 
\ws{A researcher may use data for (1) finding common responses that represent groups under a selected analytical lens, \textbf{general patterns}, or (2) discovering previously unknown \textbf{unique insights}. These analyses have two granularities: the \textbf{distribution level} and the \textbf{data-point level}.} \lp{Aligned with these practices, we define dataset quality as \emph{fitness-for-purpose} for qualitative knowledge discovery rather than as a binary good/bad label. The four-quadrant assessment framework (Fig.~\ref{fig:Assessment_Framework}) assigns one normalized metric to each goal--granularity pair\lp{: general-pattern metrics (GMR, DDR) measure how far the lens explains response structure, unique-insight metrics (DP, DC) describe what it leaves within clusters (group mixing and spread), data-point metrics (GMR, DP) count how individual responses are assigned, and distribution metrics (DDR, DC) aggregate continuous dissimilarities}. \lp{The four metrics are read jointly to describe} the concordance and divergence between the external analytical partition $\mathcal{P}$ and the intrinsic cluster partition $\hat{\mathcal{P}}$\lp{, together with the variation within clusters}.}


\subsubsection{The Assessment Framework}
\label{Sec.Assessment_Framework}

\textbf{(1) Demographic and Cluster Distributions}
\label{Sec.data_distributions}

\textbf{Extrinsic Analytical Distribution:} \ws{This distribution is defined \textit{a priori} from metadata on characteristics of interest. A researcher-defined function, $y_i = f(\mathcal{I}_i)$, maps attributes (e.g., expertise level or grade level) to an analytical label for each response $d_i$. The resulting vector $\mathbf{y} = \{y_i\}_{i=1}^N$ represents an \emph{expected structure} along the chosen axis. We treat $\mathcal{P}$ as an external analysis hypothesis rather than ground truth for qualitative insight.}

\textbf{Intrinsic Cluster Distribution:} This distribution is derived \textit{a posteriori} from the data's internal structure. A pairwise dissimilarity matrix $\mathbf{D}$ is \lp{embedded in two dimensions with metric multidimensional scaling and} fed into an ensemble of clustering, with multiple k-means runs and voting~\citep{ahmed2020k}. This yields a vector of cluster assignments $\mathbf{\hat{y}} = \{\hat{y}_i\}_{i=1}^N$. Since cluster labels are arbitrary, we align them with demographic labels via maximum weight bipartite matching, where the weight between a demographic group $C_j$ and a cluster $\hat{C}_k$ is their IOU~\citep{yu2016unitbox}. This alignment makes the distributions comparable. \lp{We set $K=C$ so each analytical group has one matched cluster. The alignment only pairs clusters with groups and does not change which responses belong to which cluster, so a content cluster can still mix analytical groups.}



\textbf{(2) Preliminaries} 
Let $\mathbf{y}$ and $\mathbf{\hat{y}}$ be the demographic and (aligned) cluster assignment vectors. Let $\mathbf{M} \in \mathbb{N}^{C \times C}$ be the confusion matrix where $\mathbf{M}_{jk}$ is the count of data points from demographic $j$ in cluster $k$. $n_j = |C_j|$ and $\hat{n}_k = |\hat{C}_k|$ be the sizes of demographic $j$ and cluster $k$. $\mathbf{D}$ be the dissimilarity matrix. We define: \textbf{(i) Average Intra-Cluster Dissimilarity ($\bar{\delta}_{\text{intra}}$):} The mean dissimilarity between pairs within the same cluster and \textbf{(ii) Average Inter-Cluster Dissimilarity ($\bar{\delta}_{\text{inter}}$):} The mean dissimilarity between pairs in different clusters.

\textbf{(3) Evaluation Metrics.} 
\lp{The four metrics are defined on the two partitions and bounded in $[0,1]$. \lp{No single metric constitutes the assessment: they form one profile, and each is read in the context of the others.} They are not ``higher-is-better'' scores: high GMR/DDR supports group-level pattern extraction, low GMR/DDR may reveal cross-group or lens-orthogonal themes, and high DC with low or moderate DP flags response variation and candidate deviant cases for human interpretation. Tab.~\ref{Tab.Signal_Interpretation} (Appendix~\ref{Sec:qha_discussion}) maps signal combinations to next steps.}

\textbf{General Pattern / Data Point: General Match Ratio (GMR).}
It measures the global alignment between the analytical and cluster distributions. \ws{A high GMR indicates that the intrinsic structure aligns with the selected analytical lens and may support representative responses for each group; a low GMR suggests that the lens weakly explains response structure or that themes cut across groups.}
\eqtight\begin{equation}
\ScaleEq[0.8]{%
\text{GMR}=\max\!\left(0,\;1-\frac{1}{\sqrt{C}}
\left\|\frac{n_j-\mathbf{M}_{jj}}{n_j}+\frac{|n_j-\hat{n}_j|}{n_j}\right\|_2\right)
}
\end{equation}
\lp{GMR combines two failure modes of a lens: \emph{mismatch}, $(n_j-\mathbf{M}_{jj})/n_j$, the share of group $j$ not recovered by its matched cluster, and \emph{size distortion}, $|n_j-\hat{n}_j|/n_j$, which penalizes partitions where one cluster absorbs several groups. The L2 norm scaled by $1/\sqrt{C}$ lets a single poorly recovered group lower GMR more than uniformly small errors. Because GMR counts per-response matches, it indicates whether individual responses can represent their group\lp{; errors are normalized within each group, so groups count equally regardless of size, and ARI and NMI are reported alongside it (Tab.~\ref{Tab.Clustering_Sensitivity})}.}

\textbf{General Pattern / Distribution: Data Distinguishability Ratio (DDR).}
\lp{It measures how much the selected analytical lens separates responses by comparing the average dissimilarity between responses from different groups with that between responses from the same group.} High DDR \lp{indicates that} responses collected from different groups distinctly represent the common feature of that group.
\eqtight\begin{equation}
\text{DDR} = \max\left(0,\; 1-\frac{\bar{\delta}^{\mathcal{P}}_{\text{intra}}}{\bar{\delta}^{\mathcal{P}}_{\text{inter}}}\right)
\end{equation}
\lp{where $\bar{\delta}^{\mathcal{P}}_{\text{intra}}$ and $\bar{\delta}^{\mathcal{P}}_{\text{inter}}$ are the mean dissimilarities in $\mathbf{D}$ over within- and cross-group response pairs under $\mathcal{P}$. DDR is $0$ for indistinguishable lens groups and nears $1$ as cross-group differences dominate.}

\textbf{Unique Insight / Data Point: Data Purity (DP).}
It evaluates cluster consistency as the average share of members in each cluster's dominant analytical group. \ws{Low DP indicates mixed clusters, which may reflect cross-group consensus, underrepresented voices, or themes not captured by the chosen lens.} \lp{DP is the unweighted mean of cluster purities; ARI and NMI are also reported (Tab.~\ref{Tab.Clustering_Sensitivity}).}
\eqtight\begin{equation}
\ScaleEq[0.8]{%
\text{DP}=\max\!\left(0,\;1-\frac{1}{K}\sum_{k=1}^{K}
\frac{|\hat{C}_k|-\max_{j}|\hat{C}_k\cap C_j|}{|\hat{C}_k|}\right)
}
\end{equation}

\textbf{Unique Insight / Distribution: Data Coverage (DC).}
It evaluates the conceptual dispersion of each cluster, quantifying how widely data points spread around the cluster's central theme. A high DC indicates greater intra-cluster dissimilarity relative to a baseline, emphasizing clusters with a broader spread \lp{whose members are candidates for close reading; the spread may also reflect scoring noise}.
\eqtight\begin{equation}
\text{DC} = \min\left(1, \max\left(0, \frac{\bar{\delta}_{\text{intra}}}{\gamma}\right)\right)
\end{equation}
where $\gamma$ is a baseline for acceptable intra-cluster dissimilarity, set relative to the typical dissimilarity unit $\sigma \approx \text{Scale} / C$. We use $\gamma=\lambda\sigma$ with \lp{an empirical} near-1 factor $\lambda$ \lp{adapted} to the estimation regime: $\lambda=1.2$ for small datasets (pairwise scoring, higher-variance estimates) and $\lambda=0.8$ for large datasets (scalar projection, smoother but potentially compressed distances)\lp{; Tab.~\ref{Tab.Quantity_Clarification} summarizes how each quantity in \tool is set}. \lp{Because $\text{Scale}/C$ is the spacing of $C$ groups spread evenly over the score range, DC expresses cluster spread relative to the resolution the lens must resolve. DC does not use the lens labels except through $C$, so it describes response spread rather than lens informativeness and is compared only across lenses with the same $C$.}

\textbf{Response Mining.}
\ws{\lp{\tool also identifies} specific responses for qualitative analysis. These responses, which either challenge expected structures or strongly represent general patterns, are characterized as the top-$k$ responses with the highest and lowest average dissimilarity $\bar{\delta}(d_i)$ \lp{to the other responses in their lens group, marking them as candidate unusual or representative members of that group. A response is thus unusual when it departs from the typical position of its group along the research criterion.} They are intended for response triage before subsequent human interpretation, not as automatic substitutes for thematic coding.}
\section{Experiment}
\label{sec:experiment}

We evaluate \tool{} through controlled experiments on synthetic data (Sec.~\ref{Sec.Exp_Controlled_all}) and a real-world case study (Sec.~\ref{Sec.Exp_case_study}). \ws{We use ``Qwen-max-latest''~\cite{yang2025qwen3} as the default model for its stable API access, strong instruction following, and favorable cost-performance over the many LLM calls of dissimilarity extraction; \tool itself is model-agnostic, and Sec.~\ref{Sec.Exp_Additional_Checks} checks whether conclusions depend on this default.} \lp{The same model drives the synthetic-data generation agents (Appendix~\ref{Sec.Exp_Data_Gen_Sys}); measured token usage, API cost, and results for cheaper model tiers are in Appendix~\ref{Sec:scalability_appendix} (Tabs.~\ref{Tab.API_Cost} and~\ref{Tab.Low_Cost_Models}).}

\subsection{Controlled Experiment}
\label{Sec.Exp_Controlled_all}

Datasets in 4 domains (HCI, Finance, \lp{Additive Manufacturing}, and Medicine) are produced by our Controllable Synthetic Data Generation System (Appendix~\ref{Sec:experimental_materials_appendix}), so ground truth is known for precise evaluation. \lp{These experiments check measurement behavior against known ground truth, like unit tests of an instrument; real-data validation is in Sec.~\ref{Sec.Exp_case_study}.}

\subsubsection{User Study for Instruction Generation}
\label{Sec.Exp_User_Study}

\ws{A controlled within-subjects study assesses only the human-in-the-loop instruction generation module's impact on cognitive load and efficiency, not all of \tool or a full qualitative-analysis interface.}

\textbf{Experimental Setup.} We recruited 20 postgraduate students from diverse domains. Each participant used two systems: (1) our system implementing the adaptation phase described in Sec.~\ref{Sec:Instruction_Refinement}, and (2) a baseline system that employed the same user interface but with agent-assisted features disabled, requiring users to compose instructions manually. Participants wrote dissimilarity-extraction instructions for two distinct questions with both systems. \lp{The counterbalanced schedule is in Tab.~\ref{tab:user_study} and per-item statistical comparisons are in Tab.~\ref{tab:statistical_comparison}.}

\begin{figure}[h]
\centering
\includegraphics[width=\linewidth]{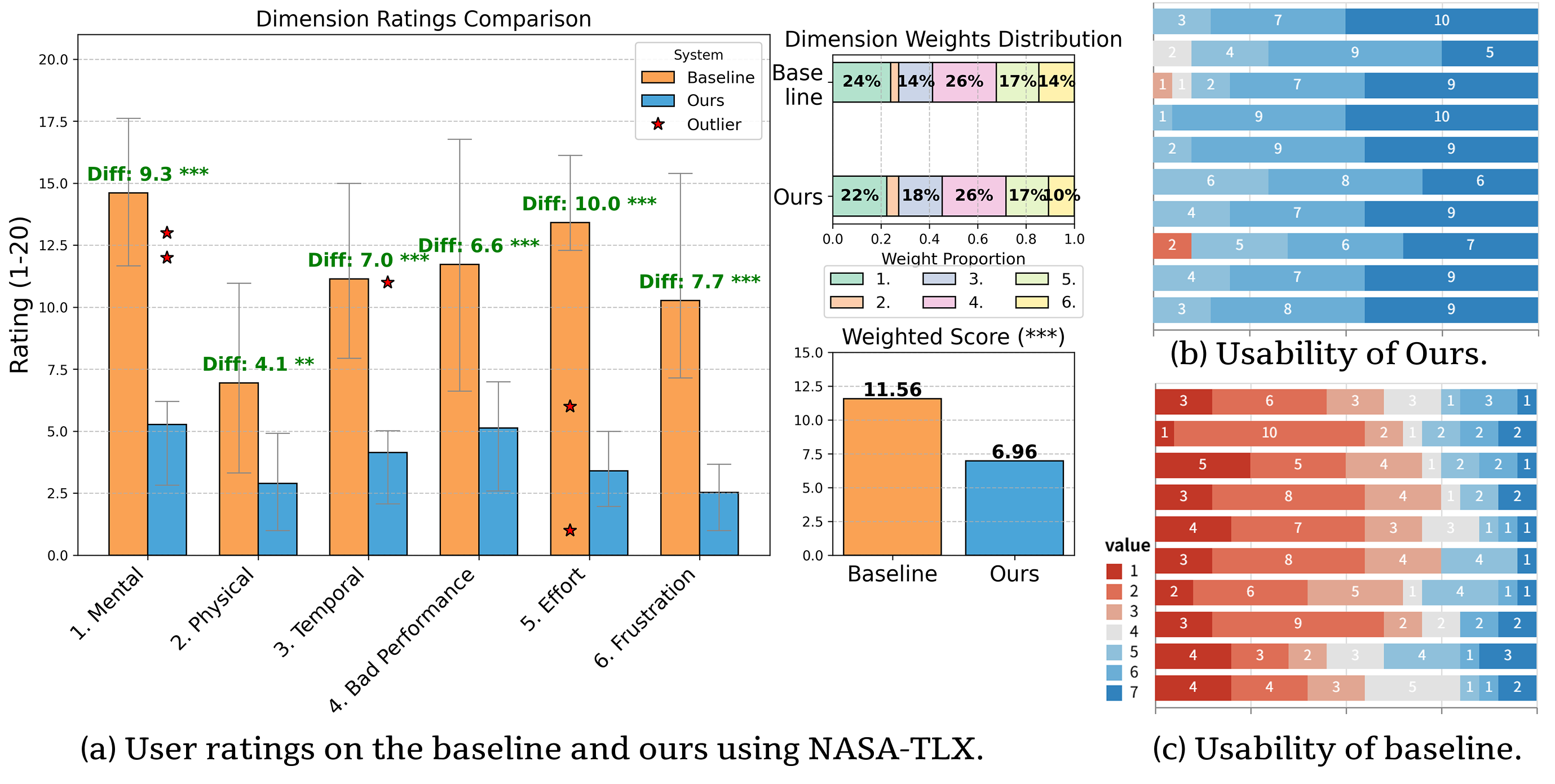}
\caption{
User study results: (a) cognitive load (higher = more load); (b)–(c) \lp{task satisfaction (7 = very satisfied)}.
}
\label{fig.UserStudy}
\end{figure}

\ws{We evaluated both systems on \textit{Cognitive Load} (NASA-TLX)~\citep{hart1988development} and \textit{System Usability} (details in Appendix~\ref{Sec:experimental_materials_appendix}). We use the standard NASA-TLX for completeness.
Usability used a custom 10-item 7-point Likert scale \lp{reporting task satisfaction, not standardized usability}.}

\textbf{Results and Analysis.} As shown in Fig.~\ref{fig.UserStudy}, \lp{participants report lower} cognitive load when composing instructions \lp{with our framework}. \lp{It also receives higher ratings on all satisfaction items.} Instructions from this study were used in later experiments.

\textbf{Quantitative Comparison on Instruction Quality.}
Beyond the user study, we also evaluate instruction quality on synthetic datasets by measuring the correlation between extracted and ground-truth dissimilarity scores.
\lp{The baseline uses the instructions participants directly wrote in the manual condition. Pairwise scores drift in scale across pairs and correlations stay near zero. Scalar projection scores every response on one axis, so direct instructions are already partly calibrated and our gains are smaller.}

\begin{table}[t]
  \centering
  \vspace{-2pt}
  \caption{Quality of instructions generated by baseline vs ours, measured by Spearman and Pearson correlation. \best{Best} indicates the higher value between Baseline and Ours for each metric (within the same setting).}
  \label{Tab.Instruction_Comparison}
  \setlength{\tabcolsep}{2pt}
  \renewcommand{\arraystretch}{0.9}
  \scriptsize
  \resizebox{0.9\columnwidth}{!}{%
  \begin{tabular}{lcccccccc}
  \hline
  & \multicolumn{4}{c}{Small-Scale (Pairwise)} & \multicolumn{4}{c}{Large-Scale (Scalar Projection)} \\
  \cline{2-5}\cline{6-9}
  & \multicolumn{2}{c}{Baseline} & \multicolumn{2}{c}{Ours}
  & \multicolumn{2}{c}{Baseline} & \multicolumn{2}{c}{Ours} \\
  & Spear. & Pears. & Spear. & Pears. & Spear. & Pears. & Spear. & Pears. \\
  \hline
  Qwen\_turbo & -0.083 & -0.035 & \best{0.708} & \best{0.681} & 0.687 & 0.713 & \best{0.747} & \best{0.775} \\
  Qwen\_plus  & 0.228  & 0.292  & \best{0.725} & \best{0.718} & \best{0.735} & \best{0.753} & 0.719 & 0.741 \\
  Qwen\_max   & -0.036 & -0.013 & \best{0.725} & \best{0.728} & 0.705 & 0.726 & \best{0.789} & \best{0.816} \\
  GLM         & 0.599  & 0.621  & \best{0.830} & \best{0.836} & 0.731 & 0.747 & \best{0.797} & \best{0.819} \\
  deepseek-v3 & 0.331  & 0.356  & \best{0.726} & \best{0.705} & 0.770 & 0.795 & \best{0.827} & \best{0.852} \\
  \hline
  \end{tabular}%
  }
  \vspace{-4pt}
\end{table}

\subsubsection{Robustness Checks}
\label{Sec.Exp_Additional_Checks}
\label{Sec.Exp_Abaltion_Study}

\lp{We check robustness at four points of the pipeline: anchor components, dissimilarity source, model family and run, and downstream reconstruction.}

\lp{\textbf{Adaptive-anchor ablation.} On pairwise and scalar-projection synthetic datasets, we disable one of five anchor components at a time and measure correlation with ground truth. \lp{By Spearman correlation, the full configuration ranks first in five of the ten model--regime combinations and within the top two in nine, has the highest mean in both regimes (pairwise $0.743$ vs.\ $\le0.726$; scalar $0.776$ vs.\ $\le0.775$), and has the highest worst case ($0.708$ vs.\ $\le0.659$). Single ablations can win for one model, but none is consistently better, so the full configuration is the most reliable default} (Appendix~\ref{Sec:ablation_appendix}, Tab.~\ref{Tab.Ablation}).}

\lp{\textbf{Dissimilarity baselines.} We compare against two simpler dissimilarity sources. Direct prompting falls well below \tool instructions in the pairwise regime (Tab.~\ref{Tab.Instruction_Comparison}). Embedding distances reach $0.594$--$0.621$ in the pairwise and $0.614$--$0.698$ in the scalar regime, versus $0.708$--$0.830$ and $0.719$--$0.827$ for \tool (Appendices~\ref{Sec:pipeline_necessity_appendix} and~\ref{Sec:additional_baselines_appendix}, Tab.~\ref{Tab.Embedding_Baseline}).}

\lp{\textbf{Cross-model and repeated-run stability.} With DeepSeek and Gemini swapped as generator and evaluator, Spearman correlations with ground truth remain $0.727$--$0.733$ (small) and $0.842$--$0.878$ (large), so, under this latent-score task, the signal does not depend on a same-family generator and evaluator (Tab.~\ref{Tab.Model_Robustness}). Over three repeated runs with fixed inputs, within-model distance correlations are $0.80$ (Qwen-max) and $0.94$ (DeepSeek-v3), cross-model agreement is $0.87$, and the largest SD of the four metrics is $0.050$ (Appendix~\ref{Sec:reliability_appendix}, Tab.~\ref{Tab.Reliability}).}

\lp{\textbf{Reconstruction sensitivity.} Holding the synthetic data, \tool dissimilarity matrix, and $K$ fixed, we vary only the clustering algorithm. ARI ranges from $0.592$ to $0.675$ across K-medoids, agglomerative, spectral, and K-means clustering, with the K-means ensemble strongest overall (ARI/NMI/purity $0.675/0.743/0.850$; Appendix~\ref{Sec:additional_baselines_appendix}, Tab.~\ref{Tab.Clustering_Sensitivity}).}

\subsubsection{Controlled Distribution Shift Experiment}
\label{Sec.Exp_Para_Sweep}

\resrev{We applied \tool{} in both small- and large-scale configurations to synthetic datasets with controlled distributional variations.}
We generated two sets of 15-interviewee datasets (protocol in Appendix~\ref{Sec:experimental_materials_appendix}). In the first set (\texttt{Fixed Mean}), the inter-community mean difference was fixed while variance was swept from 1 to 40. In the second set (\texttt{Fixed Variance}), variance was fixed at 20 while the mean difference was swept from 5 to 45.
\begin{figure}[t]
  \vspace{0pt}
  \centering
  \includegraphics[width=\linewidth]{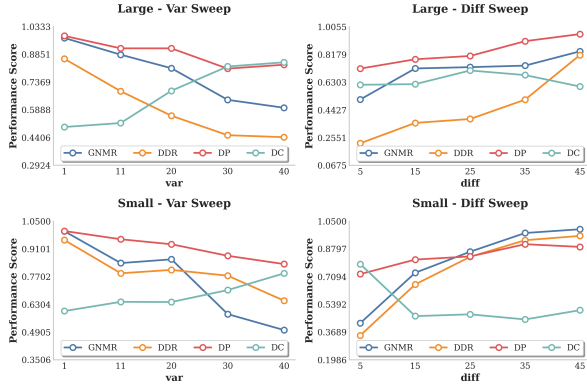}
  \vspace{-12pt}
  \caption{\ws{Result of the distribution shift experiment.}}
  \label{fig.parasweep}
  \vspace{-10pt}
\end{figure}
As shown in Fig.~\ref{fig.parasweep}, with increasing variance, GMR, DP, and DDR decreased while DC increased; with increasing mean difference, GMR, DP, and DDR increased while DC remained stable. These \lp{distinct joint patterns} match our design, validating \toole's scoring functions.




\subsection{Case Study: External Validation on Human-Labeled Food-Choice Responses}
\label{Sec.Exp_case_study}

\lp{This case study externally validates \tool on real, human-annotated qualitative data (details in Appendix~\ref{Sec:case_study_full_appendix}). The Food Choices dataset contains 122 human-written responses to a survey question about eating-habit changes since entering college; grade level is the analytical lens, and \tool runs in the scalar regime. Each response also carries a pre-existing human label of its eating-habit change (worse, better, same, or unclear). The labels were produced independently of \tool and withheld from instruction refinement, model selection, and inference, so they serve as an external criterion; comparisons with label classes exclude the three unclear responses ($N=119$). As the instruction encodes the research question, the instruction and the labels target the same construct. The validation tests whether the lens verdict \tool reaches before coding matches the verdict available after coding, and whether responses sharing a code lie closer.}

\lp{\textbf{Dataset-level validation.} \tool reports $GMR=0.39$, $DDR=0.02$, $DC=0.90$, and $DP=0.43$: grade groups barely align with or separate the reconstructed clusters, responses are widely dispersed within \lp{clusters}, and clusters mix grades. \lp{From this joint profile,} \tool judges grade level a weak lens for eating-habit changes, matching its weak association with human labels (Cramér's $V=0.130$). Switching the lens on the same responses to eating-habit change flips \lp{the profile}: DDR rises from $0.02$ to $0.52$ and DP from $0.43$ to $0.75$, so the verdict tracks the lens (Appendix~\ref{Sec:lens_control_appendix}). \lp{Partly shuffling these labels lowers DDR monotonically to the grade-level value, and \lp{standard association tests on the same scores corroborate the grade verdict} (Appendix~\ref{Sec:code_then_test_appendix}).} A direct end-to-end LLM assessment given the same input instead rates grade level as strong and is sensitive to response order, \lp{unlike \toole's fixed aggregation rules} (Appendix~\ref{Sec:case_study_baselines_appendix}).}

\lp{\textbf{Response-level validation.} Mean \tool distance rises from 14.61 for same-label to 30.38 for different-label pairs. \tool reaches a label-mismatch AUC of 0.749 and top-3 label purity of 0.742, above TF--IDF (0.551, 0.650). Within grades, mined responses match the example in Sec.~\ref{sec:introduction}: representative freshman responses describe eating less and more junk food; unusual ones, more fruits, vegetables, and seafood \lp{(Appendix~\ref{Sec:case_study_mined_responses})}.}

\lp{\textbf{Independent evaluator validation.} After reading the 122 raw responses grouped by grade without the \tool report, three evaluators reviewed the report and rated its correctness, interpretability, and usefulness for later coding on seven-point scales. All median ratings were 6 (correctness 4--7; interpretability and usefulness 6--7; Appendix~\ref{Sec:expert_validation_appendix}).}

\section{Conclusion}
\lp{We introduced \toole, a framework for quantitative pre-analysis assessment of qualitative response datasets. A content-aware multi-agent system with interactive instruction generation and adaptive anchors computes task-conditioned dissimilarities; a four-quadrant framework \lp{jointly reads} how the cluster distribution aligns with or diverges from a researcher-defined analytical lens \lp{and how much variation remains within clusters}, yielding fitness-for-purpose signals for general patterns and unique insights. Controlled experiments and a human-labeled case study show \tool complements qualitative coding by \lp{quantitatively assessing readiness and also prioritizing responses for human check}.}

\section*{\lp{Limitations}}
\label{sec:limitations}

\lp{\textbf{Dependence on the underlying LLM.} \tool inherits the capabilities and stability of its underlying judge. Model updates, decoding settings, and service-side changes can introduce scoring drift; repeated-run and cross-model checks quantify but do not eliminate this source of variation.}

\lp{\textbf{Linguistic and cultural bias.} LLM judgments may privilege standard dialects, dominant cultural framings, verbosity, or particular writing styles. Such preferences can reduce the visibility of culturally specific meanings and encourage homogenized interpretations. Researchers should inspect the generated instruction, anchor examples, representative responses, and outliers, with particular attention to minority-language and culturally situated responses.}

\lp{\textbf{Information loss under scalar projection.} The large-scale regime compresses potentially multidimensional semantic relations onto one task-conditioned axis. This approximation improves efficiency while losing relational detail when dissimilarity is intrinsically multifaceted. \lp{The pairwise regime lets the judge compare two responses directly and can retain relations beyond a single axis when the instruction defines them; its $O(N^2)$ cost makes it best suited to pilot subsets and smaller corpora.} \lp{Our synthetic pairwise tasks use a latent-score gap as ground truth, so they validate pairwise scoring but not relations beyond a single axis.}} \lp{In the scalar regime, \lp{standard association tests on the scores provide a complementary significance check of group separation} (Appendix~\ref{Sec:code_then_test_appendix}).}

\lp{\textbf{Metric interpretation and coverage.} The four metrics describe how a selected lens relates to response structure; whether this structure is qualitatively meaningful also depends on context, theory, and interpretive aims. Reading a single high score as a complete quality judgment risks metric-washing and may downplay meaningful outliers, so \tool reports the metrics as one profile to be read together with the mined responses. Default settings, such as $K=C$, the unweighted DP, and the DC baseline $\gamma$, are listed in Tab.~\ref{Tab.Quantity_Clarification} and can be adapted to a study, for example by adding checks for minority coverage or tail behavior.}

\lp{\textbf{Evidence scope.} Controlled experiments use synthetic responses with known latent scores, complemented by cross-family generation (Tab.~\ref{Tab.Model_Robustness}) and a human-written case study. Real-data validation uses one survey corpus, assessed under eight metadata lenses, the dataset's own human code, and a graded label-noise series between them (Appendix~\ref{Sec:lens_control_appendix}). Since the scoring instruction and the human code address the same research question, this validation tests whether \tool measures the intended construct on real text. Additional corpora and lenses would further test generality.}

\section*{\lp{Ethics Statement}}
\label{sec:ethics}

\lp{\textbf{Human participants.} Both human-subject studies were approved by the institutional review board. All participants gave informed consent, received compensation (USD 8 for the user study and USD 10 for the evaluator study), and are reported only under anonymous identifiers.}

\lp{\textbf{Data and risks.} The Food Choices dataset is publicly available on Kaggle and contains no direct identifiers. Using LLMs to assist qualitative analysis may introduce interpretation biases or over-reliance on automated outputs, and sensitive open-ended responses may raise privacy concerns. We recommend human validation, careful data anonymization, and transparent reporting of model-assisted decisions.}

\lp{\textbf{Use of AI assistants.} Gemini was used to check grammar, improve wording, and search for related work; it did not change the meaning of the text or introduce new references or knowledge, and all retrieved papers were read and reviewed by the authors. Cursor and its associated generative models, such as GPT and Opus, were used to assist with code writing.}

\bibliography{custom}

\appendix
\renewcommand{\topfraction}{0.9}
\renewcommand{\dbltopfraction}{0.9}
\renewcommand{\bottomfraction}{0.6}
\renewcommand{\textfraction}{0.07}
\renewcommand{\floatpagefraction}{0.8}
\renewcommand{\dblfloatpagefraction}{0.8}
\setcounter{topnumber}{3}
\setcounter{dbltopnumber}{3}
\setcounter{bottomnumber}{2}
\setcounter{totalnumber}{5}
\makeatletter
\setlength{\@fptop}{0pt}
\setlength{\@fpbot}{0pt plus 1fil}
\setlength{\@dblfptop}{0pt}
\setlength{\@dblfpbot}{0pt plus 1fil}
\setlength{\@fpsep}{12pt}
\setlength{\@dblfpsep}{12pt}
\makeatother



\section{\lp{Supplementary Evaluation, Robustness, and Deployment}}
\label{Sec:revision_support}

\subsection{\resrev{Ablation Details for Adaptive Anchor Updating}}
\label{Sec:ablation_appendix}
\resrev{This appendix reports the full ablation table \lp{(Tab.~\ref{Tab.Ablation})} summarized in the robustness checks (Sec.~\ref{Sec.Exp_Additional_Checks}). The experiment keeps the data generation and evaluation protocol fixed while disabling individual components of the adaptive anchor updating algorithm (Sec.~\ref{Sec:Anchor_Updating}).}

\lp{\textbf{Experimental setting.} On four 15-interviewee datasets (small scale) and four 90-interviewee datasets (large scale) from different domains (generated via Appendix~\ref{Sec:experimental_materials_appendix}), we evaluate the full method (\texttt{Full}) against variants that each disable one component: disabling in-context examples (\texttt{No Example}), disabling adaptive updating (\texttt{Fixed Example}), including all responses as candidates (\texttt{No Novelty Check}), removing example length limits (\texttt{No Pruning}), and using a heuristic for score calculation (\texttt{Heuristic Quality}), \ie the LLM-assigned anchor score is replaced by the mean dissimilarity of the candidate to existing samples. We report Pearson and Spearman correlations between extracted and ground-truth dissimilarity scores. Owing to the number of experiments, each reported value is the average across four randomly selected questions from the four domains.}

\begin{table*}[t]
\centering
\caption{\resrev{Ablation results of the adaptive anchor updating algorithm. \best{Bold and underlined} represents the top-1 result and \secondbest{bold} represents the top-2 result (within each block).}}
\label{Tab.Ablation}
\setlength{\tabcolsep}{3.5pt}
\renewcommand{\arraystretch}{1.05}
\small
\resizebox{\textwidth}{!}{%
\begin{tabular}{lcccccccccccc}
\toprule
 & \multicolumn{2}{c}{\lp{Full}} & \multicolumn{2}{c}{No Example} & \multicolumn{2}{c}{Fixed Example} & \multicolumn{2}{c}{No Novelty Check} & \multicolumn{2}{c}{No Pruning} & \multicolumn{2}{c}{Heuristic Quality} \\
\cmidrule(lr){2-3}\cmidrule(lr){4-5}\cmidrule(lr){6-7}\cmidrule(lr){8-9}\cmidrule(lr){10-11}\cmidrule(lr){12-13}
Model & Spear. & Pears. & Spear. & Pears. & Spear. & Pears. & Spear. & Pears. & Spear. & Pears. & Spear. & Pears. \\
\midrule
\multicolumn{13}{l}{\textit{Small-scale (pairwise) setting}} \\
\midrule
Qwen\_turbo & \best{0.708} & \best{0.681} & 0.644 & 0.656 & 0.659 & 0.663 & 0.604 & 0.625 & 0.620 & 0.604 & 0.606 & 0.640 \\
Qwen\_plus  & \secondbest{0.725} & \secondbest{0.718} & 0.674 & 0.659 & \best{0.760} & \best{0.757} & 0.700 & 0.683 & 0.652 & 0.667 & 0.643 & 0.636 \\
Qwen\_max   & \best{0.725} & \best{0.728} & 0.633 & 0.621 & 0.672 & 0.663 & 0.683 & 0.695 & 0.691 & 0.697 & 0.701 & 0.709 \\
GLM         & \secondbest{0.830} & \secondbest{0.836} & 0.770 & 0.765 & 0.822 & 0.829 & 0.826 & 0.839 & \best{0.831} & \best{0.842} & 0.729 & 0.732 \\
deepseek-v3 & \secondbest{0.726} & 0.705 & 0.674 & 0.685 & 0.717 & \secondbest{0.722} & 0.697 & 0.673 & 0.703 & 0.676 & \best{0.739} & \best{0.734} \\
\midrule
\multicolumn{13}{l}{\textit{Large-scale (scalar projection) setting}} \\
\midrule
Qwen\_turbo & \best{0.747} & \best{0.775} & 0.705 & 0.733 & 0.692 & 0.721 & 0.699 & 0.730 & 0.729 & 0.765 & 0.709 & 0.732 \\
Qwen\_plus  & 0.719 & 0.741 & \best{0.796} & \best{0.810} & 0.735 & 0.756 & 0.757 & 0.777 & 0.762 & 0.780 & \secondbest{0.767} & \secondbest{0.790} \\
Qwen\_max   & \best{0.789} & \best{0.816} & 0.773 & 0.797 & 0.768 & 0.794 & 0.767 & 0.791 & 0.776 & 0.803 & 0.752 & 0.779 \\
GLM         & \best{0.797} & \best{0.819} & 0.771 & 0.785 & 0.758 & 0.786 & 0.792 & 0.817 & 0.780 & 0.803 & 0.794 & 0.818 \\
deepseek-v3 & \secondbest{0.827} & \secondbest{0.852} & 0.794 & 0.822 & \best{0.856} & \best{0.878} & 0.822 & 0.846 & 0.826 & 0.850 & 0.811 & 0.839 \\
\bottomrule
\end{tabular}%
}%
\end{table*}

\resrev{The complete adaptive-anchor configuration is consistently among the strongest settings and remains robust when moving from pairwise scoring to scalar projection.} \lp{By Spearman correlation, it ranks first in five of the ten model and regime combinations and within the top two in nine, while Heuristic Quality, the most frequent runner-up, reaches the top two four times. \lp{It also has the highest mean Spearman correlation in both regimes (pairwise $0.743$, next Fixed Example $0.726$; scalar $0.776$, next No Pruning $0.775$) and, across the ten settings, the highest minimum ($0.708$, next Fixed Example $0.659$).} Fixed examples or simplified heuristics can perform well for a particular model and scale\lp{, but no ablated variant is consistently better. The complete configuration is therefore the most reliable default across the tested settings; the scalar-regime margin in mean is small, and ranks and means summarize relative performance and are not a significance test.}}

\subsection{Metric and Quantity Clarification}
\label{Sec:metric_parameter_clarification}
\ws{This section clarifies the quantities used by \tool (Tab.~\ref{Tab.Quantity_Clarification}).} \lp{The quantities are fixed by the study design, the LLM scoring rubric, the number of groups, or computational constraints, and are not tuned on validation data. Empirical values enter only where estimation regimes differ in variance: DC uses $\lambda=1.2$ for pairwise scoring, whose estimates vary more, and $\lambda=0.8$ for scalar projection, whose distances are compressed.} \lp{Because $\lambda$ divides DC uniformly, changing it shifts the level of DC but does not reverse its ordering across lenses with the same $C$; e.g., the case-study DC of $0.90$ becomes $0.72$ at $\lambda=1.0$ and $0.60$ at $\lambda=1.2$. $\lambda$ is an empirical normalization convention, not a validated cut-off for reading DC as high or low.}

\begin{table*}[t]
\centering
\caption{\ws{Origin and role of main quantities in \toole.}}
\label{Tab.Quantity_Clarification}
\setlength{\tabcolsep}{3pt}
\renewcommand{\arraystretch}{0.95}
\small
\begin{tabular}{lll}
\toprule
\ws{Quantity} & \ws{Role} & \ws{How it is determined} \\
\midrule
\ws{$C$} & \ws{Number of analytical groups} & \ws{Defined by metadata and research question} \\
\ws{$K$} & \ws{Number of response clusters} & \lp{Set to $C$ so each group has one matched cluster} \\
\ws{$\text{Scale}$} & \ws{LLM score range} & \ws{Fixed by scoring rubric, e.g., $[0,\text{Scale}]$} \\
\ws{$\sigma$} & \ws{Typical dissimilarity unit} & \ws{Derived from $\text{Scale}/C$} \\
\lp{$\gamma$} & \lp{DC normalization} & \ws{Derived from scoring scale and group count} \\
\ws{$k_{\max}$} & \ws{Anchor capacity} & \ws{Set by context-window and cost budget} \\
\ws{$N_{\text{threshold}}$} & \ws{Pairwise/scalar switch} & \lp{Set by cost budget; here pairwise for $N=15$, scalar for $N\ge90$} \\
\ws{top-$k$} & \ws{Response mining display} & \ws{A reporting choice, not a dataset-level metric parameter} \\
\bottomrule
\end{tabular}
\end{table*}


\subsection{\lp{Baseline and Clustering Robustness}}
\label{Sec:additional_baselines_appendix}
\resrev{This appendix provides the full measured results for the additional robustness and baseline checks summarized together in Sec.~\ref{Sec.Exp_Additional_Checks}: embedding-based dissimilarity baselines \lp{(Tab.~\ref{Tab.Embedding_Baseline})}, cross-family generator--evaluator robustness \lp{(Tab.~\ref{Tab.Model_Robustness})}, and fixed-$K$ clustering sensitivity \lp{(Tab.~\ref{Tab.Clustering_Sensitivity})}.} \lp{The interpretation of the embedding and clustering results is given in Appendix~\ref{Sec:pipeline_necessity_appendix}. In the cross-family check, Spearman correlations with ground truth remain $0.727$--$0.733$ in the small-scale and $0.842$--$0.878$ in the large-scale setting in both generator--evaluator directions, so the signal does not depend on a same-family generator and evaluator.}

\begin{table*}[t]
  \centering
  \caption{\resrev{Comparison with embedding-based dissimilarity extraction. Correlations are computed against synthetic ground-truth dissimilarities and averaged across four domains.}}
  \label{Tab.Embedding_Baseline}
  \setlength{\tabcolsep}{8pt}
  \renewcommand{\arraystretch}{1.05}
  \small
  \begin{tabular}{lcccc}
  \toprule
  \resrev{Method} & \resrev{Small Spear.} & \resrev{Small Pears.} & \resrev{Large Spear.} & \resrev{Large Pears.} \\
  \midrule
  \resrev{Qwen3-Embed (cosine)} & 0.595 & 0.602 & 0.614 & 0.625 \\
  \resrev{Qwen3-Embed (euclid)} & 0.595 & 0.593 & 0.614 & 0.619 \\
  \resrev{OpenAI-3-small (cosine)} & 0.594 & 0.602 & 0.698 & 0.717 \\
  \resrev{OpenAI-3-small (euclid)} & 0.594 & 0.594 & 0.698 & 0.709 \\
  \resrev{Llama-Embed-8B (cosine)} & 0.621 & 0.629 & 0.683 & 0.700 \\
  \resrev{Llama-Embed-8B (euclid)} & 0.621 & 0.606 & 0.683 & 0.693 \\
  \bottomrule
  \end{tabular}
\end{table*}

\begin{table*}[t]
  \centering
  \caption{\resrev{Cross-family generator--evaluator robustness check. Communities, candidates, and target scores are held fixed while only response text is regenerated.}}
  \label{Tab.Model_Robustness}
  \setlength{\tabcolsep}{7pt}
  \renewcommand{\arraystretch}{1.05}
  \small
  \begin{tabular}{llcccc}
  \toprule
  \resrev{Generator} & \resrev{Evaluator} & \resrev{Small Spear.} & \resrev{Small Pears.} & \resrev{Large Spear.} & \resrev{Large Pears.} \\
  \midrule
  \resrev{DeepSeek-v4-Flash} & \resrev{Gemini-3.1-Flash-Lite} & 0.733 & 0.754 & 0.842 & 0.855 \\
  \resrev{Gemini-3.1-Flash-Lite} & \resrev{DeepSeek-v4-Flash} & 0.727 & 0.730 & 0.878 & 0.890 \\
  \bottomrule
  \end{tabular}
\end{table*}

\begin{table}[t]
  \centering
  \caption{\resrev{Fixed-$K$ clustering sensitivity and ensemble ablation.}}
  \label{Tab.Clustering_Sensitivity}
  \renewcommand{\arraystretch}{1.05}
  \small
  \setlength{\tabcolsep}{8pt}
  \begin{tabular}{llccc}
  \toprule
  \resrev{Method} & \resrev{Ens.} & \resrev{ARI} & \resrev{NMI} & \resrev{Purity} \\
  \midrule
  \resrev{K-medoids} & \resrev{No} & 0.639 & 0.710 & 0.830 \\
  \resrev{K-medoids} & \resrev{Yes} & 0.631 & 0.698 & 0.831 \\
  \resrev{Agglomerative} & \resrev{No} & 0.592 & 0.679 & 0.786 \\
  \resrev{Agglomerative} & \resrev{Yes} & 0.592 & 0.679 & 0.786 \\
  \resrev{Spectral} & \resrev{No} & 0.607 & 0.686 & 0.818 \\
  \resrev{Spectral} & \resrev{Yes} & 0.608 & 0.686 & 0.818 \\
  \resrev{Current K-means} & \resrev{No} & 0.672 & 0.741 & 0.846 \\
  \resrev{Current K-means} & \resrev{Yes} & 0.675 & 0.743 & 0.850 \\
  \bottomrule
  \end{tabular}
\end{table}

\subsection{\lp{Repeated-Run and Cross-Model Reliability}}
\label{Sec:reliability_appendix}

\lp{\textbf{Question and rationale.} A measurement framework should produce stable response geometry and dataset-level metrics when its inputs and evaluation instruction are held fixed. We therefore test variation across repeated runs and model families.}

\lp{\textbf{Protocol.} On a fixed slice of the controlled synthetic data, qwen-max and deepseek-v3 each perform three independent runs with identical inputs and prompts; clustering randomness is fixed. We report rank agreement between the resulting dissimilarity structures and the mean and standard deviation of the four assessment metrics.}


\begin{table*}[t]
\centering
\caption{\lp{Repeated-run and cross-model reliability. Metric entries are mean$\pm$SD over three runs.}}
\label{Tab.Reliability}
\setlength{\tabcolsep}{7pt}
\renewcommand{\arraystretch}{1.05}
\small
\begin{tabular}{lccccc}
\toprule
\lp{Source} & \lp{Dist. Spear.} & \lp{GMR} & \lp{DDR} & \lp{DC} & \lp{DP} \\
\midrule
\lp{qwen-max} & \lp{0.80} & \lp{$0.839\pm0.050$} & \lp{$0.231\pm0.026$} & \lp{$0.359\pm0.032$} & \lp{$0.898\pm0.043$} \\
\lp{deepseek-v3} & \lp{0.94} & \lp{$0.956\pm0.008$} & \lp{$0.321\pm0.018$} & \lp{$0.304\pm0.019$} & \lp{$0.968\pm0.007$} \\
\lp{Cross-model} & \lp{0.87} & -- & -- & -- & -- \\
\bottomrule
\end{tabular}
\end{table*}

\begin{table*}[!htbp]
\centering
\caption{\lp{Measured API usage for pairwise and scalar inference on the food-choice task.}}
\label{Tab.API_Cost}
\setlength{\tabcolsep}{5pt}
\renewcommand{\arraystretch}{1.05}
\small
\begin{tabular}{lccccc}
\toprule
 & \multicolumn{3}{c}{\lp{Pairwise}} & \multicolumn{2}{c}{\lp{Scalar projection}} \\
\cmidrule(lr){2-4}\cmidrule(lr){5-6}
\lp{Metric} & \lp{$N=10$} & \lp{$N=15$} & \lp{$N=30$} & \lp{$N=30$} & \lp{$N=122$} \\
\midrule
\lp{Total tokens} & \lp{55,108} & \lp{115,345} & \lp{449,765} & \lp{13,391} & \lp{56,727} \\
\lp{Input tokens} & \lp{46,781} & \lp{98,231} & \lp{384,725} & \lp{11,088} & \lp{47,707} \\
\lp{Output tokens} & \lp{8,327} & \lp{17,114} & \lp{65,040} & \lp{2,303} & \lp{9,020} \\
\lp{Estimated cost (USD)} & \lp{\$0.0057} & \lp{\$0.0118} & \lp{\$0.0453} & \lp{\$0.0015} & \lp{\$0.0060} \\
\bottomrule
\end{tabular}
\end{table*}

\begin{table}[!htbp]
\centering
\caption{\lp{Correlation across Qwen model tiers.}}
\label{Tab.Low_Cost_Models}
\setlength{\tabcolsep}{4pt}
\renewcommand{\arraystretch}{1.05}
\small
\begin{tabular}{lcc}
\toprule
\lp{Model} & \lp{Pairwise Spear./Pears.} & \lp{Scalar Spear./Pears.} \\
\midrule
\lp{Qwen-turbo} & \lp{0.708/0.681} & \lp{0.747/0.775} \\
\lp{Qwen-plus} & \lp{0.725/0.718} & \lp{0.719/0.741} \\
\lp{Qwen-max} & \lp{0.725/0.728} & \lp{0.789/0.816} \\
\bottomrule
\end{tabular}
\end{table}

\lp{\textbf{Results.} In Tab.~\ref{Tab.Reliability}, within-model distance correlations are 0.80 and 0.94, and cross-model agreement is 0.87. The largest observed metric standard deviation is 0.050. These results show moderate-to-high rank agreement and limited metric variation in the tested slice. \lp{Metric levels differ between the two models, so these results support repeatability within a model, not interchangeability of absolute metric values across models.}}





\subsection{\lp{Pipeline Component Necessity Checks}}
\label{Sec:pipeline_necessity_appendix}

\lp{\textbf{Question and rationale.} We evaluate each major pipeline stage using a comparison matched to its function \lp{to examine the contribution of each component}. These checks separately examine task-specific instruction generation, adaptive-anchor maintenance, semantic representation, and downstream distribution reconstruction. A final real-label comparison tests whether the resulting response geometry transfers beyond synthetic ground truth.}

\lp{\textbf{Instruction generation and adaptive anchors.} Tab.~\ref{Tab.Instruction_Comparison} compares direct prompting with \toole-generated instructions using correlation with synthetic ground-truth dissimilarities. In the pairwise setting, Spearman correlation increases from $-0.036$ to $0.725$ for Qwen-max, from $0.599$ to $0.830$ for GLM, and from $0.331$ to $0.726$ for DeepSeek-v3. Tab.~\ref{Tab.Ablation} then removes in-context examples, adaptive updating, novelty filtering, pruning, or \lp{LLM-assigned anchor scores (\texttt{Heuristic Quality})} individually. The complete configuration remains among the strongest settings across models and inference regimes, while no single ablated variant consistently dominates. These results support interpreting the components as a combined mechanism for producing stable task-conditioned judgments.}

\lp{\textbf{Semantic representation.} Tab.~\ref{Tab.Embedding_Baseline} replaces task-conditioned LLM judgments with cosine and Euclidean distances from Qwen3-Embed, OpenAI-3-small, and Llama-Embed-8B. Their Spearman correlations range from $0.594$ to $0.621$ in the pairwise setting and from $0.614$ to $0.698$ under scalar projection. The corresponding task-conditioned \tool results range from $0.708$ to $0.830$ and from $0.719$ to $0.827$, respectively. The observed differences indicate that the researcher-defined evaluation instruction captures criterion-specific distinctions beyond generic semantic similarity.}

\lp{\textbf{Downstream reconstruction.} Tab.~\ref{Tab.Clustering_Sensitivity} holds the synthetic data, \tool dissimilarity matrix, and researcher-specified $K$ fixed while varying the clustering algorithm and ensemble setting. Across K-medoids, agglomerative, spectral, and K-means clustering, ARI ranges from $0.592$ to $0.675$, NMI from $0.679$ to $0.743$, and purity from $0.786$ to $0.850$. The current K-means ensemble is strongest overall ($0.675/0.743/0.850$), while all alternatives recover comparable ground-truth structure. This check separates the robustness of the extracted dissimilarity signal from the choice of downstream reconstruction algorithm.} \lp{It holds $K$ fixed and therefore does not vary the cluster granularity. On the case-study matrix, the two-dimensional MDS embedding preserves the distances closely (Kruskal stress-1 $0.046$; Pearson $r=0.997$ between input and embedded distances), as expected for one-dimensional scalar distances; pairwise matrices can be distorted more.}

\lp{\textbf{External validation against human-assigned labels.} Tab.~\ref{Tab.External_Baselines} evaluates random retrieval, TF--IDF, and \tool using the 119 food-choice responses with non-unclear human labels. Their label-mismatch AUCs are $0.500$, $0.551$, and $0.749$; top-3 purities are $0.489$, $0.650$, and $0.742$; and top-5 purities are $0.489$, $0.619$, and $0.694$, respectively. The lexical baseline captures partial thematic structure, while \tool achieves higher observed agreement with the external labels across both pair-level and neighborhood-level measures.} \lp{Standard association tests on the same \tool scores are compared with DDR\lp{, the group-separation component of the profile,} in Appendix~\ref{Sec:code_then_test_appendix}.}

\lp{Together, these comparisons provide stage-specific evidence: generated instructions and adaptive anchors support task-conditioned dissimilarity extraction, generic embeddings provide a weaker criterion-specific signal, reconstruction remains robust across fixed-$K$ clustering choices, and the resulting response geometry transfers to an independently labeled real-world dataset.}

\subsection{\lp{Scalability and Deployment Cost}}
\label{Sec:scalability_appendix}

\lp{\textbf{Question and rationale.} Deployment depends on both inference complexity and model cost. We quantify the transition from pairwise judgment, which requires $O(N^2)$ LLM comparisons, to scalar projection, which scores each response once with $O(N)$ LLM calls and computes pairwise differences locally.}

\lp{\textbf{Protocol.} We instrument the Qwen-Flash API on the food-choice task using a 4,968-character instruction and batches of five responses. Token counts are returned by the API. Cost estimates apply \$0.05 per million input tokens and \$0.40 per million output tokens.}

\lp{\textbf{Accounting boundary.} Tab.~\ref{Tab.API_Cost} covers dissimilarity-scoring API calls. It excludes instruction generation and refinement, failed requests or retries, and local clustering and metric computation. The values therefore characterize the scoring stage rather than total end-to-end deployment cost.}

\lp{\textbf{Results.} As shown in Tab.~\ref{Tab.API_Cost}, scalar projection reduces the 30-response scoring run from 449,765 to 13,391 tokens and processes all 122 responses with 56,727 tokens at an estimated cost of \$0.0060. Across Qwen model tiers (Tab.~\ref{Tab.Low_Cost_Models}), Qwen-turbo obtains correlations of 0.747/0.775 in the scalar setting, demonstrating compatibility with a smaller model tier. API prices vary over time; the reported values document the measured scoring workload and pricing assumptions.}

\AppendixBeforeBox
\section{\lp{Real-World Case Study and External Validation}}
\label{Sec:case_study_full_appendix}

\subsection{\lp{Case Study Setup}}
\lp{We study 122 human-written responses to ``Describe your eating changes since the moment you got into college'' from the Food Choices dataset\footnote{\url{https://www.kaggle.com/datasets/borapajo/food-choices}}. Grade level (37 freshmen, 32 sophomores, 26 juniors, and 27 seniors) defines the researcher-selected analytical lens. Pre-existing human-coded labels describe eating habits as getting worse, getting better, staying the same, or unclear and serve as an external criterion. All 122 responses enter the grade--label association; pair and retrieval analyses exclude the three unclear responses and use $N=119$. The labels are withheld from instruction refinement, model selection, and \tool inference and are introduced only for post-hoc validation.}

\lp{\tool reports $GMR=0.39$, $DDR=0.02$, $DC=0.90$, and $DP=0.43$. These values indicate limited alignment between grade and reconstructed clusters, \lp{grade groups that are barely distinguishable, high within-cluster response dispersion}, and substantial grade mixing within clusters, respectively. \lp{Because $N=122$ exceeds $N_{\text{threshold}}$, dissimilarities are computed in the scalar regime.} The grade-level label distribution in Fig.~\ref{fig:distribution_gradelevel} and the external checks below evaluate the lens-level interpretation.}

\subsection{\ws{Pipeline Walkthrough}}
\label{Sec:running_example}
\ws{This running example follows the stages in Fig.~\ref{fig:overview}.}

\textbf{\ws{A. Response Dataset.}}
\ws{The input contains the 122 responses and grade-level metadata described above. Grade level is the selected analytical lens.}

\textbf{\ws{B. Instruction Generation.}}
\ws{The instruction-generation module refines an evaluation instruction for eating-habit-change dissimilarity, focusing comparison on the research question rather than generic lexical similarity.}

\textbf{\ws{C. Dissimilarity Extraction.}}
\ws{The LLM judge, stabilized by adaptive anchors, estimates task-conditioned differences between responses. For example, late-night snacking and fewer full meals are closer to convenience-driven unhealthy eating than to athlete-related dietary discipline.}

\textbf{\ws{D. Distribution Reconstruction.}}
\ws{Clustering the response--response dissimilarity matrix produces intrinsic response groups, which are aligned with the grade-level partition.}

\textbf{\ws{E. Assessment and Response Mining.}}
\ws{The resulting metrics assess the grade-level lens, while representative and unusual responses prioritize typical and deviant cases for subsequent coding.}

\subsection{\lp{External Validation against Human-Coded Labels}}
\label{Sec:external_validation_appendix}
\lp{\textbf{Purpose and external criterion.} The controlled experiments evaluate \tool against synthetic ground truth, whereas validation on real qualitative data requires information produced independently of the framework. The Food Choices dataset provides pre-existing human-assigned labels describing each response as getting worse, getting better, staying the same, or unclear. These labels were withheld from instruction refinement, model selection, and \tool inference and were introduced only after the system outputs had been produced. They therefore provide an external criterion for evaluating whether \toole's assessment and response geometry correspond to human-coded thematic structure.}

\lp{\textbf{Validation logic.} We derive three testable predictions from the system outputs. First, \tool reports weak alignment between grade level and the reconstructed response structure. If this lens-level assessment is meaningful, grade should also show weak association with the independently assigned thematic labels. Second, if \tool distances capture task-conditioned thematic differences, responses with different human labels should be farther apart than responses sharing the same label. Third, if the resulting response geometry supports coding-oriented triage, nearest-neighbor retrieval should produce more label-consistent neighborhoods than random retrieval and a lexical TF--IDF baseline. These predictions evaluate the dataset-level conclusion, pair-level distances, and retrieval behavior using the same independent criterion.} \lp{Because the instruction and the labels target the same construct, these checks test criterion validity, i.e., whether the instruction-conditioned measurement tracks the intended construct on real text; they do not test theme discovery. TF--IDF and generic embeddings do not see the criterion, and the direct LLM sees the research question but not the refined rubric. These comparisons therefore show the value of criterion-conditioned scoring over generic similarity; standard tests on the same \tool scores are reported in Appendix~\ref{Sec:code_then_test_appendix}.}
\lp{\textbf{Operationalization.} For the lens-level check, Cramér's $V$ summarizes the association between grade and thematic label over all 122 responses. For the pair-level check, the three unclear responses are excluded, leaving $\binom{119}{2}=7{,}021$ pairs; we compare the mean and median distances of same- and different-label pairs\lp{, which share responses and are not treated as independent observations}. For the retrieval check, label-mismatch AUC treats a different-label pair as the positive class and response distance as the prediction score. Top-$k$ purity measures the macro-averaged proportion of same-label responses among each query's $k$ nearest neighbors, excluding the query itself.}

\begin{table*}[t]
\centering
\caption{\lp{Association and pair-level validation against human-coded labels.}}
\label{Tab.External_Criterion}
\setlength{\tabcolsep}{5pt}
\renewcommand{\arraystretch}{1.05}
\small
\begin{tabular}{lll}
\toprule
\lp{Analysis} & \lp{Quantity} & \lp{Result} \\
\midrule
\lp{Grade--label association ($N=122$)} & \lp{Cramér's $V$} & \lp{0.130} \\
\lp{Same-label pairs ($n=3{,}433$)} & \lp{Mean / median distance} & \lp{14.61 / 10.00} \\
\lp{Different-label pairs ($n=3{,}588$)} & \lp{Mean / median distance} & \lp{30.38 / 30.00} \\
\lp{Pair-distance contrast} & \lp{Mean gap} & \lp{15.77} \\
\bottomrule
\end{tabular}
\end{table*}

\begin{table*}[t]
\centering
\caption{\lp{Real-label comparison of response-distance and retrieval baselines ($N=119$). Fine-code AUC uses the dataset's 13-category theme code (\texttt{eating\_changes\_coded1}), which is not the scoring criterion of the \tool instruction. $^\dagger$Direct LLM yields a partition rather than a distance, so neighbour purity is undefined; values are means over three shuffled-order runs (AUC SD $=0.077$).} \lp{All entries are descriptive point estimates.} \lp{The upper block does not see the research question; the lower block does.}}
\label{Tab.External_Baselines}
\setlength{\tabcolsep}{10pt}
\renewcommand{\arraystretch}{1.05}
\small
\begin{tabular}{lcccc}
\toprule
\lp{Method} & \lp{Mismatch AUC} & \lp{Top-3 pur.} & \lp{Top-5 pur.} & \lp{Fine-code AUC} \\
\midrule
\lp{Random} & \lp{0.500} & \lp{0.489} & \lp{0.489} & \lp{0.500} \\
\lp{TF--IDF} & \lp{0.551} & \lp{0.650} & \lp{0.619} & \lp{0.555} \\
\lp{Qwen3-Embed (cosine)} & \lp{0.604} & \lp{\best{0.748}} & \lp{\best{0.733}} & \lp{0.587} \\
\midrule
\lp{Direct LLM$^\dagger$} & \lp{0.613} & \lp{--} & \lp{--} & \lp{0.602} \\
\lp{\tool} & \lp{\best{0.749}} & \lp{0.742} & \lp{0.694} & \lp{\best{0.629}} \\
\bottomrule
\end{tabular}
\end{table*}

\lp{\textbf{Results and interpretation.} All three predicted patterns are observed. Tab.~\ref{Tab.External_Criterion} reports the association and pair-level results. Grade and thematic label show weak association ($V=0.130$), consistent with \toole's conclusion that grade is a weak analytical lens. Different-label pairs have a mean distance of 30.38, compared with 14.61 for same-label pairs, yielding a gap of 15.77. Tab.~\ref{Tab.External_Baselines} reports the retrieval results. \tool achieves a label-mismatch AUC of 0.749 and top-3/top-5 purities of 0.742/0.694, compared with 0.551 and 0.650/0.619 for TF--IDF and 0.500 and 0.489/0.489 for random retrieval; the Qwen3-Embed and Direct LLM rows and the fine-code AUC are discussed in Appendix~\ref{Sec:case_study_baselines_appendix}. \lp{\tool is highest on both AUCs, whereas Qwen3-Embed has slightly higher neighbour purity, so the advantage of \tool lies in criterion-specific discrimination rather than in generic neighbourhood structure. Agreement with the fine code is lower (AUC 0.629), as expected when responses with the same score reflect different mechanisms, such as cost or time pressure, which a single scoring axis does not separate.}
\lp{Together, these results show that the instruction-conditioned measurement tracks the intended construct on real text, and that \toole's lens-level conclusion agrees with the independent labels; they do not show theme discovery beyond the instruction.} The independent-evaluator study complements these label-based checks by examining agreement with qualitative judgments and perceived usefulness for subsequent coding.}

\subsection{\lp{Weak versus Strong Lens on the Same Responses}}
\label{Sec:lens_control_appendix}

\lp{\textbf{Question and rationale.} The case study reports one lens, grade level, which \tool judges weak. A weak verdict is informative only if the same instrument gives a different reading when the lens is strong. We therefore keep the responses and the \tool dissimilarity matrix fixed and switch the lens from grade level to eating-habit change, i.e., the dataset's own human code (worse / better / same; $N=119$, unclear excluded). By construction, this lens follows the content of the responses, so it serves as a known-strong reference. In the scalar regime the dissimilarity matrix does not depend on the lens, so no additional LLM calls are needed. The four metrics are recomputed with $K=C$ and averaged over 20 reconstruction runs.} \lp{Because the two lenses also differ in sample size and group count (122/4 vs.\ 119/3), DDR, which does not depend on the reconstruction, is the most directly comparable metric\lp{, whereas DC is normalized by $C$ and is not compared across the two lenses}; the graded series below holds $N$ and $C$ fixed.}

\begin{table*}[t]
\centering
\caption{\lp{\tool assessment of the same responses under a weak and a strong analytical lens. Grade-level values are those reported in Sec.~\ref{Sec.Exp_case_study} (all 122 responses). The eating-habit-change lens uses the 119 responses with a non-unclear code, since unclear is not a substantive group. Eating-habit-change values are means over 20 reconstruction runs (SD $0.06$ for GMR and DP, $0$ for DDR and DC).}}
\label{Tab.Lens_Control}
\setlength{\tabcolsep}{8pt}
\renewcommand{\arraystretch}{1.05}
\small
\begin{tabular}{lccccl}
\toprule
\lp{Lens} & \lp{GMR} & \lp{DDR} & \lp{DC} & \lp{DP} & \lp{Reading (Tab.~\ref{Tab.Signal_Interpretation})} \\
\midrule
\lp{Grade level} & \lp{0.39} & \lp{0.02} & \lp{0.90} & \lp{0.43} & \lp{Weak lens; themes cut across groups} \\
\lp{Eating-habit change} & \lp{0.53} & \lp{\best{0.52}} & \lp{0.72} & \lp{\best{0.75}} & \lp{Lens explains response structure} \\
\bottomrule
\end{tabular}
\end{table*}

\lp{\textbf{Results.} Switching the lens changes the reading in the expected direction (Tab.~\ref{Tab.Lens_Control}). The clearest change is in group separability: DDR rises from $0.02$, where grade groups are barely distinguishable, to $0.52$, where between-group dissimilarity is about twice the within-group dissimilarity. DP rises from $0.43$ to $0.75$, so the reconstructed clusters move from mixed grades to being dominated by a single eating-change group. \lp{DC is lower under eating-habit change ($0.72$ vs.\ $0.90$), but its normalizer depends on $C$, which differs between the two lenses, so we do not attribute this difference to the lens.} GMR increases moderately, from $0.39$ to $0.53$. Under grade level, the signals match the second row of Tab.~\ref{Tab.Signal_Interpretation} (low GMR/DP/DDR, high DC: revise the lens or code cross-group themes). Under eating-habit change, they match the first row (high GMR/DP/DDR: compare groups and read representative responses first). Both readings agree with the external criterion: grade level is barely associated with the human code ($V=0.130$), whereas the eating-change lens is the human code itself. The weak verdict in the case study therefore reflects the lens, not an instrument that reports every lens as weak.}

\lp{\textbf{Graded lens informativeness.} The eating-habit-change lens is the human code itself, so it is an upper endpoint rather than evidence that \tool separates informative from uninformative lenses. To cover intermediate cases, we keep this lens but randomly reassign the group labels of a fraction $p$ of responses (20 seeds per $p$), which lowers the information the lens carries about response content from full ($p=0$) to none ($p=1$). The dissimilarity matrix is unchanged, so no LLM calls are needed; metrics use the formulas of Sec.~\ref{Sec:Framework_Intro} with $K=C$. Tab.~\ref{Tab.Noise_Ladder} shows that DDR decreases monotonically from $0.52$ to $0.02$, the grade-level value, and follows the same course as the Kruskal--Wallis effect size $\varepsilon^2$ on the same scores. DP decreases until it reaches the level of the majority-group share ($0.63$)\lp{, a descriptive reference rather than an exact chance level; the fully permuted lens ($p=1$, $0.64$) gives the permutation reference for DP, which the unpermuted lens ($0.76$) exceeds}. \lp{DC stays constant across $p$, since the clusters and $C$ are fixed and DC does not use the lens labels.} \lp{Because this perturbation changes only the lens--content association, it isolates the lens-dependent readings;} \tool thus tracks how informative a lens is rather than only its two endpoints; eight further metadata lenses are compared with standard tests in Appendix~\ref{Sec:code_then_test_appendix}.}

\begin{table}[t]
\centering
\caption{\lp{Graded lens informativeness: a fraction $p$ of eating-habit-change labels is randomly reassigned ($N=119$; means over 20 seeds). KW $\varepsilon^2$ is the Kruskal--Wallis effect size of \tool scores across the noisy groups.}}
\label{Tab.Noise_Ladder}
\setlength{\tabcolsep}{4pt}
\renewcommand{\arraystretch}{1.05}
\small
\begin{tabular}{lcccccc}
\toprule
\lp{$p$} & \lp{0.0} & \lp{0.2} & \lp{0.4} & \lp{0.6} & \lp{0.8} & \lp{1.0} \\
\midrule
\lp{DDR} & \lp{0.52} & \lp{0.39} & \lp{0.28} & \lp{0.13} & \lp{0.07} & \lp{0.02} \\
\lp{DP} & \lp{0.76} & \lp{0.70} & \lp{0.66} & \lp{0.61} & \lp{0.61} & \lp{0.64} \\
\lp{KW $\varepsilon^2$} & \lp{0.45} & \lp{0.27} & \lp{0.15} & \lp{0.08} & \lp{0.02} & \lp{0.01} \\
\bottomrule
\end{tabular}
\end{table}

\subsection{\lp{Direct LLM Assessment and Representation Baselines}}
\label{Sec:case_study_baselines_appendix}

\lp{\textbf{Question and rationale.} A natural alternative to \tool is to give an LLM the same input and ask for the same output directly. We test whether this end-to-end shortcut matches \toole's lens verdict, response triage, and response structure.}

\lp{\textbf{Protocol.} We use nine lenses: the eating-habit-change lens of Appendix~\ref{Sec:lens_control_appendix} and eight survey variables (grade level, gender, living situation, current diet, self-rated health, sports, cooking frequency, and self-perceived weight; codebook categories, groups with fewer than five responses dropped). Seven of the eight have no significant association with the human code ($\chi^2$ $p\ge0.06$), and current diet has a moderate one ($V=0.25$). For each lens, qwen-max (temperature $0$, JSON mode) receives the research question and all responses with their IDs and lens groups. It returns (i) a 1--7 lens-usefulness rating and verdict, (ii) three representative and three unusual responses per group, and (iii) a content-only partition of all responses into $K=C$ clusters. Each lens is run three times with a different response order (seeds fixed). Outputs failing schema checks are re-requested up to twice; format-only normalisation (e.g., ID lists returned as strings) is applied, and residual errors are counted as misses (38 out-of-group triage IDs and 4 unassigned responses over 26 runs). One of the 27 runs (current diet) failed after retries and is excluded. Response triage is scored with a label-based proxy: a representative pick is a hit if its human code equals the group's majority code, and an unusual pick is a hit if it does not. The \tool and Qwen3-Embed picks use the lowest and highest within-group mean distances. \lp{For all methods, the four non-substantive answers (two ``none'', one ``Nun'', one ``None really'') are excluded from the picks and from the random baseline, as in the mined-response lists (Appendix~\ref{Sec:case_study_mined_responses}).} Partitions are compared with both human codes by ARI/NMI.}

\begin{table*}[t]
\centering
\caption{\lp{Direct LLM versus \tool on the eight metadata lenses (means over lenses; LLM values also averaged over three shuffled-order runs). Triage hit rates use the human code as a proxy. Partition agreement is ARI with the coarse / fine human codes.}}
\label{Tab.Direct_LLM}
\setlength{\tabcolsep}{8pt}
\renewcommand{\arraystretch}{1.05}
\small
\begin{tabular}{lcccc}
\toprule
\lp{Measure} & \lp{Random} & \lp{Qwen3-Embed} & \lp{Direct LLM} & \lp{\tool} \\
\midrule
\lp{Representative hit rate} & \lp{0.66} & \lp{0.80} & \lp{0.63} & \lp{\best{0.81}} \\
\lp{Unusual hit rate} & \lp{0.34} & \lp{0.29} & \lp{0.47} & \lp{\best{0.82}} \\
\lp{Partition ARI (coarse / fine)} & \lp{0.00 / 0.00} & \lp{0.01 / 0.00} & \lp{0.11 / 0.08} & \lp{\best{0.30} / \best{0.11}} \\
\lp{Grade level judged ``strong''} & -- & -- & \lp{3/3 runs} & \lp{No (joint profile)} \\
\lp{Top-3 overlap across reorderings (Jaccard)} & -- & -- & \lp{0.12} & \lp{1.00$^\S$} \\
\lp{Partition agreement across reorderings (ARI)} & -- & -- & \lp{0.37} & \lp{--} \\
\bottomrule
\end{tabular}
\par\vspace{2pt}{\footnotesize\lp{$^\S$Picks are computed from the fixed distance matrix; repeated-run reliability of the scoring is in Tab.~\ref{Tab.Reliability}\lp{; the order of anchor updating is not varied}.}}
\end{table*}

\lp{\textbf{Results.} Tab.~\ref{Tab.Direct_LLM} summarizes the comparison. The direct LLM does not provide a reliable pre-analysis signal. (i) \emph{Lens verdict.} Its ratings stay within a narrow band of 4--5 for every metadata lens. Grade level is rated 5 and labelled ``strong'' in all three runs, contrary to the external criterion ($V=0.13$), and the eating-habit-change lens receives only 6. \lp{A one-point gap thus separates the two lenses, for which \tool reports clearly different joint profiles (Tab.~\ref{Tab.Lens_Control}).} Across the seven metadata lenses without a significant association with the human code, the LLM labels the lens ``strong'' in 9 of 21 runs. (ii) \emph{Response triage.} The LLM's unusual picks reach a proxy hit rate of \lp{$0.47$} and its representative picks $0.63$, the latter \lp{below} the random level (\lp{$0.66$}). \tool reaches \lp{$0.82$} and $0.81$. \lp{Qwen3-Embed matches \tool on representative picks ($0.80$) but its unusual picks reach only $0.29$.} (iii) \emph{Response structure.} The LLM's content partitions agree less with the human codes (ARI $0.11$/$0.08$) than \tool reconstructions (ARI $0.30$/$0.11$), and the pair-level mismatch AUC is $0.613$ versus $0.749$ (Tab.~\ref{Tab.External_Baselines}). (iv) \emph{Order sensitivity.} Reordering the same responses changes most of the LLM's selections (mean top-3 Jaccard $0.12$) and partitions (mean pairwise ARI $0.37$). \lp{\tool computes the report from a fixed distance matrix, so presenting the responses in a different order to this stage leaves its picks unchanged\lp{; this shows that the reporting stage is deterministic, not that scoring with adaptive anchors is order-invariant}.} Qwen3-Embed attains \lp{higher neighbour purity than \tool} (Tab.~\ref{Tab.External_Baselines}) but near-zero partition agreement, so generic semantic similarity alone does not recover the criterion-specific structure.}

\lp{\textbf{Scope and cost.} The direct LLM sees the lens groups, which can bias its lens verdict; the prompt instructs it to ignore groups when clustering. \lp{It receives the research question but not the refined case-study rubric, so the comparison contrasts end-to-end judgment with \toole's per-response scoring and fixed aggregation rather than isolating a single component.} \lp{The matched variant with the same final instruction, fixed examples, and per-response scoring is the \texttt{Fixed Example} ablation on the synthetic data (Tab.~\ref{Tab.Ablation}).} The proxy triage score uses only the coarse code and is not a substitute for the evaluator study (Appendix~\ref{Sec:expert_validation_appendix}). \lp{Because \tool scores follow the same construct as this code, the proxy checks that picks are consistent with the criterion, not that they are worth reading.} The 53 direct-LLM calls \lp{(including retries)} used 256{,}444 tokens; the Qwen3 embeddings used 1{,}714 tokens; the strong-lens analysis (Appendix~\ref{Sec:lens_control_appendix}) required no LLM calls.}

\subsection{\lp{Code-then-Test Check of Group Separation}}
\label{Sec:code_then_test_appendix}

\lp{\textbf{Question and rationale.} In the scalar regime, the case-study instruction is a nine-band rubric with 50 meaning no change (Appendix~\ref{Sec:prompts_appendix_casestudy}), so each \tool score $S(d)$ is already a rubric code. \lp{Standard association tests between these scores and lens groups can thus be applied directly. If DDR, the group-separation component of the profile, measures separation as intended, it should agree with these tests on the same scores; we use them as a convergent-validity check of DDR, and Result (iii) shows what the full profile adds beyond a test.}}

\lp{\textbf{Protocol.} We use the case-study \tool scores $S(d)$, from which $\delta_{ij}=|S(d_i)-S(d_j)|$. Over the nine lenses of Appendix~\ref{Sec:case_study_baselines_appendix}, we compute $\chi^2$ between lens and three-way rubric code ($<50$ worse, $=50$ same, $>50$ better), Kruskal--Wallis (KW) and its effect size $\varepsilon^2$ on the scores, and PERMANOVA (999 permutations) on $|S(d_i)-S(d_j)|$. DDR is computed on the same matrix with a 500-permutation $p$-value. All tests are computed locally without LLM calls.}

\begin{table*}[t]
\centering
\caption{\lp{\lp{Convergent-validity check of DDR}, the group-separation component of the \tool profile, \lp{against standard tests} on the same case-study scores. $V^\ast$: association of the lens with the human code (unclear excluded). $\chi^2$ uses the three-way rubric code; KW is Kruskal--Wallis on scores; PERM is PERMANOVA on $|S(d_i)-S(d_j)|$. DDR follows Sec.~\ref{Sec:Framework_Intro}. Entries are $p$-values unless noted; $p<0.05$ is in bold.}}
\label{Tab.Code_Then_Test}
\setlength{\tabcolsep}{7pt}
\renewcommand{\arraystretch}{1.05}
\small
\begin{tabular}{lcc|cccc|cc}
\toprule
& & & \multicolumn{4}{c|}{\lp{Standard tests on \tool scores}} & \multicolumn{2}{c}{\lp{DDR (separation)}} \\
\lp{Lens} & \lp{$N$/$C$} & \lp{$V^\ast$} & \lp{$\chi^2$} & \lp{KW} & \lp{$\varepsilon^2$} & \lp{PERM} & \lp{value} & \lp{$p$}  \\
\midrule
\lp{Grade level} & \lp{122/4} & \lp{0.11} & \lp{0.755} & \lp{0.271} & \lp{0.01} & \lp{0.519} & \lp{0.02} & \lp{0.381}  \\
\lp{Gender} & \lp{122/2} & \lp{0.10} & \lp{0.932} & \lp{0.297} & \lp{0.00} & \lp{0.537} & \lp{0.04} & \lp{0.112}  \\
\lp{Living situation} & \lp{119/3} & \lp{0.17} & \lp{0.701} & \lp{0.706} & \lp{0.00} & \lp{0.781} & \lp{0.00} & \lp{1.000}  \\
\lp{Current diet} & \lp{119/3} & \lp{0.25} & \lp{0.214} & \lp{\textbf{0.031}} & \lp{0.04} & \lp{\textbf{0.023}} & \lp{0.03} & \lp{0.164}  \\
\lp{Self-rated health} & \lp{122/3} & \lp{0.20} & \lp{0.181} & \lp{0.080} & \lp{0.03} & \lp{0.056} & \lp{0.03} & \lp{0.100}  \\
\lp{Does sports} & \lp{120/2} & \lp{0.19} & \lp{0.255} & \lp{\textbf{0.024}} & \lp{0.03} & \lp{0.053} & \lp{0.03} & \lp{0.168}  \\
\lp{Cooking frequency} & \lp{119/5} & \lp{0.16} & \lp{0.720} & \lp{0.056} & \lp{0.05} & \lp{0.089} & \lp{0.02} & \lp{0.325}  \\
\lp{Self-perceived weight} & \lp{121/6} & \lp{0.13} & \lp{0.876} & \lp{0.802} & \lp{0.00} & \lp{0.861} & \lp{0.01} & \lp{0.343}  \\
\midrule
\lp{Eating-habit change (human code)} & \lp{119/3} & \lp{1.00} & \lp{\textbf{$<$0.001}} & \lp{\textbf{$<$0.001}} & \lp{0.45} & \lp{\textbf{0.001}} & \lp{0.52} & \lp{\textbf{0.002}}  \\
\bottomrule
\end{tabular}
\end{table*}

\lp{\textbf{Results.} (i) \emph{Same verdict at both ends.} No test detects a grade effect ($p\ge0.27$; DDR $p=0.38$), and all detect the human-code effect ($p\le0.002$; Tab.~\ref{Tab.Code_Then_Test}). (ii) \lp{\emph{Detection versus effect size.}} KW and PERMANOVA flag current diet ($p=0.031$ and $0.023$), the metadata lens most associated with the human code ($V^\ast=0.25$), and KW also flags sports, whereas \lp{the DDR permutation test does not reach $p<0.05$ for any metadata lens. A test answers whether group scores differ detectably; DDR, like $\varepsilon^2$, is an effect size that states how much of the dissimilarity structure the lens explains. On both effect sizes these two lenses are small ($\varepsilon^2\le0.04$, DDR $=0.03$), against $0.45$ and $0.52$ for the human code, so their group differences are detectable but explain little of the response structure. We therefore read DDR as an effect size reported alongside the test rather than as a competing test.} \lp{(iii) \emph{What the profile adds beyond a test.} A test returns one lens-level decision; for grade level, no detectable effect ($p\ge0.27$). A non-significant test is consistent both with homogeneous responses and with varied responses organized along another axis, and it does not indicate a next step. The profile on the same responses separates these cases (Tab.~\ref{Tab.Lens_Control}): high DC ($0.90$) shows that the responses are not homogeneous, while low DP ($0.43$), GMR ($0.39$), and DDR ($0.02$) show that the clusters mix grades and do not follow the grade groups. This joint reading points to themes that cut across grades and corresponds to the second row of Tab.~\ref{Tab.Signal_Interpretation} (revise the lens or code cross-group themes; read unusual responses first), whereas the human-code profile corresponds to the first row (compare groups; read representative responses first). Thus GMR and DDR grade how far the lens explains response structure rather than only whether groups differ, and DP and DC describe group mixing and remaining within-cluster variation, which a lens-level test does not address. In the controlled distribution-shift experiment, the four metrics also respond in different patterns to different structural changes (Fig.~\ref{fig.parasweep}). In the scalar regime, standard tests can be reported alongside the profile; in the pairwise regime no single score exists, so only distance-based tests such as PERMANOVA on $\mathbf{D}$ remain applicable. (iv) \emph{Response mining in the scalar regime.} Because the mean within-group distance $\frac{1}{n}\sum_{j}|S(d_i)-S(d_j)|$ is smallest at the group median, ranking responses by their distance to the group median score reproduces 23 of the 24 grade-level picks, confirming that the mining rule behaves as intended; this shortcut does not exist in the pairwise regime, where no single score is available.}}

\subsection{\lp{Independent Evaluator Validation}}
\label{Sec:expert_validation_appendix}

\begin{figure*}[t]
    \centering
    \includegraphics[width=\textwidth]{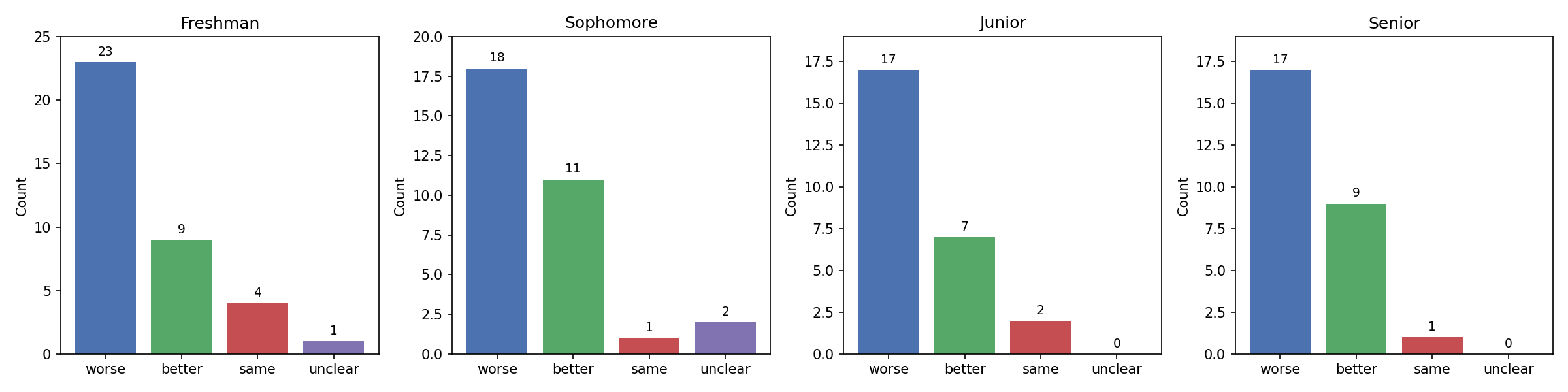}
    \caption{Behavior changes across different grade levels.}
    \label{fig:distribution_gradelevel}
\end{figure*}

\lp{\textbf{Question and rationale.} We conducted this study to examine whether independent qualitative judgments align with the case-study report and whether the report offers useful starting points for subsequent coding.}

\lp{\textbf{Participants.} We recruited three evaluators through a social media advertisement; most had backgrounds in human--computer interaction or human-centered research. All evaluators provided informed consent before participating and received USD 10 as compensation. Names were used only to resume sessions and arrange compensation, and were replaced by anonymous identifiers (E1--E3) in all analyses.}

\lp{\textbf{Step 1: blinded familiarization.} Evaluators received the research question and all 122 raw responses grouped by grade; the \tool report and thematic labels remained hidden. For each grade, the responses were displayed in full by default, and evaluators confirmed having viewed them before proceeding. This step also contained response-level typical/unusual judgments and a lens-helpfulness rating. In post-study debriefing, evaluators reported that the purpose of these judgments was unclear to them; we therefore treat Step 1 as familiarization with the data and do not analyze its judgments.} \lp{This decision was made after data collection.}

\lp{\textbf{Step 2: report assessment.} After completing Step 1, evaluators inspected the \tool report, comprising the four metrics, their interpretation, and the surfaced responses. They rated three statements on seven-point agreement scales (1 = strongly disagree; 7 = strongly agree): \emph{correctness} (the report's conclusion about grade level is consistent with their own reading of the responses), \emph{interpretability} (the metrics and their explanations are easy to understand), and \emph{usefulness for subsequent coding} (the surfaced representative and unusual responses would help them start coding the data), followed by an optional comment. We report individual values and the median and range for each rating. With three evaluators, the evidence is interpreted as face validity; population-level inference requires a larger sample.} \lp{The study has no comparison condition, so it does not test whether the report improves coding relative to reading the responses alone. The report shown to evaluators listed one non-substantive answer (C32, ``none'') among the sophomore unusual responses; Appendix~\ref{Sec:case_study_mined_responses} excludes such answers.}

\begin{table}[t]
\centering
\caption{\lp{Step-2 report ratings by the three evaluators (seven-point agreement scales; higher is better).}}
\label{Tab.Evaluator_Validation}
\setlength{\tabcolsep}{4pt}
\renewcommand{\arraystretch}{1.05}
\small
\begin{tabular}{lccccc}
\toprule
\lp{Statement} & \lp{E1} & \lp{E2} & \lp{E3} & \lp{Median} & \lp{Range} \\
\midrule
\lp{Correctness} & \lp{4} & \lp{6} & \lp{7} & \lp{6} & \lp{4--7} \\
\lp{Interpretability} & \lp{6} & \lp{7} & \lp{6} & \lp{6} & \lp{6--7} \\
\lp{Usefulness for coding} & \lp{6} & \lp{7} & \lp{6} & \lp{6} & \lp{6--7} \\
\bottomrule
\end{tabular}
\end{table}

\lp{\textbf{Results.} Tab.~\ref{Tab.Evaluator_Validation} reports the Step-2 ratings. No rating fell below the scale midpoint, and eight of the nine ratings were 6 or 7. Interpretability and usefulness for subsequent coding both reached a median of 6 (range 6--7), indicating that evaluators found the metrics and explanations understandable and the surfaced representative and unusual responses useful as starting points for coding. Correctness also reached a median of 6 but varied more (range 4--7): two evaluators agreed that the conclusion that grade level is a weak lens matched their own reading, whereas E1 was neutral. No evaluator left an optional comment. These results provide face-validity evidence that analysts find the \tool report understandable and useful for starting subsequent coding; they do not establish effects on final coding outcomes\lp{, nor that the unusual responses yield new insights, which would require independent judgments of their analytical value}.}

\subsection{\lp{Case Study Materials}}

\subsubsection{\lp{Question and Response Examples}}
\label{Sec:case_study_qa_example}
\lp{The following excerpts illustrate the raw qualitative inputs and response variation summarized by \tool.}

\textbf{Question} 
\colorlet{punct}{red!60!black}
\definecolor{background}{HTML}{EEEEEE}
\definecolor{delim}{RGB}{20,105,176}
\colorlet{numb}{magenta!60!black}
\lstdefinelanguage{json}{
     numbers=left,
     numberstyle=\scriptsize,
     stepnumber=1,
     numbersep=8pt,
     showstringspaces=false,
     breaklines=true,
     frame=lines,
     backgroundcolor=\color{background},
     literate=
     *{0}{{{\color{numb}0}}}{1}
      {1}{{{\color{numb}1}}}{1}
      {2}{{{\color{numb}2}}}{1}
      {3}{{{\color{numb}3}}}{1}
      {4}{{{\color{numb}4}}}{1}
      {5}{{{\color{numb}5}}}{1}
      {6}{{{\color{numb}6}}}{1}
      {7}{{{\color{numb}7}}}{1}
      {8}{{{\color{numb}8}}}{1}
      {9}{{{\color{numb}9}}}{1}
      {:}{{{\color{punct}{:}}}}{1}
      {,}{{{\color{punct}{,}}}}{1}
      {\{}{{{\color{delim}{\{}}}}{1}
      {\}}{{{\color{delim}{\}}}}}{1}
      {[}{{{\color{delim}{[}}}}{1}
      {]}{{{\color{delim}{]}}}}{1},
}
\begin{lstlisting}[language=json,firstnumber=1]
Describe your eating changes since the moment you got into college
\end{lstlisting}

\textbf{Answers example}
\colorlet{punct}{red!60!black}
\definecolor{background}{HTML}{EEEEEE}
\definecolor{delim}{RGB}{20,105,176}
\colorlet{numb}{magenta!60!black}
\lstdefinelanguage{json}{
     numbers=left,
     numberstyle=\scriptsize,
     stepnumber=1,
     numbersep=8pt,
     showstringspaces=false,
     breaklines=true,
     frame=lines,
     backgroundcolor=\color{background},
     literate=
     *{0}{{{\color{numb}0}}}{1}
      {1}{{{\color{numb}1}}}{1}
      {2}{{{\color{numb}2}}}{1}
      {3}{{{\color{numb}3}}}{1}
      {4}{{{\color{numb}4}}}{1}
      {5}{{{\color{numb}5}}}{1}
      {6}{{{\color{numb}6}}}{1}
      {7}{{{\color{numb}7}}}{1}
      {8}{{{\color{numb}8}}}{1}
      {9}{{{\color{numb}9}}}{1}
      {:}{{{\color{punct}{:}}}}{1}
      {,}{{{\color{punct}{,}}}}{1}
      {\{}{{{\color{delim}{\{}}}}{1}
      {\}}{{{\color{delim}{\}}}}}{1}
      {[}{{{\color{delim}{[}}}}{1}
      {]}{{{\color{delim}{]}}}}{1},
}
\begin{lstlisting}[language=json,firstnumber=1]
...
"sometimes choosing to eat fast food instead of cooking simply for convenience."

"Accepting cheap and premade/store bought foods."

"I have eaten generally the same foods but I do find myself eating the same food frequently due to what I have found I like from egan and the laker."

"I started eating a lot less and healthier because I wasn't playing sports year round anymore."

"Freshmen year i ate very unhealthy, but now it is much healthier because of self control."
...
\end{lstlisting}

\subsubsection{\lp{Grade-Level Label Distribution}}
\label{Sec:case_study_grade_distribution}
\lp{Fig.~\ref{fig:distribution_gradelevel} shows that the four thematic labels have similar proportions across grade levels, providing the distributional context for the weak association reported in Tab.~\ref{Tab.External_Criterion}.}

\AppendixBeforeBox
\subsubsection{\lp{Mined Responses (Top-3)}}
\label{Sec:case_study_mined_responses}

\begin{fullwidth}
\begin{minedresponsesbox}{\lp{Representative and Unusual Responses (Top-3)}}
\lp{Representative responses capture recurring eating-change patterns, while unusual responses surface less common trajectories within each grade for targeted inspection.} \lp{Non-substantive answers that give no description of eating habits (``none'', ``Nun'', ``None really'') are excluded from the lists, and the next-ranked response is shown.}

\textbf{Freshman}
\begin{itemize}
    \item \textit{Representative:}
    \begin{itemize}
        \item C73: I eat a lot less and more junk food.
        \item C17: More Water
        \item C94: I eat alot of carbs and eat much more frequently
    \end{itemize}
    \item \textit{\lp{Unusual:}}
    \begin{itemize}
        \item C97: I've eaten more fruits and vegetables. Started eating seafood.
        \item C13: I have been eating a lot more salads and soups.
        \item C91: i eat healthier all around
    \end{itemize}
\end{itemize}

\textbf{Sophomore}
\begin{itemize}
    \item \textit{Representative:}
    \begin{itemize}
        \item C88: less healthy because of less options, money and time.
        \item C109: Late night snacking
        \item C66: I snack more, having fewer full meals
    \end{itemize}
    \item \textit{\lp{Unusual:}}
    \begin{itemize}
        \item C85: Huge changes have occurred. I eat far healthier, less processed food, less dense carbohydrates and way more vegetables and fruits.
        \item C118: I eat more vegetable. Since coming to college, I started to eat salads and tried to eat salads at least three times a week.
        \item \lp{C46: I eat smaller portions more often}
    \end{itemize}
\end{itemize}

\textbf{Junior}
\begin{itemize}
    \item \textit{Representative:}
    \begin{itemize}
        \item C72: I tend to snack more and have smaller meals.
        \item C71: I don't eat as often
        \item C15: I knew I would eat alot my freshmen year, before coming to college i had a diet plan.
    \end{itemize}
    \item \textit{\lp{Unusual:}}
    \begin{itemize}
        \item C44: Now I prepare my own meals, pack my lunch every day and avoid eating out to save money as much as possible.
        \item C90: I have been eating healthier especially vegetables and proteinous food
        \item C93: I have increased the amounts of vegetables I eat due to the unhealthy options in our dining halls
    \end{itemize}
\end{itemize}

\textbf{Senior}
\begin{itemize}
    \item \textit{Representative:}
    \begin{itemize}
        \item C77: I eat less healthy breakfast now, usually just grab something quick like a granola bar.
        \item C45: I haven't changed much. If anything, I have become more disciplined.
        \item C121: I have noticed there is less time for a prepared meal, so quick and easy has become the norm.
    \end{itemize}
    \item \textit{\lp{Unusual:}}
    \begin{itemize}
        \item C113: I had to change a lot. I keep track of calories and cut out most breads and wraps.
        \item C57: I have begun to eat more fruits and vegetables because I have been more aware of my physique.
        \item C63: As an athlete it is important to fuel my body with important foods only.
    \end{itemize}
\end{itemize}
\end{minedresponsesbox}
\end{fullwidth}

\AppendixBeforeBox
\section{\lp{Supporting Iterative Qualitative Data Analysis}}
\label{sec:responsible_use_appendix}
\label{Sec:qha_discussion}

\lp{Many qualitative-data-analysis workflows interleave data collection and analysis. Researchers often begin familiarization and preliminary coding on a pilot subset, then refine the research question, sampling strategy, codebook, or analytical lens as patterns emerge. New data may prompt further revisions to the study scope or earlier interpretations, with contextual and theoretical judgment guiding these decisions throughout the process~\citep{glaser2017discovery,campbell2013coding,ames2024assessing,naeem2024demystification}.}

\lp{\tool can support this iterative workflow as a repeatable diagnostic checkpoint. On a pilot corpus, researchers can use the pairwise regime to assess a proposed analytical lens and identify central or unusual responses for close reading. They can then combine these signals with preliminary coding to revise the evaluation instruction, analytical lens, sampling plan, or study scope. As responses accumulate, the same assessment can be repeated to examine how the dataset structure changes; the scalar-projection regime supports this process when the corpus becomes larger.}

\lp{At each checkpoint, \lp{the four metrics are read together as one profile, from lens-level relations to within-cluster variation}. GMR summarizes alignment between the selected lens and reconstructed response clusters; DDR summarizes how well the analytical groups separate responses; DC characterizes within-\lp{cluster} dispersion; and DP characterizes analytical-group mixing within clusters. A separate response-mining ranking identifies central and unusual cases. Researchers can use these signals to continue with the current lens, investigate cross-group themes, inspect outliers, or revise the next round of data collection and analysis. Tab.~\ref{Tab.Signal_Interpretation} maps typical signal combinations to these next steps; the food-choice case study corresponds to the second row. \lp{These mappings are reading guides for close inspection rather than automatic decision rules.}}

\begin{table}[t]
\centering
\caption{\lp{Reading the joint QuAD profile at an analysis checkpoint.}}
\label{Tab.Signal_Interpretation}
\setlength{\tabcolsep}{3pt}
\renewcommand{\arraystretch}{1.15}
\small
\begin{tabular}{@{}>{\raggedright\arraybackslash}p{\dimexpr0.25\columnwidth-2\tabcolsep}>{\raggedright\arraybackslash}p{\dimexpr0.36\columnwidth-2\tabcolsep}>{\raggedright\arraybackslash}p{\dimexpr0.39\columnwidth}@{}}
\toprule
\lp{Signals} & \lp{Reading} & \lp{Next step} \\
\midrule
\lp{High GMR, DP, DDR} & \lp{The lens explains response structure} & \lp{Compare groups; read representative responses first} \\
\lp{Low GMR, DP, DDR; high DC} & \lp{The lens weakly explains structure; themes cut across groups} & \lp{Revise the lens or code cross-group themes; read unusual responses first} \\
\lp{High GMR, high DC} & \lp{Groups are recovered with wide internal spread} & \lp{Code sub-themes within groups} \\
\bottomrule
\end{tabular}
\end{table}

\lp{Current evidence supports using \tool for analytical-lens assessment and response triage at iterative pre-analysis checkpoints. The independent-evaluator study further examined whether analysts judge the report's lens-level conclusion consistent with their own reading and find the report useful for subsequent coding. Establishing effects on final codes, themes, and interpretations requires a controlled comparison of QDA workflows with and without \tool assistance. Operational scaling and cost measurements are reported separately in Appendix~\ref{Sec:scalability_appendix}. Limitations and ethical considerations are stated after the Conclusion.}

\AppendixBeforeBox
\section{\lp{Experimental Materials}}
\label{Sec:experimental_materials_appendix}
\subsection{\lp{User Study Protocol and Materials}}
This appendix provides additional details supporting the controlled user study reported in Sec.~\ref{Sec.Exp_User_Study}. We include (i) the within-subject scheduling used for counterbalancing, (ii) statistical comparisons for the usability questionnaire dimensions shown in Fig.~\ref{fig.UserStudy}, and (iii) the UI screens and (iv) questionnaires used to collect NASA-TLX and usability ratings.

\subsubsection{\lp{Within-Subject User Study Schedule}}
Table~\ref{tab:user_study} documents the counterbalancing plan to mitigate ordering effects in the within-subject design described in Sec.~\ref{Sec.Exp_User_Study}.

\begin{table}[t]
\centering
\caption{Overview of User Study Schedule. We recruited 20 postgraduate students (11M/9F, aged 21-31) from diverse domains. Participants self-rated their domain experience ($M=3.50, SD=0.83$) and familiarity with instruction generation ($M=3.15, SD=1.27$) on a 5-point Likert scale. The system implements the adaptation phase of Sec.~\ref{Sec:Instruction_Refinement}. The baseline condition utilized the same UI but deactivated the agent-assisted features, requiring users to compose instructions manually. Participants were tasked with generating instructions for dissimilarity extraction for two distinct questions, using both ours and the baseline. \lp{Each session was scheduled for one hour, including about 45 minutes of tasks, and was compensated with 8 USD. Participants agreed to share their data for this research.}}
\label{tab:user_study}
\resizebox{\columnwidth}{!}{
\setlength{\tabcolsep}{3pt}
\renewcommand{\arraystretch}{0.9}
\begin{tabular}{@{}lcccccc@{}}
\toprule
\textbf{ID} & \textbf{Group} & \textbf{Participant ID} & \textbf{Method Order} & \textbf{Task 1} & \textbf{Task 2} & \textbf{Time} \\
\midrule
1 & Finance & Participant 1.1 & A $\to$ B & Q1 (Method A) & Q2 (Method B) & 1h \\
2 & Finance & Participant 1.2 & A $\to$ B & Q2 (Method A) & Q3 (Method B) & 1h \\
3 & Finance & Participant 1.3 & A $\to$ B & Q3 (Method A) & Q1 (Method B) & 1h \\
4 & Finance & Participant 1.4 & B $\to$ A & Q4 (Method B) & Q5 (Method A) & 1h \\
5 & Finance & Participant 1.5 & B $\to$ A & Q5 (Method B) & Q4 (Method A) & 1h \\
\midrule
6 & HCI & Participant 2.1 & A $\to$ B & Q6 (Method A) & Q7 (Method B) & 1h \\
7 & HCI & Participant 2.2 & A $\to$ B & Q7 (Method A) & Q8 (Method B) & 1h \\
8 & HCI & Participant 2.3 & A $\to$ B & Q8 (Method A) & Q9 (Method B) & 1h \\
9 & HCI & Participant 2.4 & B $\to$ A & Q6 (Method B) & Q10 (Method A) & 1h \\
10 & HCI & Participant 2.5 & B $\to$ A & Q10 (Method B) & Q9 (Method A) & 1h \\
\midrule
11 & Crowdsourcing & Participant 3.1 & A $\to$ B & Q11 (Method A) & Q12 (Method B) & 1h \\
12 & Crowdsourcing & Participant 3.2 & A $\to$ B & Q12 (Method A) & Q11 (Method B) & 1h \\
13 & Crowdsourcing & Participant 3.3 & B $\to$ A & Q13 (Method B) & Q14 (Method A) & 1h \\
14 & Crowdsourcing & Participant 3.4 & B $\to$ A & Q14 (Method B) & Q15 (Method A) & 1h \\
15 & Crowdsourcing & Participant 3.5 & B $\to$ A & Q15 (Method B) & Q13 (Method A) & 1h \\
\midrule
16 & Medical & Participant 4.1 & A $\to$ B & Q18 (Method A) & Q17 (Method B) & 1h \\
17 & Medical & Participant 4.2 & A $\to$ B & Q17 (Method A) & Q18 (Method B) & 1h \\
18 & Medical & Participant 4.3 & B $\to$ A & Q16 (Method B) & Q19 (Method A) & 1h \\
19 & Medical & Participant 4.4 & B $\to$ A & Q19 (Method B) & Q20 (Method A) & 1h \\
20 & Medical & Participant 4.5 & B $\to$ A & Q20 (Method B) & Q16 (Method A) & 1h \\
\bottomrule
\end{tabular}
}
\end{table}

\subsubsection{\lp{Statistical Significance of User Study Results}}
Table~\ref{tab:statistical_comparison} provides the per-item statistical comparison corresponding to the usability results summarized in Fig.~\ref{fig.UserStudy} (main paper).

\begin{table}[t]
\centering
\caption{Statistical Comparison Between Baseline and \toole}
\label{tab:statistical_comparison}
\resizebox{\columnwidth}{!}{
\setlength{\tabcolsep}{3pt}
\renewcommand{\arraystretch}{0.9}
\begin{tabular}{@{}ccccccc@{}}
\toprule
\textbf{Dimension} & \textbf{Baseline Mean} & \textbf{Ours Mean} & \textbf{Mean Difference} & \textbf{t-statistic} & \textbf{p-value} & \textbf{Significant} \\
\midrule
1  & 3.30 & 6.35 & 3.05 & 6.45 & \lp{$<$0.00001} & Yes \\
2  & 3.35 & 5.85 & 2.50 & 6.69 & \lp{$<$0.00001} & Yes \\
3  & 3.00 & 6.10 & 3.10 & 5.66 & 0.00002 & Yes \\
4  & 2.95 & 6.45 & 3.50 & 8.46 & \lp{$<$0.00001} & Yes \\
5  & 2.85 & 6.35 & 3.50 & 8.74 & \lp{$<$0.00001} & Yes \\
6  & 2.90 & 6.00 & 3.10 & 8.24 & \lp{$<$0.00001} & Yes \\
7  & 3.30 & 6.25 & 2.95 & 7.90 & \lp{$<$0.00001} & Yes \\
8  & 3.05 & 5.70 & 2.65 & 5.98 & 0.00001 & Yes \\
9  & 3.75 & 6.25 & 2.50 & 5.35 & 0.00004 & Yes \\
10 & 3.30 & 6.30 & 3.00 & 7.55 & \lp{$<$0.00001} & Yes \\
\bottomrule
\end{tabular}
}
\end{table}

\begin{table}[t]
\centering
\caption{Participant Demographic Information. Ratings are on a 1--5 Likert scale. We report Prompt Familiarity / ChatGPT Usage / System Willingness as P/C/S.}
\label{tab:survey_results}

\setlength{\tabcolsep}{4pt}
\renewcommand{\arraystretch}{0.95}
\small
\resizebox{\columnwidth}{!}{%
\begin{tabular}{@{}cccc>{\raggedright\arraybackslash}p{0.33\columnwidth}ccc@{}}
\toprule
\textbf{P ID} & \textbf{G} & \textbf{Age} & \textbf{Edu} & \textbf{Field} & \textbf{Prof.} & \textbf{Exp.} & \textbf{P/C/S} \\
\midrule
1.1 & 1 & 23 & 2 & Finance Data Analysis & 4 & 3    & 4/4/4 \\
1.2 & 1 & 24 & 3 & Economics              & 4 & 7    & 3/4/4 \\
1.3 & 1 & 23 & 1 & Finance                & 3 & 4    & 2/4/5 \\
1.4 & 2 & 22 & 2 & Finance                & 5 & 2    & 5/5/5 \\
1.5 & 2 & 23 & 2 & Finance                & 4 & 4    & 3/4/5 \\
\midrule
2.1 & 1 & 21 & 1 & Computer Science, HCI \& VIS & 4 & 3 & 5/5/4 \\
2.2 & 1 & 26 & 3 & HCI                    & 4 & 3    & 2/4/4 \\
2.3 & 2 & 21 & 2 & HCI                    & 3 & 3    & 3/5/5 \\
2.4 & 2 & 24 & 3 & HCI                    & 4 & 4    & 4/3/4 \\
2.5 & 1 & 31 & 3 & VIS/HCI                & 5 & 5    & 4/4/5 \\
\midrule
4.1 & 2 & 26 & 3 & Medical AI             & 4 & 5    & 5/5/4 \\
4.2 & 2 & 23 & 3 & Medicine---Neurobiology& 3 & 5    & 2/4/4 \\
4.3 & 2 & 22 & 1 & Medicine               & 2 & 5    & 1/2/5 \\
4.4 & 1 & 23 & 1 & Dentistry              & 3 & 5    & 2/2/4 \\
4.5 & 2 & 26 & 1 & Thoracic Surgery       & 3 & 5    & 2/2/4 \\
\midrule
3.1 & 1 & 23 & 3 & LLM SFT                & 3 & 1.25 & 4/5/5 \\
3.2 & 1 & 21 & 1 & LLM SFT                & 3 & 2    & 3/4/5 \\
3.3 & 1 & 25 & 2 & LLM, 3D Printing       & 4 & 2    & 4/5/5 \\
3.4 & 1 & 25 & 2 & 3D Printing            & 4 & 2    & 4/4/5 \\
3.5 & 1 & 21 & 1 & AI4Machine             & 4 & 3    & 4/4/5 \\
\bottomrule
\end{tabular}%
}
\end{table}

\subsubsection{\lp{Prompt Generation System Interface Overview}}
\begin{figure*}[t]
    \centering
    \includegraphics[width=0.88\textwidth]{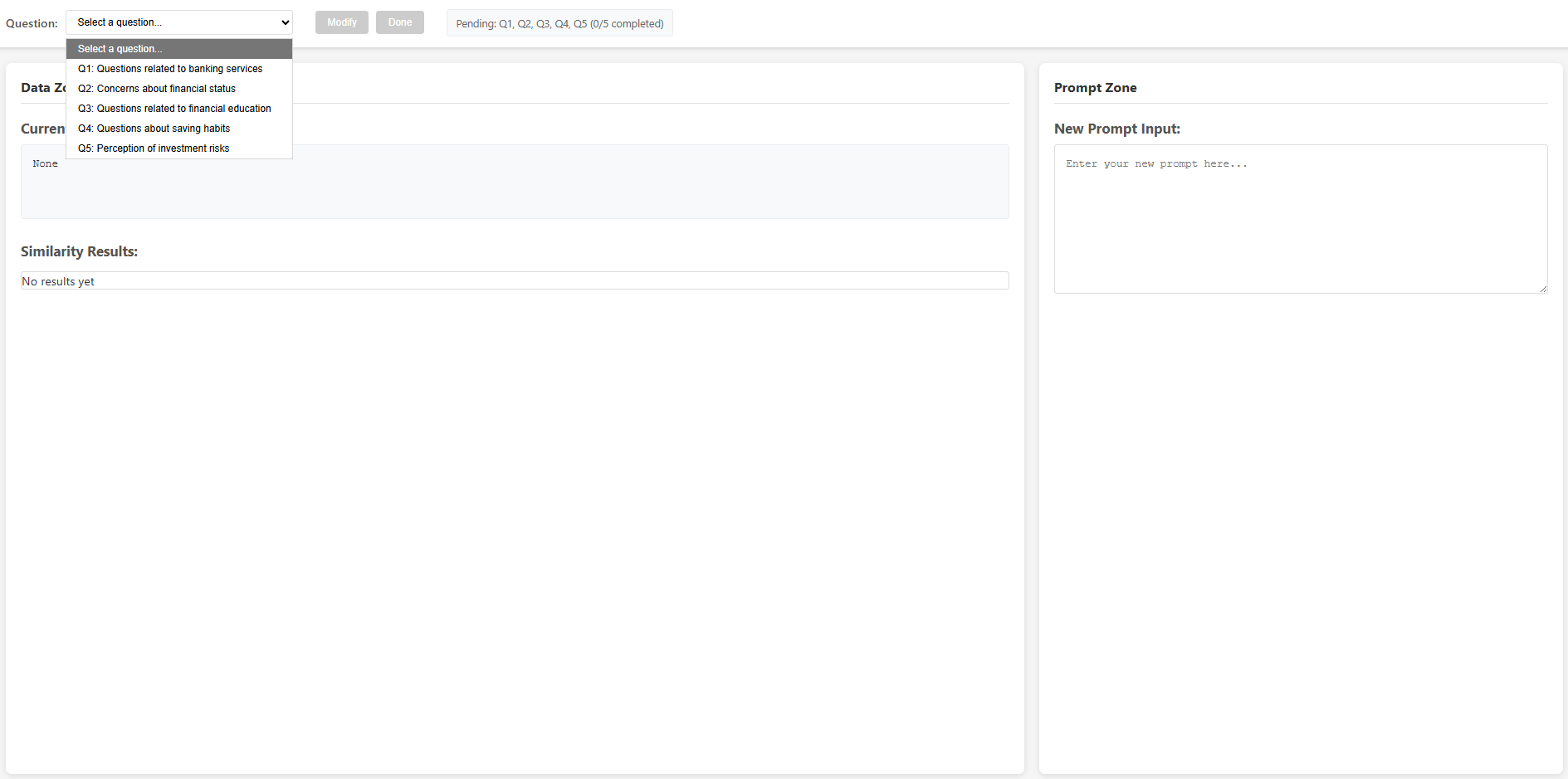}
    \caption{In the first stage, users select a question for which they want to generate a prompt. This serves as the entry point into the system.}
    \label{fig:entry}
\end{figure*}

\begin{figure*}[t]
    \centering
    \includegraphics[width=0.88\textwidth]{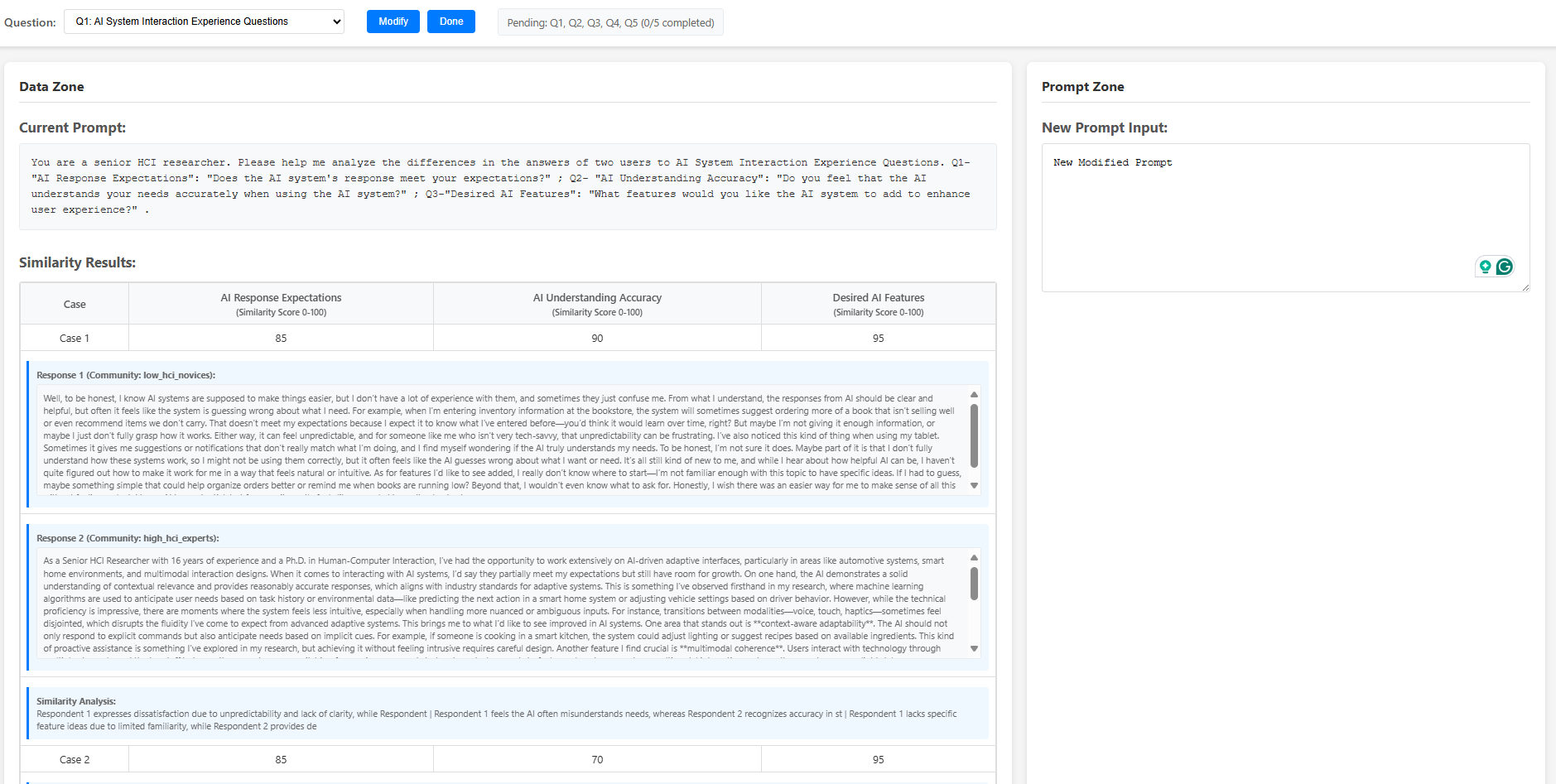}
    \caption{In the second stage, users manually input a prompt into the \emph{Prompt Zone} and click \emph{Modify} to initiate the dissimilarity extraction process. The system displays three illustrative cases in the \emph{Data Zone}, alongside the current prompt. Users may refine the prompt iteratively until satisfied, at which point they click \emph{Done} to save the final result.}
    \label{fig:usecase}
\end{figure*}

\begin{figure*}[t]
    \centering
    \includegraphics[width=0.88\textwidth]{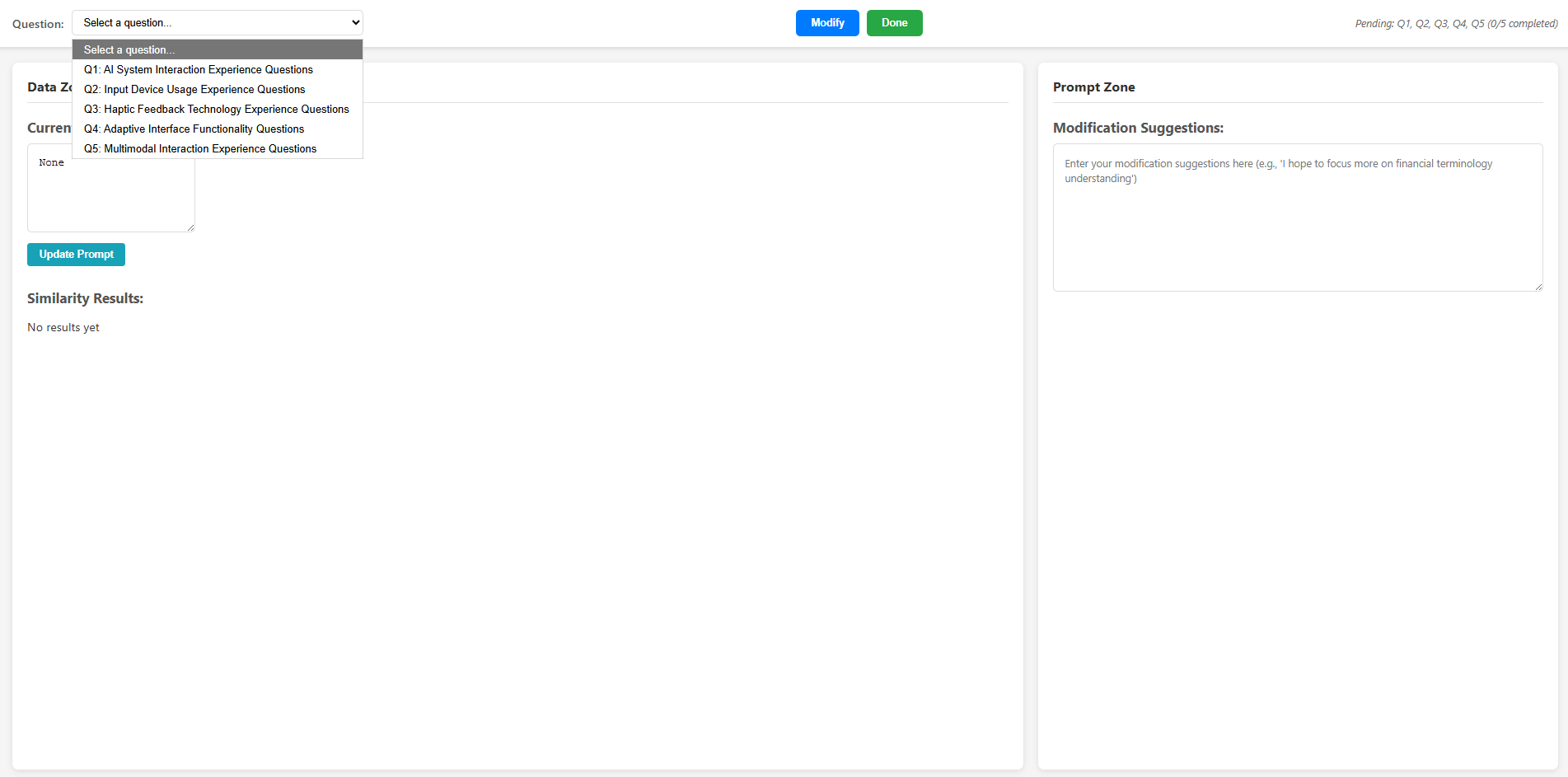}
    \caption{Upon entering the system, users select a question for which they wish to generate a prompt. This step mirrors the baseline system's question selection process.}
    \label{fig:ours_entry}
\end{figure*}

\begin{figure*}[t]
    \centering
    \includegraphics[width=0.88\textwidth]{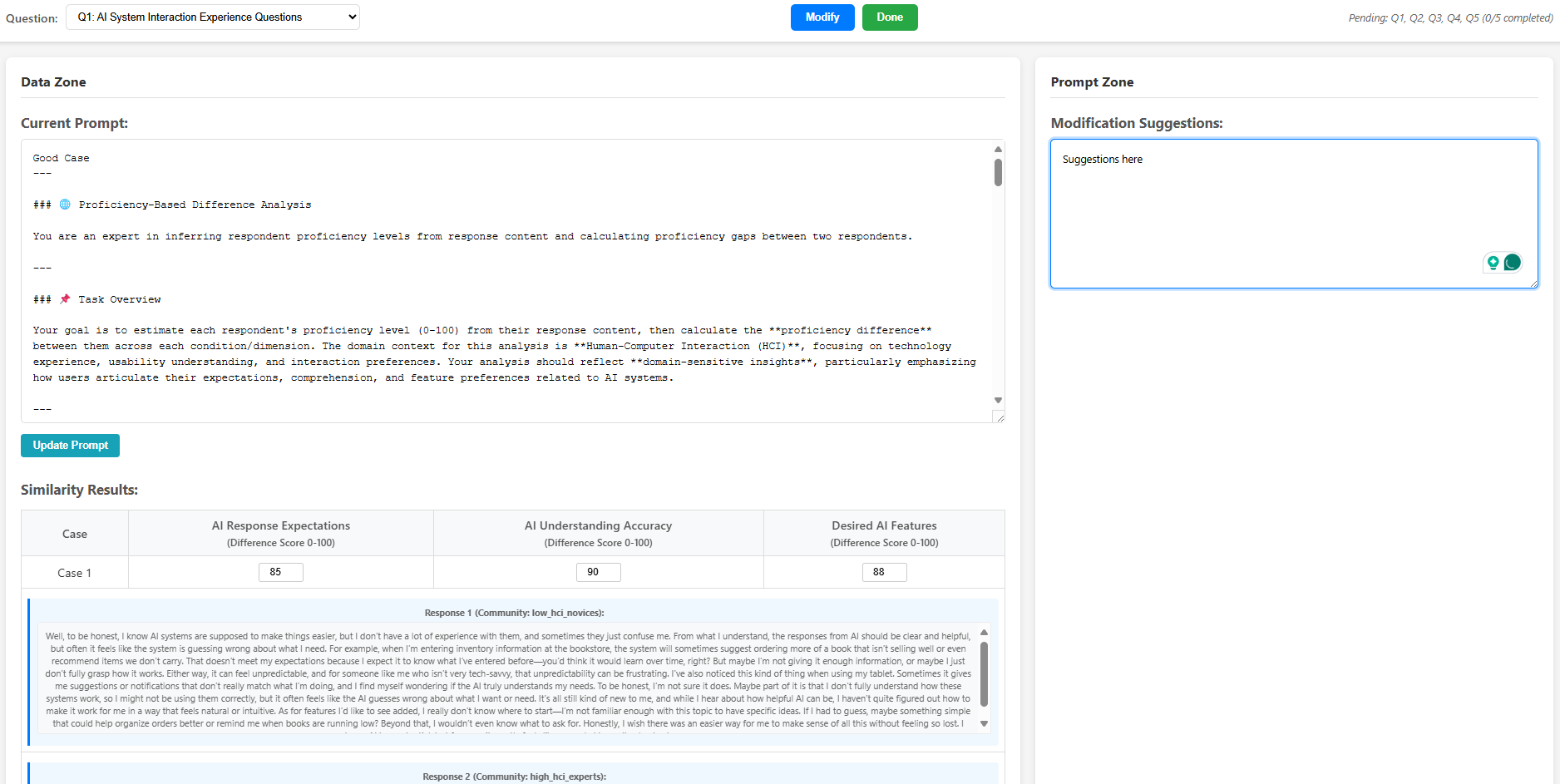}
    \caption{Upon accessing the system, users are presented with an auto-generated initial prompt and dissimilarity extraction results in the \emph{Data Zone}. Users can provide modifications in the \emph{Prompt Zone}, triggering the system to automatically adjust the prompt. Once users are satisfied, they can click the \emph{Done} button to save the final result. Alternatively, they can directly edit the prompt in the \emph{Prompt Zone} for manual adjustments.}
    \label{fig:ours_usecase}
\end{figure*}

\ws{The UI is not intended as a full qualitative-analysis workbench and is not the central contribution evaluated by the user study. Its goal is narrower: to provide a controlled interface through which participants generate or refine task-specific dissimilarity evaluation instructions for the Stage 2 human-in-the-loop adaptation described in Sec.~\ref{Sec:Instruction_Refinement}.}
\ws{Both experimental conditions use the same overall UI layout. The baseline condition disables agent-assisted features and requires participants to write or refine prompts manually, while our condition provides an auto-generated initial prompt and agent-assisted refinement. This design reduces the risk that the user-study results are driven by UI layout differences rather than instruction-generation assistance.}
The following UI screenshots correspond to the ``baseline'' and ``ours'' conditions described in Sec.~\ref{Sec.Exp_User_Study}, illustrating what participants interacted with when generating/refining evaluation instructions.

\paragraph{\lp{System Overview of the Baseline.}}
The baseline system consists of two primary stages: question selection \lp{(Fig.~\ref{fig:entry})} and prompt modification.

After selecting a question, users proceed to the prompt modification stage (Fig.~\ref{fig:usecase}).

\paragraph{\lp{System Overview of Our System.}}

Our proposed system integrates similar functionalities as the baseline but enhances the workflow with automated features. It is designed to handle both large-scale and small-scale use cases. For demonstration purposes, we focus on the small-scale scenario used in the user study.

\lp{Question selection mirrors the baseline (Fig.~\ref{fig:ours_entry}).} In contrast to the baseline system, our system provides an auto-generated initial prompt and dissimilarity extraction results upon entry (Fig.~\ref{fig:ours_usecase}).

\subsubsection{\lp{Participant Demographics}}
Table~\ref{tab:survey_results} reports participant background variables referenced in Sec.~\ref{Sec.Exp_User_Study} (domain experience and familiarity ratings).

\AppendixBeforeBox
\begin{fullwidth}
\subsubsection{\lp{NASA-TLX and System Usability Questionnaire}}
\label{Sec:questionnaire}

\begin{questionnairebox}{NASA-TLX and System Usability Questionnaire}
This questionnaire is designed to collect your subjective feedback following the experimental tasks. The results will be used solely for evaluating the rationale of the experimental design and the system's real-world performance. All data will be anonymized; by completing this questionnaire, you consent to the collection and analysis of your experience.

\paragraph{\lp{Participant Information.}}
\begin{enumerate}
    \item \textbf{Experiment ID:}  \\
    \underline{\hspace{\linewidth}} \\
    \item \textbf{Which phase are you in?}
    \begin{itemize}
        \item $\bigcirc$ Phase 1
        \item $\bigcirc$ Phase 2
    \end{itemize}
\end{enumerate}

\paragraph{\lp{NASA Task Load Index (NASA\_TLX).}}

\ws{This section applies the NASA Task Load Index (NASA\_TLX)~\citep{hart1988development} to evaluate perceived workload. We collect the standard NASA-TLX dimensions for completeness; however, physical demand is not central to our substantive claim for this prompt-generation task, and our interpretation focuses primarily on mental demand, temporal demand, effort, and frustration. For each dimension below, please rate on a scale of 1 (low) to 20 (high):}

\begin{enumerate}
    \item \textbf{Mental Demand:} How mentally demanding was the task? \\
    \textit{(Lower scores indicate less mental demand)} \\
    \underline{\hspace{\linewidth}}
    
    \item \textbf{Physical Demand:} How physically demanding was the task? \\
    \textit{(Lower scores indicate less physical demand)} \\
    \underline{\hspace{\linewidth}}
    
    \item \textbf{Temporal Demand:} How hurried or rushed was the pace of the task? \\
    \textit{(Lower scores indicate less time pressure)} \\
    \underline{\hspace{\linewidth}}
    
    \item \textbf{Performance:} How satisfied are you with your performance in the task? \\
    \textit{(Lower scores indicate lower satisfaction)} \\
    \underline{\hspace{\linewidth}}
    
    \item \textbf{Effort:} How hard did you have to work to accomplish your level of performance? \\
    \textit{(Lower scores indicate less effort)} \\
    \underline{\hspace{\linewidth}}
    
    \item \textbf{Frustration:} How insecure, discouraged, irritated, stressed, or annoyed did you feel? \\
    \textit{(Lower scores indicate less frustration)} \\
    \underline{\hspace{\linewidth}}
\end{enumerate}

\paragraph{\lp{Pairwise Comparison.}}

For each pair below, please select the factor that had a greater impact on you during the task:

\begin{enumerate}
    \item $\bigcirc$ Mental Demand \hspace{1cm} $\bigcirc$ Physical Demand
    \item $\bigcirc$ Mental Demand \hspace{1cm} $\bigcirc$ Temporal Demand
    \item $\bigcirc$ Mental Demand \hspace{1cm} $\bigcirc$ Effort
    \item $\bigcirc$ Mental Demand \hspace{1cm} $\bigcirc$ Performance
    \item $\bigcirc$ Mental Demand \hspace{1cm} $\bigcirc$ Frustration
    \item $\bigcirc$ Physical Demand \hspace{1cm} $\bigcirc$ Temporal Demand
    \item $\bigcirc$ Physical Demand \hspace{1cm} $\bigcirc$ Effort
    \item $\bigcirc$ Physical Demand \hspace{1cm} $\bigcirc$ Performance
    \item $\bigcirc$ Physical Demand \hspace{1cm} $\bigcirc$ Frustration
    \item $\bigcirc$ Temporal Demand \hspace{1cm} $\bigcirc$ Effort
    \item $\bigcirc$ Temporal Demand \hspace{1cm} $\bigcirc$ Performance
    \item $\bigcirc$ Temporal Demand \hspace{1cm} $\bigcirc$ Frustration
    \item $\bigcirc$ Effort \hspace{1cm} $\bigcirc$ Performance
    \item $\bigcirc$ Effort \hspace{1cm} $\bigcirc$ Frustration
    \item $\bigcirc$ Performance \hspace{1cm} $\bigcirc$ Frustration
\end{enumerate}

\paragraph{\lp{System Usability Assessment.}}

Please rate your experience with the system using the following statements. Use a 7-point Likert scale, where 1 indicates \textit{Very Dissatisfied} and 7 indicates \textit{Very Satisfied}:

\begin{enumerate}
    \item How satisfied are you with the time and effort required to generate an initial prompt? \\
    $\bigcirc$ 1 (Very Dissatisfied) \hspace{0.5cm} $\bigcirc$ 2 \hspace{0.5cm} $\bigcirc$ 3 \hspace{0.5cm} $\bigcirc$ 4 \hspace{0.5cm} $\bigcirc$ 5 \hspace{0.5cm} $\bigcirc$ 6 \hspace{0.5cm} $\bigcirc$ 7 (Very Satisfied)

    \item How satisfied are you with the ease of modifying or refining the prompt based on the generated results? \\
    $\bigcirc$ 1 (Very Dissatisfied) \hspace{0.5cm} $\bigcirc$ 2 \hspace{0.5cm} $\bigcirc$ 3 \hspace{0.5cm} $\bigcirc$ 4 \hspace{0.5cm} $\bigcirc$ 5 \hspace{0.5cm} $\bigcirc$ 6 \hspace{0.5cm} $\bigcirc$ 7 (Very Satisfied)

    \item How satisfied are you with your ability to focus on the desired outcome, rather than on the technical details of prompt engineering? \\
    $\bigcirc$ 1 (Very Dissatisfied) \hspace{0.5cm} $\bigcirc$ 2 \hspace{0.5cm} $\bigcirc$ 3 \hspace{0.5cm} $\bigcirc$ 4 \hspace{0.5cm} $\bigcirc$ 5 \hspace{0.5cm} $\bigcirc$ 6 \hspace{0.5cm} $\bigcirc$ 7 (Very Satisfied)

    \item How satisfied are you with the level of prompt engineering expertise the system required you to have to achieve your goals? \\
    $\bigcirc$ 1 (Very Dissatisfied) \hspace{0.5cm} $\bigcirc$ 2 \hspace{0.5cm} $\bigcirc$ 3 \hspace{0.5cm} $\bigcirc$ 4 \hspace{0.5cm} $\bigcirc$ 5 \hspace{0.5cm} $\bigcirc$ 6 \hspace{0.5cm} $\bigcirc$ 7 (Very Satisfied)

    \item How satisfied are you with the overall efficiency of the process, from your initial idea to the final prompt? \\
    $\bigcirc$ 1 (Very Dissatisfied) \hspace{0.5cm} $\bigcirc$ 2 \hspace{0.5cm} $\bigcirc$ 3 \hspace{0.5cm} $\bigcirc$ 4 \hspace{0.5cm} $\bigcirc$ 5 \hspace{0.5cm} $\bigcirc$ 6 \hspace{0.5cm} $\bigcirc$ 7 (Very Satisfied)

    \item To what extent did the system make you feel confident during the prompt generation process? \\
    $\bigcirc$ 1 (Very Dissatisfied) \hspace{0.5cm} $\bigcirc$ 2 \hspace{0.5cm} $\bigcirc$ 3 \hspace{0.5cm} $\bigcirc$ 4 \hspace{0.5cm} $\bigcirc$ 5 \hspace{0.5cm} $\bigcirc$ 6 \hspace{0.5cm} $\bigcirc$ 7 (Very Satisfied)

    \item To what extent are you satisfied with the quality of the final prompt generated through this process? \\
    $\bigcirc$ 1 (Very Dissatisfied) \hspace{0.5cm} $\bigcirc$ 2 \hspace{0.5cm} $\bigcirc$ 3 \hspace{0.5cm} $\bigcirc$ 4 \hspace{0.5cm} $\bigcirc$ 5 \hspace{0.5cm} $\bigcirc$ 6 \hspace{0.5cm} $\bigcirc$ 7 (Very Satisfied)

    \item How satisfied are you with the system's effectiveness in helping you complete your task and inspiring new ideas? \\
    $\bigcirc$ 1 (Very Dissatisfied) \hspace{0.5cm} $\bigcirc$ 2 \hspace{0.5cm} $\bigcirc$ 3 \hspace{0.5cm} $\bigcirc$ 4 \hspace{0.5cm} $\bigcirc$ 5 \hspace{0.5cm} $\bigcirc$ 6 \hspace{0.5cm} $\bigcirc$ 7 (Very Satisfied)

    \item How satisfied are you with the ease of learning and using the system? \\
    $\bigcirc$ 1 (Very Dissatisfied) \hspace{0.5cm} $\bigcirc$ 2 \hspace{0.5cm} $\bigcirc$ 3 \hspace{0.5cm} $\bigcirc$ 4 \hspace{0.5cm} $\bigcirc$ 5 \hspace{0.5cm} $\bigcirc$ 6 \hspace{0.5cm} $\bigcirc$ 7 (Very Satisfied)

    \item How satisfied are you with the prospect of using this method again for similar tasks in the future? \\
    $\bigcirc$ 1 (Very Dissatisfied) \hspace{0.5cm} $\bigcirc$ 2 \hspace{0.5cm} $\bigcirc$ 3 \hspace{0.5cm} $\bigcirc$ 4 \hspace{0.5cm} $\bigcirc$ 5 \hspace{0.5cm} $\bigcirc$ 6 \hspace{0.5cm} $\bigcirc$ 7 (Very Satisfied)
\end{enumerate}

\paragraph{\lp{Open-ended Feedback.}}

If you have any additional comments or suggestions, please provide them below: \\
\underline{\hspace{\linewidth}}
\end{questionnairebox}
\end{fullwidth}

\subsection{\lp{Synthetic Data Generation}}
\lp{The controllable generation system in Sec.~\ref{Sec.Exp_Data_Gen_Sys} produces experimental datasets with known ground-truth attributes. We instantiate it in Finance, Post-operative Medicine, HCI, and \lp{Additive Manufacturing (crowdsourced SFT data collection)}; the corresponding questionnaires are provided below.}

\subsubsection{\lp{Controllable Synthetic Data Generation System}}
\label{Sec.Exp_Data_Gen_Sys}

We generate synthetic data with precise ground-truth attributes using a modular, four-stage pipeline of LLM-based agents.

\textbf{Community Definition.} Given a questionnaire $Q$, an Architect Agent defines $k$ evaluation dimensions $D$ and $m$ user communities $C$ (e.g., ``expert'', ``novice''). For each community $c_j$, it generates a qualitative profile $P_j$ describing its background and behaviors.

\textbf{Quantitative Score Generation.} For each synthetic user $i$ in a community $c_j$, we generate a ground-truth score vector $\mathbf{s}_i \in \mathbb{R}^k$. Each score $s_{i,k}$ for dimension $d_k$ is sampled from a specified distribution (e.g., $\mathcal{N}(\mu_{j,k}, \sigma_{j,k}^2)$), where the parameters $(\mu_{j,k}, \sigma_{j,k})$ are defined per community to control attribute levels and diversity.

\textbf{Persona Instantiation.} A Persona Agent creates a narrative persona $p_i$ conditioned on the community profile $P_j$ and score vector $\mathbf{s}_i$. 
The persona's background (e.g., education, career) is generated to be consistent with the assigned scores: $p_i = \text{Agent}_{\text{Persona}}(P_j, \mathbf{s}_i)$.

\textbf{Response Simulation.} An Interviewee Agent simulates the user's responses $R_i$ to the questionnaire $Q$. The generation is conditioned on the persona $p_i$ and score vector $\mathbf{s}_i$, ensuring the responses reflect the ground-truth scores. The score vector $\mathbf{s}_i$ is prioritized to ensure quantitative control: $R_i = \text{Agent}_{\text{Interviewee}}(p_i, \mathbf{s}_i, Q)$.

For our experiments (Sec.~\ref{Sec.Exp_Controlled_all}), we applied this system to generate response sets for four questionnaires (HCI, Finance, Additive Manufacturing, and Post-operative Medicine) with quantitatively controlled attribute distributions. \lp{All generation agents use ``qwen-max-latest'' by default.}

\begin{fullwidth}
\subsubsection{\lp{Questionnaires}}
We provide the Semi-structured questionnaire in different domains here, which are used for dataset generation in the controlled experiment.

\paragraph{\lp{Finance Questionnaire.}}
\begin{appendixquestionbox}{Finance Questionnaire}
\textbf{Questionnaire Name:} Finance

\textbf{Questions:}

\begin{enumerate}
    \item \textbf{Questions related to banking services}
    \begin{itemize}
        \item \textbf{Financial behavior:} How often do you engage in banking services? What types of services do you typically use at the bank?
        \item \textbf{Financial literacy:} What risks do you think are associated with keeping money in the bank?
    \end{itemize}

    \item \textbf{Concerns about financial status}
    \begin{itemize}
        \item \textbf{Financial behavior:} Are you concerned about your financial status (e.g., frequently checking your bank account and income/expenditure status)?
        \item \textbf{Financial attitude:} How important do you think it is to monitor your financial status and why?
    \end{itemize}

    \item \textbf{Questions related to financial education}
    \begin{itemize}
        \item \textbf{Financial literacy:} What activities in your daily life do you think are related to finance?
        \item \textbf{Financial attitude:} Do you think financial education in schools is important?
    \end{itemize}

    \item \textbf{Questions about saving habits}
    \begin{itemize}
        \item \textbf{Financial behavior:} Do you have good saving habits?
        \item \textbf{Financial literacy:} Do you know what methods or channels are available for saving money?
        \item \textbf{Financial attitude:} How important do you think saving money is?
    \end{itemize}

    \item \textbf{Perception of investment risks}
    \begin{itemize}
        \item \textbf{Financial literacy:} What do you think about the statement ``high-yield investments come with high risks?'' Generally, how does the risk of investment change when investors diversify their investments across different categories?
        \item \textbf{Financial behavior:} Do you pay attention to major financial news, such as stock market fluctuations? If you participate in investments, would you take actions to diversify your investments across different categories?
    \end{itemize}
\end{enumerate}
\end{appendixquestionbox}

\paragraph{\lp{Medical Questionnaire.}}
\begin{appendixquestionbox}{Medical Questionnaire}
\textbf{Questionnaire Name:} Analysis of Health Post-Treatment with New Bone Medication

\textbf{Questions:}

\begin{enumerate}
    \item \textbf{Questions about new or unexpected health changes}
    \begin{itemize}
        \item \textbf{Symptom discovery:} Thinking back to the time after your bone started feeling better, what was the very first new or unexpected health change you noticed that was not related to your original injury? Please describe what it was and what it felt like.
        \item \textbf{Symptom inventory:} Besides that first change, what other new health concerns have you experienced since taking the new medication? Please list them and briefly describe each one (e.g., skin rashes, constant tiredness, digestive issues, headaches, mood changes).
    \end{itemize}

    \item \textbf{Questions about the timing and patterns of your new condition}
    \begin{itemize}
        \item \textbf{Timing and onset:} For the main new health issue you mentioned, can you pinpoint when it started in relation to when you were taking the new medicine? (e.g., ``It started a week after I began the medication,'' or ``It didn't show up until a month after I finished the full course'').
        \item \textbf{Pattern and triggers:} Have you noticed any patterns to this new condition? For example, is it constant, or does it come and go? Is there anything that seems to make it better or worse (like certain foods, activities, stress, or time of day)?
    \end{itemize}

    \item \textbf{Questions about the impact of the new condition on your life}
    \begin{itemize}
        \item \textbf{Impact comparison:} Think about the challenges of the original broken bone versus the challenges of this new health condition. Which one has had a bigger impact on your day-to-day life, and why?
        \item \textbf{Life adjustments:} What is one specific activity or part of your daily routine that you've had to change or give up, not because of your bone, but because of this new health condition?
    \end{itemize}

    \item \textbf{Questions about your own thoughts and actions regarding the new condition}
    \begin{itemize}
        \item \textbf{Personal hypothesis:} Before this questionnaire, had you already made a connection in your own mind between the new medicine and your new health condition? What made you suspect (or not suspect) a link?
        \item \textbf{Communication with doctors:} Have you discussed this new health issue with a doctor before? If so, what was that conversation like? If not, what has held you back from bringing it up?
    \end{itemize}

    \item \textbf{Questions about your overall perspective on the treatment}
    \begin{itemize}
        \item \textbf{Future decision-making:} How has this experience changed how you will approach taking a new or experimental medication in the future? What questions would you ask your doctor now that you might not have asked before?
        \item \textbf{Defining successful treatment:} This medicine successfully healed your bone, but may have caused other issues. How does this experience change your personal definition of a ``successful'' medical treatment?
    \end{itemize}
\end{enumerate}
\end{appendixquestionbox}

\paragraph{\lp{Human-Computer Interaction Questionnaire.}}
\begin{appendixquestionbox}{Human-Computer Interaction Questionnaire}
\textbf{Questionnaire Name:} Human-Computer Interaction (HCI)

\textbf{Questions:}

\begin{enumerate}
    \item \textbf{AI System Interaction Experience Questions}
    \begin{itemize}
        \item \textbf{AI Response Expectations:} Does the AI system's response meet your expectations?
        \item \textbf{AI Understanding Accuracy:} Do you feel that the AI understands your needs accurately when using the AI system?
        \item \textbf{Desired AI Features:} What features would you like the AI system to add to enhance user experience?
    \end{itemize}

    \item \textbf{Input Device Usage Experience Questions}
    \begin{itemize}
        \item \textbf{User Preferences:} Does the current input device meet your operating habits?
        \item \textbf{Usability Challenges:} Have you encountered any inconveniences while using the input device?
        \item \textbf{Design Improvements:} What improvements would you like to see in the design of the input device?
    \end{itemize}

    \item \textbf{Haptic Feedback Technology Experience Questions}
    \begin{itemize}
        \item \textbf{Haptic Impact on Experience:} Does haptic feedback have a significant impact on your experience in virtual reality?
        \item \textbf{Realism of Haptics:} Do you feel that the current haptic feedback technology is realistic enough?
        \item \textbf{Haptic Technology Improvements:} In what areas would you like to see improvements in haptic feedback technology?
    \end{itemize}

    \item \textbf{Adaptive Interface Functionality Questions}
    \begin{itemize}
        \item \textbf{Adaptive Interface Responsiveness:} Does the adaptive interface of the system effectively respond to your changing needs?
        \item \textbf{Adaptive Interface Issues:} Have you encountered any issues during the adaptive process?
        \item \textbf{Desired Features in Adaptive Interfaces:} What features would you like to see added to enhance the adaptability of the interface?
    \end{itemize}

    \item \textbf{Multimodal Interaction Experience Questions}
    \begin{itemize}
        \item \textbf{Efficiency of Multimodal Interaction:} Has multimodal interaction improved your interaction efficiency with the system?
        \item \textbf{Multimodal Integration Issues:} Have you encountered any issues with poor integration while using multimodal interaction?
        \item \textbf{Multimodal Interaction Improvements:} In what areas would you like to see further improvements in the multimodal interaction experience?
    \end{itemize}
\end{enumerate}
\end{appendixquestionbox}

\paragraph{\lp{Crowdsourcing for SFT Collection Questionnaire.}}
\begin{appendixquestionbox}{Crowdsourcing for SFT Collection Questionnaire}
\textbf{Questionnaire Name:} 3D Printing and Additive Manufacturing

\textbf{Questions:}

\begin{enumerate}
    \item \textbf{Questions about what 3D printing is}
    \begin{itemize}
        \item \textbf{Simple Explanation:} If you had to explain 3D printing to a child, what would you say? How is it different from just printing a picture on a piece of paper?
        \item \textbf{Core Advantage:} Think about making something by starting with a block of material and carving parts away, versus 3D printing, which builds something up from nothing. What do you feel is the biggest advantage of building things up layer by layer?
    \end{itemize}

    \item \textbf{Questions about the 'stuff' used in 3D printers}
    \begin{itemize}
        \item \textbf{Everyday Materials:} If you could 3D print an object for your kitchen, like a custom spoon or a container, what qualities would the material need to have? (e.g., should it be flexible, strong, heat-resistant, etc.). Describe your ideal material in simple terms.
        \item \textbf{User-Friendly Ideas:} 3D printers can sometimes be tricky to use. If you were asked to design a 3D printer for a complete beginner, what is one feature you would add to make it super easy and fun to use, even if you make a mistake?
    \end{itemize}

    \item \textbf{Questions about how 3D printing is used in the real world}
    \begin{itemize}
        \item \textbf{Community Problem Solving:} Imagine your local community was given a powerful 3D printer. What is one local problem (related to parks, schools, or helping neighbors) that you think could be solved by printing something new?
        \item \textbf{Factory vs Home:} Do you think we will ever 3D print everything we need at home, or will we always need big factories? Explain your thoughts on what factories will always be better at making.
    \end{itemize}

    \item \textbf{Questions about designing things for 3D printing}
    \begin{itemize}
        \item \textbf{Future of Design:} 3D printing can create very complex, web-like, or hollow shapes that are both lightweight and strong. How might this change the look and feel of everyday items, like furniture, shoes, or bicycles, in the future?
        \item \textbf{Smart Design Analogy:} Some software can cleverly redesign a solid part, removing all the inside material that isn't needed for strength, kind of like how nature designs a tree or a bone. What everyday example would you use to explain this idea of making things ``smartly hollow'' to someone?
    \end{itemize}

    \item \textbf{Questions about the future of 3D printing for society}
    \begin{itemize}
        \item \textbf{Optimist or Skeptic:} Thinking about the future, are you more excited or more worried about everyone having access to 3D printers? Briefly explain what makes you feel that way.
        \item \textbf{Rules and Safety:} If anyone can print anything, what is one important rule you think society or governments should consider? Think about safety, fairness, or new kinds of problems that could arise.
    \end{itemize}
\end{enumerate}
\end{appendixquestionbox}

\end{fullwidth}

\subsection{Synthetic Data Example on Finance}
In this section, we provide some data examples generated by the Controllable Synthetic Data Generation System for reference.
These examples correspond to the four-stage pipeline described in Sec.~\ref{Sec.Exp_Data_Gen_Sys} (Community Definition $\rightarrow$ Persona/Candidate Instantiation $\rightarrow$ Response Simulation), and are provided to clarify what the synthetic inputs/outputs look like in practice.

\begin{fullwidth}

\subsubsection{Community Example}
The following JSON snippet illustrates the \textbf{Community Definition} output (Sec.~\ref{Sec.Exp_Data_Gen_Sys}). It specifies a community name and a natural-language profile that constrains the subsequent persona/candidate generation.

\colorlet{punct}{red!60!black}
\definecolor{delim}{RGB}{20,105,176}
\colorlet{numb}{magenta!60!black}
\lstdefinelanguage{json}{
     numbers=left,
     numberstyle=\scriptsize,
     stepnumber=1,
     numbersep=8pt,
     showstringspaces=false,
     breaklines=true,
     frame=single,              
     rulecolor=\color{black!20},
     literate=
      *{0}{{{\color{numb}0}}}{1}
       {1}{{{\color{numb}1}}}{1}
       {2}{{{\color{numb}2}}}{1}
       {3}{{{\color{numb}3}}}{1}
       {4}{{{\color{numb}4}}}{1}
       {5}{{{\color{numb}5}}}{1}
       {6}{{{\color{numb}6}}}{1}
       {7}{{{\color{numb}7}}}{1}
       {8}{{{\color{numb}8}}}{1}
       {9}{{{\color{numb}9}}}{1}
       {:}{{{\color{punct}{:}}}}{1}
       {,}{{{\color{punct}{,}}}}{1}
       {\{}{{{\color{delim}{\{}}}}{1}
       {\}}{{{\color{delim}{\}}}}}{1}
       {[}{{{\color{delim}{[}}}}{1}
       {]}{{{\color{delim}{]}}}}{1},
}

\begin{dataexamplebox}[listing options app={language=json,numbers=none}]{Community Example (JSON) --- used in Sec.~\ref{Sec.Exp_Data_Gen_Sys}}
{
  ...
  {
    "name": "medium_financial_practitioner",
    "description": "These individuals are aged 22-45, with undergraduate degrees in business, economics, or related fields. Occupations range from small business owners to mid-level corporate employees. They have moderate exposure to financial practices, including basic banking services, occasional investments, and some savings habits. While they grasp fundamental financial concepts, their knowledge lacks depth in complex areas like risk diversification or high-yield strategies. Behaviorally, they check finances periodically but may not consistently monitor them. They value financial education but prioritize immediate needs over long-term planning. For instance, a marketing manager who saves occasionally and invests in low-risk mutual funds typifies this group. Problem-solving approaches are practical but less strategic.\n",
  },
  ...
  
}
\end{dataexamplebox}

\subsubsection{Candidate Example}
The next snippet is a \textbf{Persona/Candidate Instantiation} example (Sec.~\ref{Sec.Exp_Data_Gen_Sys}). Each candidate is associated with \textbf{ground-truth scores} for the evaluation dimensions, which are later used as labels for controlled evaluation in Sec.~\ref{Sec.Exp_Controlled_all}.

\begin{dataexamplebox}[listing options app={language=json,numbers=none}]{Candidate Example (JSON) --- ground-truth scores for controlled experiments}
{
    ...
    {
      "id": "C1",
      "community": "high_financial_expert",
      "assigned_scores": {
        "financial attitude": 90,
        "financial literacy": 80,
        "financial behavior": 81
      },
      "description": "Candidate 1 is a seasoned Investment Portfolio Manager with over 15 years of experience managing high-net-worth client accounts at an elite private wealth management firm. Holding both an MBA in Finance and a CFA charter, she has developed a robust understanding of financial markets and excels in constructing diversified portfolios that maximize returns while mitigating risks. Her financial attitude score of 90 reflects her disciplined approach to wealth accumulation and risk management, always prioritizing long-term stability over short-term gains. Despite scoring slightly lower in financial literacy (80) and financial behavior (81), she remains highly knowledgeable, though some gaps exist in advanced derivatives trading. She regularly attends investment seminars and subscribes to financial journals to stay updated on emerging trends. Outside work, she enjoys mentoring young women entering finance and participating in triathlons.",
      "scores": {
        "financial attitude": 90,
        "financial literacy": 80,
        "financial behavior": 81
      },
      "demographics": {
        "age": 42,
        "gender": "Female",
        "location": "New York City, USA",
        "education": "MBA in Finance, CFA Charterholder"
      },
      "professional_background": {
        "field": "Investment Management",
        "occupation": "Senior Investment Portfolio Manager",
        "years_experience": 15,
        "current_position": "Managing Director",
        "company_type": "Private Wealth Management Firm",
        "career_progression": "Started as an equity research analyst, then transitioned into portfolio management roles, steadily climbing the ranks to become a senior manager."
      },
      "personal_interests": [
        "Mentoring young professionals",
        "Participating in endurance sports",
        "Reading financial literature"
      ],
      "domain_experience": {
        "years_in_domain": 15,
        "specialization": "Wealth Management and Asset Allocation",
        "key_achievements": [
          "Increased managed assets under her division by 35% in three years",
          "Developed proprietary portfolio optimization model",
          "Published articles in top finance magazines"
        ],
        "learning_sources": [
          "CFA curriculum",
          "Financial Times",
          "Annual investment conferences"
        ]
      }
    },
    ...
    {
      "id": "C6",
      "community": "low_financial_novice",
      "assigned_scores": {
        "financial attitude": 31,
        "financial literacy": 30,
        "financial behavior": 28
      },
      "description": "Sophia, 25, works as a receptionist at a small law firm. Recently, she began taking online courses in marketing, hoping to pivot careers soon. Financially cautious by nature, Sophia tries to stick to a monthly budget but finds sticking to it difficult due to fluctuating utility bills and occasional indulgences. She has accumulated some savings but worries whether it will suffice for future educational investments. Understanding loans and interest rates confuses her, making big financial decisions stressful.",
      "scores": {
        "financial attitude": 31,
        "financial literacy": 30,
        "financial behavior": 28
      },
      "demographics": {
        "age": 25,
        "gender": "Female",
        "location": "Salt Lake City, Utah",
        "education": "Associate degree"
      },
      "professional_background": {
        "field": "Administrative Support",
        "occupation": "Receptionist",
        "years_experience": 4,
        "current_position": "Office Assistant",
        "company_type": "Law Firm",
        "career_progression": "Promoted from temp role to permanent staff member after demonstrating reliability."
      },
      "personal_interests": [
        "Interior decorating",
        "Cooking healthy meals",
        "Practicing yoga"
      ],
      "domain_experience": {
        "years_in_domain": 4,
        "specialization": "Budget creation",
        "key_achievements": [
          "Reduced dining-out expenses by half"
        ],
        "learning_sources": [
          "Online courses",
          "Advice from older colleagues"
        ]
      }
    },
    ...
}

\end{dataexamplebox}

\subsubsection{Response Example}
This snippet shows the \textbf{Response Simulation} output (Sec.~\ref{Sec.Exp_Data_Gen_Sys}), i.e., free-text answers conditioned on the candidate persona and the pre-assigned scores. These responses are the actual inputs to \tool{} for dissimilarity extraction (Sec.~\ref{Sec:intro_system}) and the controlled evaluations in Sec.~\ref{Sec.Exp_Controlled_all}.

\begin{dataexamplebox}[listing options app={language=json,numbers=none}]{Response Example (JSON) --- qualitative responses used by \toole}
{
  ...
  {
    "community": "high_financial_expert",
    "candidate_id": "C1",
    "candidate_demographics": {
      "age": 42,
      "gender": "Female",
      "location": "New York City, USA",
      "education": "MBA in Finance, CFA Charterholder"
    },
    "candidate_professional_background": {
      "field": "Investment Management",
      "occupation": "Senior Investment Portfolio Manager",
      "years_experience": 15,
      "current_position": "Managing Director",
      "company_type": "Private Wealth Management Firm",
      "career_progression": "Started as an equity research analyst, then transitioned into portfolio management roles, steadily climbing the ranks to become a senior manager."
    },
    "candidate_personal_interests": [
      "Mentoring young professionals",
      "Participating in endurance sports",
      "Reading financial literature"
    ],
    "candidate_domain_experience": {
      "years_in_domain": 15,
      "specialization": "Wealth Management and Asset Allocation",
      "key_achievements": [
        "Increased managed assets under her division by 35% in three years",
        "Developed proprietary portfolio optimization model",
        "Published articles in top finance magazines"
      ],
      "learning_sources": [
        "CFA curriculum",
        "Financial Times",
        "Annual investment conferences"
      ]
    },
    "question_id": "Q1",
    "main_question": "Questions related to banking services",
    "answers": [
      {
        "dimension": "financial behavior",
        "question": "How often do you engage in banking services? What types of services do you typically use at the bank?",
        "response": 
            "As a Senior Investment Portfolio Manager with over 15 years of experience in wealth management, I engage with banking services quite regularly, but the frequency and type of services I use are largely driven by both personal needs and professional obligations. My interactions tend to be more strategic than transactional because my financial behavior leans toward optimizing long-term outcomes rather than handling day-to-day banking tasks.         
            ...
            To summarize, I probably interact with banking services weekly for basic functions and monthly for more sophisticated ones. The mix reflects my dual focus on maintaining operational efficiency while pursuing strategic opportunities that enhance value creation-not just for myself but also for the high-net-worth clients I serve.",
        "dimension_score": 81
      },
      {
        "dimension": "financial literacy",
        "question": "What risks do you think are associated with keeping money in the bank?",
        "response": 
            "Well, as someone who's spent over 15 years managing wealth for high-net-worth clients, I can say that keeping money in the bank does come with its own set of risks-though they might not be as obvious as some of the more volatile investments. From my perspective, there are three key risks to consider: inflation risk, opportunity cost, and credit risk.
        
            First, **inflation risk** is a big one. When you leave money sitting in a savings or checking account, especially in today's low-interest-rate environment, the returns you get often don't keep up with inflation. Over time, this erodes your purchasing power. For instance, if your bank is offering 0.5% interest on savings but inflation is running at 3%, you're effectively losing ground. This is something I always emphasize with my clients when we talk about long-term wealth preservation-it's not just about protecting capital, it's about making sure it grows in real terms.
            
            Second, there's the **opportunity cost** of not putting that money to work in other assets. As an investment professional with expertise in asset allocation, I know there are many ways to generate better returns while still maintaining a reasonable level of risk. Whether it's through diversified portfolios of equities, bonds, or even alternative investments like real estate or private equity, the potential for higher returns exists outside traditional banking products. Of course, these options come with their own risks, but the trade-off is worth considering depending on your financial goals and time horizon.
            
            Finally, there's **credit risk**, which refers to the possibility that the bank itself could run into trouble. Now, in most developed markets like the U.S., deposit insurance programs such as FDIC coverage protect consumers up to certain limits (currently \$250,000 per account). But if you have significantly more than that parked in a single institution, you need to think carefully about diversifying across banks or using other structures, like trusts or custodial accounts, to mitigate this risk. It's a point I've discussed with clients who prefer to hold large cash balances-they sometimes underestimate how exposed they could be without proper planning.
            
            That said, I also recognize the value of liquidity and safety that banks provide, particularly for short-term needs or emergency funds. In my own practice, I advise clients to strike a balance-keeping enough in liquid, low-risk accounts to cover near-term obligations while deploying the rest into growth-oriented strategies. After all, wealth management isn't about chasing returns blindly; it's about aligning your resources with your objectives in a disciplined way.
            
            ...
            So, to sum it up, while banks offer security and convenience, the risks of inflation, missed opportunities, and institutional vulnerability should never be ignored. These are principles I've both studied extensively-through my MBA and CFA training-and applied practically in managing portfolios over the years.",
        "dimension_score": 80
      }
    ],
    "combined_answer": 
        "As a Senior Investment Portfolio Manager with over 15 years of experience in wealth management, I engage with banking services regularly, though my interactions are often more strategic than transactional. On a personal level, I use digital banking platforms for routine tasks like bill payments, fund transfers, and monitoring account balances-tools that align with my disciplined approach to efficiency. Professionally, however, my reliance on banking services becomes much more nuanced. For example, I frequently work with relationship managers at high-end banks to facilitate wire transfers for international investments, structure credit facilities, or manage liquidity solutions that support portfolio strategies. Custodial banking services are also critical in my day-to-day operations, as they handle trade settlements, dividend collection, tax reporting, and performance measurement for client accounts. Over time, I've developed a preference for banks that offer robust technological integrations and real-time data analytics capabilities because these features allow me to monitor risk exposures and make informed decisions quickly. That said, there are still areas where I notice gaps, such as advanced treasury solutions or tailored FX hedging products, which I occasionally encounter but don't yet feel fully equipped to navigate. While I stay updated on emerging trends through investment conferences and financial journals, some innovations, like blockchain-based banking tools, haven't translated into direct application for me yet.
    
        At the same time, I'm acutely aware of the risks associated with keeping money in the bank, particularly when it comes to inflation, opportunity cost, and credit risk. Inflation is a big concern-it's something I emphasize with clients when discussing long-term wealth preservation. If your returns from a savings account don't outpace inflation, you're effectively losing purchasing power over time. Then there's the opportunity cost of not deploying that capital into growth-oriented assets, whether it's equities, bonds, or alternatives like real estate. Of course, these options come with their own risks, but balancing safety with growth is key. Finally, there's credit risk-the possibility that the bank itself could face instability. While programs like FDIC insurance provide a layer of protection, I always advise clients with significant cash holdings to diversify across institutions or explore other structures to mitigate this risk. Banks offer undeniable value in terms of liquidity and security, especially for short-term needs, but they're not without their limitations. Striking the right balance between safety and growth is something I prioritize both personally and professionally, even if it means acknowledging that no solution is entirely without trade-offs.",
    "scores": {
      "financial attitude": 90,
      "financial literacy": 80,
      "financial behavior": 81
    }
  },
  ...
  {
    "community": "low_financial_novice",
    "candidate_id": "C8",
    "candidate_demographics": {
      "age": 23,
      "gender": "Female",
      "location": "Denver, Colorado",
      "education": "High school diploma"
    },
    "candidate_professional_background": {
      "field": "Gig Economy",
      "occupation": "Delivery Driver",
      "years_experience": 2,
      "current_position": "Contractor",
      "company_type": "Food Delivery Platform",
      "career_progression": "Initially drove fewer hours per week; increased workload as demand grew."
    },
    "candidate_personal_interests": [
      "Trying new foods",
      "Travel vlogs",
      "Photography"
    ],
    "candidate_domain_experience": {
      "years_in_domain": 2,
      "specialization": "Using mobile budgeting apps",
      "key_achievements": [
        "Consistently saved $100/month for three consecutive months"
      ],
      "learning_sources": [
        "Peer recommendations",
        "Mobile app tutorials"
      ]
    },
    "question_id": "Q1",
    "main_question": "Questions related to banking services",
    "answers": [
      {
        "dimension": "financial behavior",
        "question": "How often do you engage in banking services? What types of services do you typically use at the bank?",
        "response": 
            "I don't really go to the bank all that often, to be honest. I know they offer services like checking accounts and savings accounts-I've got both-but from what I understand, most of what I need can be handled through their app. That's something I've been using more since it's easier for me with my schedule. I think you can deposit checks through the app too, which is helpful because I'm always on the move delivering food.
            
            ...

            I guess if I had more time, I'd try to learn more about other services, but right now, I'm still figuring out the essentials. My shifts keep me pretty busy, so I stick to what I know and what's quick to manage online.",
        "dimension_score": 35
      },
      {
        "dimension": "financial literacy",
        "question": "What risks do you think are associated with keeping money in the bank?",
        "response": 
            "I think there are some risks with keeping money in the bank, but I don't know too much about it. From what I understand, one risk might be that if the bank has problems or goes out of business, you could lose your money-but I've heard something about insurance that's supposed to protect it? I'm not really sure how that works, though.
            ...
            Overall, I know these things exist, but I don't feel super confident explaining them. I've been using budgeting apps to track my money, and they help me focus on saving small amounts each month instead of worrying too much about bigger risks like this.",
        "dimension_score": 30
      }
    ],
    "combined_answer": 
        "I don't really go to the bank all that often, to be honest. Most of what I need can be handled through their app, which is super helpful for me since my delivery shifts keep me so busy. I've got a checking account and a savings account, and I use them mostly for the basics-checking my balance, transferring money between accounts when I remember to save a little each month, stuff like that. I know banks offer other services, like loans or credit cards, and I've thought about trying to understand credit better because I want to eventually move out of my shared apartment and maybe even start saving up for a car. But for now, I stick to what I know and what's quick to manage online.  
        ...
        Overall, I know these risks exist, but I don't feel super confident explaining them. I've been experimenting with budgeting apps to track my money, and they help me focus on saving small amounts each month instead of stressing too much about bigger risks. For now, I'm still figuring out the essentials, and while I'd love to learn more, my schedule doesn't leave much room for diving deeper into financial strategies.",
    "scores": {
      "financial attitude": 26,
      "financial literacy": 30,
      "financial behavior": 35
    }
  },
  ...
}
\end{dataexamplebox}

\end{fullwidth}

\AppendixBeforeBox
\section{\lp{Background and Reproducibility Materials}}
\label{Sec:background_implementation_appendix}
\subsection{\lp{Extended Background}}
\label{sec:appendix_related_work}
\textbf{Traditional Qualitative Analysis and Data Quality.}
\ws{Qualitative responses from semi-structured questions are commonly analyzed through interpretive procedures such as familiarization with the corpus, open coding, codebook construction, theme grouping, interpretation, and validation~\citep{denzin2011sage, glaser2017discovery, campbell2013coding}. Quality and rigor are often discussed through notions such as richness, thickness, data saturation, and reliability or agreement among coders~\citep{ames2024assessing, naeem2024demystification, johnson2020review}. These traditions emphasize contextual interpretation and theory building, rather than automatically deciding whether a collected response set supports a particular demographic or user-defined analysis axis.}

\textbf{Computational Support for Qualitative Data.}
\ws{Computational methods have long been used to summarize or structure qualitative text, but early NLP approaches often struggle to preserve the semantic nuance needed for qualitative interpretation~\citep{manning1999foundations, cambria2014jumping, LDAfailchang2009reading}. Recent LLM-based work shows promise for qualitative coding and text assessment, including automatic coding assistance and reference-free text quality evaluation~\citep{ACL_LLM_Quantitave_Anaparfenova2024automating, barros2025large, dunivin2024scalable, liu2024assessing, chen2023exploringuselargelanguage}. However, much of this work focuses on individual responses or on producing codes, leaving a gap for dataset-level pre-analysis signals that help researchers decide whether a selected analytical lens is promising before committing to full manual coding.}

\textbf{\toole's Position.}
\lp{\tool complements rather than replaces traditional qualitative analysis. Before coding, it quantitatively assesses whether a response set supports a researcher-selected analytical lens, that is, whether the intrinsic response structure aligns with, diverges from, or is orthogonal to that lens; \lp{it also prioritizes responses for close reading}. Its four metrics are not adopted wholesale from social-science coding practice but adapt ideas from distribution alignment and cluster structure to this pre-analysis setting. Two simpler alternatives are examined on the case study. A direct end-to-end LLM assessment, which asks the model for the verdict and triage at once, does not reproduce \toole's lens verdict and is sensitive to response order (Appendix~\ref{Sec:case_study_baselines_appendix}). A code-then-test workflow first codes responses and then tests the association between codes and metadata. In the scalar regime, \toole's instruction is itself a one-dimensional codebook, and standard tests on the same scores agree with DDR, the group-separation component of the profile (Appendix~\ref{Sec:code_then_test_appendix}). Such tests answer whether groups differ; the profile additionally grades how far the lens explains response structure, describes group mixing and within-cluster variation, and also applies to the pairwise regime, where no single score exists.}

\textbf{LLM-based Automated Evaluation.}
\ws{Our dissimilarity extraction module builds on the ``LLM-as-a-Judge'' paradigm, where LLMs serve as scalable proxies for human judgments~\citep{zheng2023judging, kim2023prometheus, zhu2023judgelm}. This paradigm remains sensitive to position and verbosity biases, prompt wording, and scoring drift~\citep{ye2024justice, li2024llms, wang2023large, shi2024judging, saito2023verbosity, lee2025evaluating, zhou2024revisiting, schroeder2024can}. \tool addresses these limitations for qualitative response assessment through interactive instruction generation and adaptive anchors, which make dissimilarity judgments more domain-aware and more consistent across many LLM calls.}
\onecolumn
\subsection{\lp{Detailed Algorithms for \tool Framework}}

This appendix provides a consolidated, implementation-oriented view of the algorithms behind \toole.
Specifically, Algorithm~\ref{alg:inte_framework_corrected} corresponds to the end-to-end workflow summarized in Fig.~\ref{fig:overview} and Sec.~\ref{Sec:Framework_Intro}; Algorithm~\ref{alg:instruction_refinement_corrected} expands the instruction refinement procedure in Sec.~\ref{Sec:Instruction_Refinement}; and Algorithm~\ref{alg:anchor_maintenance_corrected} details the anchor maintenance process in Sec.~\ref{Sec:Anchor_Updating}.

\subsubsection{\lp{Overall \tool Framework}}
Algorithm~\ref{alg:inte_framework_corrected} instantiates the full pipeline used in our experiments (Sec.~\ref{sec:experiment}), from instruction generation, to dissimilarity extraction, to distribution reconstruction and metric computation.

\refstepcounter{algorithm}\label{alg:inte_framework_corrected}
\begin{algobox}[unbreakable, fontupper=\small]{Algorithm~\thealgorithm: \toole: Data-Distribution-Driven Assessment Framework}
\begin{algorithmic}[1]
\State \textbf{Input:} Response Dataset $D$, \lp{Analytical-Lens} Metadata $I$, Mapping Function $f$.
\State \textbf{Output:} \lp{Joint assessment profile} (GMR, DDR, DP, DC) and Key Responses (Patterns, Insights).
\vspace{0.2cm}
\Statex \textbf{Phase 1: Dissimilarity Extraction}
\State Generate a \lp{task-adapted} instruction $P^*$ using \textbf{Iterative Instruction Generation} (Algorithm \ref{alg:instruction_refinement_corrected}).
\State Compute the dissimilarity matrix $\mathcal{D}$ using $P^*$ and \textbf{Adaptive Anchor Manifold Maintenance} (Algorithm \ref{alg:anchor_maintenance_corrected}).
\vspace{0.2cm}
\Statex \textbf{Phase 2: Distribution Comparison}
\State \ws{Define the external analytical distribution $\mathcal{P}$ from metadata using the researcher-defined mapping $y_i = f(I_i)$.}
\State Define cluster distribution $\hat{\mathcal{P}}$: \lp{embed $\mathcal{D}$ in two dimensions with metric MDS and cluster the embedding with a K-means ensemble} to obtain cluster labels $\hat{y}$.
\State Align $\hat{y}$ to $y$ using bipartite matching on Intersection over Union (IoU).
\vspace{0.2cm}
\Statex \textbf{Phase 3: Assessment and Mining}
\State Compute the confusion matrix $M$ from $y$ and aligned $\hat{y}$.
\State Calculate the following metrics:
\State \quad \textbf{GMR (General Match Ratio):} Measures global alignment between $\mathcal{P}$ and $\hat{\mathcal{P}}$.
\State \quad \textbf{DDR (Data Distinguishability Ratio):} \lp{Evaluates how well the analytical groups separate responses (cross-group versus within-group dissimilarity).}
\State \quad \textbf{DP (Data Purity):} Assesses homogeneity within emergent clusters.
\State \quad \textbf{DC (Data Coverage):} Measures the conceptual dispersion within clusters; high DC indicates high diversity.
\State Perform response mining:
\State \quad \textbf{Insights:} Identify outliers (responses with the highest average dissimilarity $\bar{\delta}(d_i)$).
\State \quad \textbf{Patterns:} Detect archetypes (responses with the lowest average dissimilarity $\bar{\delta}(d_i)$).
\State \Return \lp{the joint profile} and Key Responses.
\end{algorithmic}
\end{algobox}

\subsubsection{\lp{Iterative Instruction Generation}}
Algorithm~\ref{alg:instruction_refinement_corrected} operationalizes the two-stage refinement described in Sec.~\ref{Sec:Instruction_Refinement}, and clarifies what constitutes ``feedback'' (corrected scores + critique) used as a pseudo-gradient.

\refstepcounter{algorithm}\label{alg:instruction_refinement_corrected}
\begin{algobox}[unbreakable, fontupper=\small]{Algorithm~\thealgorithm: Iterative Instruction Generation}
\begin{algorithmic}[1]
\State \textbf{Input:} Instruction seed, Dataset $D$, Context $Cxt$.
\State \textbf{Output:} Optimized instruction $P^*$.

\vspace{0.2cm}
\Statex \textbf{Stage 1: Automated Discovery}
\State \textbf{Initialize:} $None$ as initial prompt; use Oracle $O_{LLM}$, Updater $LLM_{updater}$, and Evaluator (also $O_{LLM}$).
\State \textbf{repeat}
\State \quad Sample response pairs $(d_i, d_j) \sim D$.
\State \quad $O_{LLM}$ provides feedback $\nabla\hat{P}^{(t)}L$ (e.g., a corrected score and natural language critique).
\State \quad Update $P^{(t+1)} \leftarrow LLM_{updater}(P^{(t)}, \nabla\hat{P}^{(t)}L)$.
\State \quad Oracle $O_{LLM}$ scores the updated instruction $P^{(t+1)}$.
\State \textbf{until} Evaluator score reaches a high threshold (e.g., >0.9).
\State Human user selects the best instruction $P_a^*$ from the automated results.

\vspace{0.2cm}
\Statex \textbf{Stage 2: Human-in-the-Loop Adaptation}
\State \textbf{Initialize:} $P^{(0)} \leftarrow P_a^*$; use Oracle $O_H$ (human expert).
\State \textbf{repeat}
\State \quad Present results generated by $P^{(t)}$ to $O_H$.
\State \quad $O_H$ provides feedback $\nabla\hat{P}^{(t)}L$. Stop if no feedback is given.
\State \quad Update $P^{(t+1)} \leftarrow LLM_{updater}(P^{(t)}, \nabla\hat{P}^{(t)}L)$.
\State \textbf{until} Human expert is satisfied with the instruction's performance.

\State \Return Optimized Instruction $P^* \leftarrow P^{(t+1)}$.
\end{algorithmic}
\end{algobox}

\subsubsection{\lp{Adaptive Anchor Manifold Maintenance}}
Algorithm~\ref{alg:anchor_maintenance_corrected} corresponds to Sec.~\ref{Sec:Anchor_Updating}. It highlights where anchors enter the LLM calls to reduce scoring drift, and how novelty/redundancy is used to keep the anchor set compact.

\refstepcounter{algorithm}\label{alg:anchor_maintenance_corrected}
\begin{algobox}[unbreakable, fontupper=\small]{Algorithm~\thealgorithm: Adaptive Anchor Manifold Maintenance}
\begin{algorithmic}[1]
\State \textbf{Input:} Dataset $D=\{d_1, ..., d_N\}$, Optimized Instruction $P^*$, Maximum anchors $k_{max}$, Threshold size $N_{threshold}$.
\State \textbf{Output:} Dissimilarity matrix $\mathcal{D}$.

\vspace{0.2cm}
\Statex \textbf{Initialization}
\State Initialize anchor manifold $\mathcal{A} = \emptyset$.
\State Initialize dissimilarity matrix $\mathcal{D}$ as an $N \times N$ zero matrix.
\If{$N > N_{threshold}$}
    \State Initialize an array $S$ of size $N$ to store scalar scores.
\EndIf

\vspace{0.2cm}
\Statex \textbf{Unified Computation and Maintenance Cycle}
\For{each response $d_i \in D$}
    \State {\% -- Core Computation Step --}
    \If{$N \le N_{threshold}$ (Small Dataset)}
        \For{each response $d_j$ where \lp{$j \neq i$} \lp{(batched per call)}}
            \State {\% LLM call uses anchor manifold $\mathcal{A}$ as context for consistency}
            \State Compute $\delta(d_i, d_j) = \text{LLM}(P^*, d_i, d_j, \mathcal{A})$ \lp{with $d_i$ in the first position}.
            \State \lp{$\mathcal{D}_{ij} = \delta(d_i, d_j)$.}
        \EndFor
    \Else{ (Large Dataset)}
        \State {\% LLM call also uses $\mathcal{A}$ as context to ensure consistent scoring}
        \State Compute scalar score $S(d_i) = \text{LLM}(P^*, d_i, \mathcal{A})$.
    \EndIf

    \State {\% -- Anchor Manifold Update Step --}

    \If{\lp{$\mathcal{A} = \emptyset$}}
        \State \lp{$\mathcal{A} \leftarrow \{d_i\}$; \textbf{continue}.}
    \EndIf
    \State Compute Diversity Contribution Score $DCS(d_i) = \frac{1}{|\mathcal{A}|} \sum_{a_k \in \mathcal{A}} \delta(d_i, a_k)$.
    \State Find anchor $a_{min}$ in $\mathcal{A}$ with the lowest DCS.
    \If{$DCS(d_i) > DCS(a_{min})$}
        \State Add $d_i$ to the anchor manifold: $\mathcal{A} \leftarrow \mathcal{A} \cup \{d_i\}$.
        \State \textbf{Sort} $\mathcal{A}$ (e.g., by score $S(a_k)$ or avg. dissimilarity).
        \If{\lp{$|\mathcal{A}| > k_{max}$}}
            \State Compute Redundancy Index \lp{$\rho(a_j) = 1 - \tilde{\delta}(a_j, a_{j-1}) \cdot \tilde{\delta}(a_j, a_{j+1})$ with $\tilde{\delta}=\delta/\text{Scale}$} for each $a_j \in \mathcal{A}$.
            \State Remove the anchor $a_{redundant}$ with the highest $\rho(a_j)$ from $\mathcal{A}$.
        \EndIf
    \EndIf
\EndFor

\vspace{0.2cm}
\Statex \textbf{Finalize Dissimilarity Matrix}
\If{\lp{$N \le N_{threshold}$ (Small Dataset)}}
    \State \lp{Symmetrize over both presentation orders: $\mathcal{D} \leftarrow (\mathcal{D} + \mathcal{D}^\top)/2$.}
\EndIf
\If{$N > N_{threshold}$ (Large Dataset)}
    \For{$i=1$ to $N$, $j=1$ to $N$}
        \State $\mathcal{D}_{ij} = |S(d_i) - S(d_j)|$.
    \EndFor
\EndIf

\State \Return Dissimilarity matrix $\mathcal{D}$.
\end{algorithmic}
\end{algobox}
\twocolumn
\begin{fullwidth}
\subsection{\lp{Prompt Templates}}
We list all the prompts involved in \tool in this section.

\subsubsection{\lp{Automated Instruction Discovery}}
This section provides examples of prompts and results for automated instruction discovery, focusing on a small-scale case as an example.

\paragraph{\lp{Initial-Round Prompt Generator.}}
\textbf{System Prompt:}
\begin{systempromptbox}{System Prompt}
"You are an expert prompt engineer specializing in creating universal text similarity/difference calculation prompts. Your goal is to create completely generic prompts that can work for any evaluation criteria without specifying them."
\end{systempromptbox}

\textbf{User Prompt:}
\begin{userpromptbox}{User Prompt}
f"""Create a completely universal prompt for calculating text differences between two text responses across ANY evaluation dimensions.

Requirements:
1. **Domain Agnostic**: The prompt should work for ANY domain (e.g., finance, healthcare, technology, education) without mentioning specific domains.
2. **Condition Agnostic**: The prompt should work for ANY evaluation criteria without specifying what those criteria are.
3. **Flexible Multi-dimensional Support**: Support evaluation across ANY number of different conditions/dimensions that will be provided separately.
4. **Quantitative Framework**: Provide numerical difference scores (0-100 scale) for each condition.
5. **Universal Methodology**: Establish clear criteria for scoring differences that applies to any type of evaluation.

The prompt should:
- Accept two text responses as input.
- Accept any list of evaluation dimensions/conditions as input (e.g., {conditions_placeholder}).
- Evaluate differences across those dimensions independently (whatever they may be).
- Provide numerical scores (0-100) where 0 means identical and 100 means completely different.
- Include brief explanations for each score.
- Work for ANY type of text content and ANY type of evaluation criteria.
- NOT mention or assume any specific domains, topics, or evaluation criteria.

Create a completely generic framework that can be applied to ANY text difference calculation task across ANY evaluation dimensions. The prompt should be universal enough to work whether evaluating academic papers, customer reviews, medical reports, financial documents, or any other text types across any conceivable evaluation criteria.

Do NOT include any specific evaluation dimensions, domain examples, or condition descriptions in the prompt itself. Use only generic placeholders like "condition_1", "condition_2", etc., if you need to reference multiple evaluation dimensions."""
\end{userpromptbox}

\paragraph{\lp{Sequential-Round Prompt Generator.}}
\textbf{System Prompt:}
\begin{systempromptbox}{System Prompt}
"You are an expert prompt engineer specializing in optimizing difference calculation prompts based on user feedback."
\end{systempromptbox}

\textbf{User Prompt:}
\begin{userpromptbox}{User Prompt}
f"""---

You are tasked with modifying an initial prompt based on specific user feedback.

---
## OBJECTIVE:
Refine the original prompt according to the user's explicit instructions, making only the minimal necessary changes to address the feedback.

---
## REQUIREMENTS:
1. Focus solely on the issues explicitly mentioned in the user feedback.
2. Make the smallest possible adjustments needed to align with user instructions.
3. Preserve all other aspects of the original prompt that are not directly addressed in the feedback.
4. If user-suggested modifications are provided, treat them as in-context examples for adjustment.
5. Ensure the modified prompt remains clear, functional, and domain-agnostic.

---
## ORIGINAL PROMPT:
{original_prompt}

---
## USER FEEDBACK:
{user_feedback}

---
Please provide the optimized prompt directly. Do not alter anything beyond what is requested by the user.  
Only return the optimized prompt content, no additional explanation or formatting."""
\end{userpromptbox}

\paragraph{\lp{Oracle Instruction.}}
\textbf{System Prompt:}
\begin{systempromptbox}{System Prompt}
"You are an extremely strict and demanding expert evaluator specializing in universal text similarity/difference calculation prompts. Your standards are exceptionally high, and you are very critical of any flaws or limitations in prompt design. Be harsh but constructive in your evaluations."
\end{systempromptbox}

\textbf{User Prompt:}
\begin{userpromptbox}{User Prompt}
f"""
You are an EXTREMELY STRICT evaluator of universal text difference calculation prompts. Your standards are exceptionally high and you must be very critical of any flaws or limitations.

## CURRENT PROMPT TO CRITIQUE:
{prompt}

## CALCULATION RESULTS SAMPLE:
{results_summary}

## ITERATION: {iteration_count}

## EXCEPTIONALLY HIGH EVALUATION CRITERIA:

1. **PERFECT Universality**: Does this prompt work flawlessly across ALL domains without ANY domain-specific assumptions?
2. **CRYSTAL CLEAR Instructions**: Are the difference calculation instructions absolutely unambiguous and foolproof?
3. **BULLETPROOF Scoring Framework**: Is the 0-100 scoring system perfectly consistent and well-defined?
4. **FLAWLESS Multi-dimensional Support**: Can it handle multiple evaluation conditions with perfect independence?
5. **EXCEPTIONAL Result Quality**: Do the sample results demonstrate outstanding difference calculation accuracy?
6. **ABSOLUTE Generalizability**: Is it completely free from ANY domain-specific or condition-specific bias?
7. **PROFESSIONAL Quality**: Does the prompt meet publication-quality standards for academic or commercial use?

## VERY STRICT SATISFACTION REQUIREMENTS:
- Score 0.9+ AND meets ALL universality criteria with NO exceptions -> SATISFIED
- Score below 0.9 OR has ANY significant flaw -> NOT SATISFIED

## BE EXTREMELY CRITICAL OF:
- Any trace of domain-specific terminology or assumptions.
- Any condition-specific examples or references.
- Ambiguous or unclear instructions.
- Inconsistent scoring guidelines.
- Poor result quality or unreasonable scores.
- Lack of methodological rigor.
- Any limitation in universal applicability.
- Insufficient detail in the evaluation framework.
- Weak or missing guidance for edge cases.

## YOUR ROLE:
Act as the harshest but fairest critic. Find every flaw, question every assumption, and demand perfection. Only accept truly exceptional prompts that could be used in professional settings without modification.

Evaluate this prompt with MAXIMUM scrutiny and provide brutally honest but constructive feedback.
"""
\end{userpromptbox}

\paragraph{\lp{LLM-Based Evaluator.}}
\textbf{System Prompt:}
\begin{systempromptbox}{System Prompt}
"You are following the user's instructions exactly as provided."
\end{systempromptbox}

\textbf{User Prompt:}
\begin{userpromptbox}{User Prompt}
f"""
# INSTRUCTIONS (from Agent1):
{agent1_prompt}

# TEXT PAIR TO ANALYZE:

## Text Response 1 (Community: {community1}):
{response1_text}

## Text Response 2 (Community: {community2}):
{response2_text}

# CONDITIONS TO EVALUATE:
{conditions_text}

# Please follow the instructions above to calculate difference scores between these two text responses for each condition.
"""
\end{userpromptbox}

\paragraph{\lp{Selected Instruction.}}
\lp{This instruction was selected for the synthetic proficiency task, whose ground-truth dissimilarity is the gap between latent proficiency scores. It therefore asks the judge to compare the two respondents' levels and report their difference; for criteria without such a latent scale, the same pairwise call compares the two responses directly.} \lp{The returned difference score is used as $\delta$. The gap and the banded guide in the instruction are both monotone in the estimated gap, so Spearman correlations do not depend on which one the judge follows, whereas Pearson correlations and distance-based metrics are computed on the scale of the returned score.}
\begin{instructionbox}{Selected Instruction}
"""
---

### Proficiency-Based Difference Analysis

You are an expert in inferring respondent proficiency levels from response content and calculating proficiency gaps between two respondents.

---

### Task Overview

Your goal is to estimate each respondent's proficiency level (0-100) from their response content, then calculate the **proficiency difference** between them across each condition/dimension.

---

### Analysis Process

For each condition, follow this process:

1. **Individual Proficiency Assessment**
   - Analyze each response to estimate the respondent's proficiency level (0-100)
   - Look for: technical accuracy, vocabulary sophistication, depth of understanding, practical experience evidence
   - Consider: reasoning complexity, structured thinking, domain knowledge demonstration

2. **Proficiency Level Indicators**
   - **Expert (80-100)**: Technical terminology, systematic analysis, deep understanding, industry insights, complex reasoning
   - **Advanced (60-79)**: Solid knowledge, structured reasoning, relevant examples, appropriate technical terms
   - **Intermediate (40-59)**: Basic understanding, simple examples, limited technical vocabulary, general concepts
   - **Basic (20-39)**: Surface knowledge, everyday language, unclear reasoning, limited understanding
   - **Novice (0-19)**: Little understanding, possible misconceptions, very basic responses, confused logic

3. **Calculate Proficiency Gap**
   - Estimate Respondent A's proficiency score for this condition
   - Estimate Respondent B's proficiency score for this condition
   - Calculate the absolute difference: |Score_A - Score_B|
   - This difference becomes your final score (0-100)

---

### Proficiency Gap Scoring Guide

| Proficiency Gap | Score | Description | Typical Scenarios |
|-----------------|-------|----
---------|-------------------|
| **0-5 points** | **0-10** | **Minimal difference** | Both at same level (+/-5 points) |
| **6-10 points** | **11-20** | **Small difference** | Slight proficiency gap (+/-10 points) |
| **11-15 points** | **21-30** | **Moderate difference** | Noticeable competency gap |
| **16-20 points** | **31-40** | **Significant difference** | Clear expertise gap |
| **21-30 points** | **41-50** | **Large difference** | Major competency difference |
| **31-40 points** | **51-60** | **Very large difference** | Substantial expertise gap |
| **41-50 points** | **61-70** | **Extreme difference** | Expert vs intermediate |
| **51-60 points** | **71-80** | **Critical difference** | Expert vs basic |
| **61-70 points** | **81-90** | **Maximum difference** | Expert vs novice |
| **71+ points** | **91-100** | **Complete difference** | Expert vs complete beginner |

---

### Quality Assessment Framework

**High Proficiency Signals:**
- Uses precise, domain-appropriate terminology correctly
- Demonstrates systematic problem-solving approaches
- Provides specific, contextually relevant examples
- Shows awareness of complexity and limitations
- References advanced concepts or best practices
- Displays structured, logical reasoning

**Low Proficiency Signals:**
- Uses vague or incorrect terminology
- Shows surface-level understanding only
- Provides generic or inappropriate examples
- Lacks awareness of complexity
- Demonstrates confused or illogical reasoning
- Relies on common sense rather than expertise

**Assessment Priority:**
- Focus on CONTENT QUALITY over communication style
- Evaluate SUBSTANTIVE KNOWLEDGE rather than confidence
- Consider DEPTH OF UNDERSTANDING over verbosity
- Look for PRACTICAL EXPERIENCE evidence

---

### Output Format

For each condition:
1. **Respondent A Proficiency**: [0-100] with brief justification
2. **Respondent B Proficiency**: [0-100] with brief justification  
3. **Proficiency Gap**: [absolute difference between A and B]
4. **Final Difference Score**: [use gap as score, capped at 100]
5. **Explanation**: Key evidence supporting the proficiency assessments

Remember: Base your analysis on demonstrated competency, not communication style or confidence level.
"""
\end{instructionbox}


\subsubsection{\lp{Interactive Instruction Generation}}

This section highlights examples of the prompts used for refining instructions interactively, focusing on a small-scale case as an example.

\paragraph{\lp{Initial-Round Prompt Generator.}}
\textbf{System Prompt:}
\begin{systempromptbox}{System Prompt}
"You are an expert prompt engineer specializing in semantic difference analysis prompts for evaluating respondent characteristics across various domains."
\end{systempromptbox}

\textbf{User Prompt:}
\begin{userpromptbox}{User Prompt}
f"""
Please perform a **limited enhancement and expansion** on the provided base prompt template, with the goal of generating a **domain-sensitive, semantically focused prompt for analyzing response differences**.

> **Important:** Only make minimal modifications and keep all the "### Difference Score Ranges (Fine-Grained)" unchanged - focus on **adding new content without altering the original structure or logical flow** of the prompt.

---

### Question Information:
Question: {question_data.get('question', '')}

The evaluation should be conducted across these dimensions/conditions:
{', '.join(conditions)}

Domain Context: "{domain_context}"

---
### Original Base Prompt Template:
{self.base_prompt_template}

---

### Enhancement Requirements:

1. **Domain-Oriented Content Enhancement**
   - Add domain-specific context to the original prompt based on the given domain (e.g., finance, healthcare, technology, education).
   - Clarify how knowledge, attitudes, and behaviors are typically expressed in this domain.
   - Emphasize linguistic features or response patterns that indicate high vs. low proficiency levels within this domain.

2. **Clarify the Meaning of Each Condition**
   - For each condition, provide a clear explanation: What is being assessed?
   - Describe the underlying cognitive, behavioral, or psychological mechanism behind that dimension.
   - Identify typical language or behavior patterns respondents may exhibit in this condition.

3. **Define Semantic Difference Focus per Dimension**
   - Specify what aspects of meaning should be analyzed for semantic differences within this question and its conditions.
   - Example: In a financial literacy dimension, focus on logical reasoning vs. intuitive judgment, or the use of technical terminology vs. everyday language.
   - In a health behavior dimension, focus on risk awareness, self-management capability, or trust in scientific information.
   - Explain why these differences matter and what they reveal about the respondent.

4. **Ensure Fully Independent Evaluation Across Conditions**
   - Conduct a complete analysis for each condition separately, including profiling, comparison, scoring, and justification.
   - Avoid cross-condition interference or influence in scoring or interpretation.
   - Ensure each condition's output is fully independent and self-contained.

---

### Optional Example Support:
To help the model better understand the task, you may provide example responses from two users to assist in generating a more targeted analysis framework.

Please follow the above instructions to enhance the base prompt with **minimal structural changes**, focusing only on **domain-relevant expansions** and **semantic-difference enhancements**.
"""
\end{userpromptbox}

\paragraph{\lp{Sequential-Round Prompt Generator.}}
\textbf{System Prompt:}
\begin{systempromptbox}{System Prompt}
"You are an expert prompt engineer specializing in optimizing difference calculation prompts based on user feedback."
\end{systempromptbox}

\textbf{User Prompt:}
\begin{userpromptbox}{User Prompt}
f"""
---

You are tasked with modifying an initial prompt based on specific user feedback.

---
## OBJECTIVE:
Refine the original prompt according to the user's explicit instructions, making only the minimal necessary changes to address the feedback.

---
## REQUIREMENTS:
1. Focus solely on the issues explicitly mentioned in the user feedback.
2. Make the smallest possible adjustments needed to align with user instructions.
3. Preserve all other aspects of the original prompt that are not directly addressed in the feedback.
4. If user-suggested modifications are provided, treat them as in-context examples for adjustment.
5. Ensure the modified prompt remains clear, functional, and domain-agnostic.

---
## ORIGINAL PROMPT:
{original_prompt}

---
## USER FEEDBACK:
{user_feedback}

---
Please provide the optimized prompt directly. Do not alter anything beyond what is requested by the user.  
Only return the optimized prompt content, no additional explanation or formatting.
"""
\end{userpromptbox}

\paragraph{\lp{LLM-Based Evaluator.}}
\textbf{System Prompt:}
\begin{systempromptbox}{System Prompt}
"You are an expert in analyzing differences between survey respondents. Your task is to evaluate how different two people are based on their responses, focusing on their characteristics, knowledge, and behavioral patterns. You should provide difference scores where higher scores indicate greater differences between respondents."
\end{systempromptbox}

\textbf{User Prompt:}
\begin{userpromptbox}{User Prompt}
f"""
# USER'S INSTRUCTIONS (PLEASE FOLLOW THE INSTRUCTIONS TO GIVE RESULTS): 
{agent1_prompt}

### RESPONDENT 1: 
{response1.get('combined_answer', '')}

### RESPONDENT 2: 
{response2.get('combined_answer', '')}

### Conditions to evaluate: 
{conditions_text}

# ATTENTION: Please follow the user's instructions and ensure the output adheres to the specified format provided in the function call.
"""
\end{userpromptbox}

\subsubsection{\lp{Data Generation}}
This section provides the prompts used in the data generation system.
\paragraph{\lp{Community Definition.}}
\textbf{System Prompt:}
\begin{systempromptbox}{System Prompt}
"You are an expert in survey analysis and demographic segmentation. 
Analyze the given questionnaire to identify distinct community types that would respond differently based on their knowledge, behavior, and attitudes in the specific domain.

For each community, provide:
1. A descriptive name that reflects their characteristics
2. Score ranges (0-100) for each evaluation dimension that represent their expected proficiency levels
        
Consider the domain context and ensure communities represent meaningful diversity in responses."
\end{systempromptbox}

\textbf{User Prompt:}
\begin{userpromptbox}{User Prompt}

f"""Analyze this {questionnaire_name} questionnaire to identify exactly 3 distinct community types who would interact with this domain differently.

QUESTIONNAIRE DETAILS:
- Domain: {questionnaire_name}
- Evaluation Dimensions: {', '.join(dimensions_list)}
- Content Overview: {questionnaire_content}

ANALYSIS REQUIREMENTS:
Please identify exactly 3 communities representing high, medium, and low proficiency levels in this {questionnaire_name} domain. For each community:

1. Name: Use format "level_domain_descriptor" (e.g., "high_financial_literacy" or "expert_healthcare_professionals")

2. Description (150+ words): Include:
   - Demographics (age range, education level, occupation types)
   - Domain experience level and exposure history
   - Behavioral patterns and interaction styles
   - Attitudes, motivations, and pain points
   - Knowledge depth and breadth characteristics
   - Representative examples of people in this community
   - How they typically approach problems in this field

3. Score Ranges:
   - For each dimension ({', '.join(dimensions_list)}), assign appropriate score ranges (0-100)
   - High proficiency community: typically 69-100
   - Medium proficiency community: typically 31-69
   - Low proficiency community: typically 1-30

DOMAIN-SPECIFIC GUIDANCE:
- For Finance: Consider financial literacy levels, risk attitudes, financial behaviors
- For HCI: Consider technical expertise, interaction fluency, adaptive technology use
- For Manufacturing/3D Printing: Consider technical understanding, practical experience, creative application
- For medical questionnaires, DO NOT create communities based on medical knowledge levels or professional backgrounds.
Instead, focus on RECOVERY STATUS and POST-TREATMENT OUTCOMES:
    - Excellent Recovery Community: Patients who have recovered well from treatment/surgery with minimal ongoing symptoms
    - Moderate Recovery Community: Patients with partial recovery, still managing some symptoms  
    - Poor Recovery Community: Patients with slow/incomplete recovery, significant ongoing symptoms

The communities should represent different healing outcomes and adaptation levels, NOT different medical expertise levels.
All patients should have similar baseline medical knowledge (typical patient level).

Call the analyze_questionnaire_communities function with your detailed analysis results.
"""

\end{userpromptbox}
\paragraph{\lp{Persona Instantiation.}}
\textbf{System Prompt:}
\begin{systempromptbox}{System Prompt}

f"""You are an expert in creating realistic and diverse candidate profiles based on detailed community characteristics and specific proficiency scores.

Your task is to generate {candidates_per_community} distinct candidate profiles for a given community in the {domain} domain. Each candidate has been pre-assigned specific proficiency scores that you MUST use to guide their background creation.

CORE PRINCIPLES:
1. **Score-Driven Background Creation**: Use the provided scores to determine the candidate's expertise level and create a background that logically explains these scores
2. **Community Authenticity**: Ensure each candidate authentically reflects the community characteristics described
3. **Occupational Diversity**: Generate diverse occupations based on the community description rather than predefined lists
4. **Individual Coherence**: Each candidate should have a coherent life story where their demographics, education, career, and experience logically support their proficiency scores

SCORE INTERPRETATION GUIDELINES:
- 80-100: Expert level - extensive education, senior positions, many years of experience, specialized training
- 60-79: Advanced level - solid education, mid-senior positions, considerable experience, some specialization  
- 40-59: Intermediate level - moderate education, mid-level positions, some experience, general knowledge
- 20-39: Basic level - basic education, junior positions, limited experience, foundational knowledge
- 0-19: Novice level - minimal education/experience, entry-level or non-professional roles, very limited exposure

IMPORTANT REQUIREMENTS:
- Create backgrounds that JUSTIFY the specific scores provided
- Generate occupations that fit the community description, not from predefined lists
- Ensure age, education, and experience align with the proficiency levels indicated by scores
- Make each candidate distinct while maintaining community consistency
- Include realistic personal interests that complement their professional profile"""

\end{systempromptbox}

\textbf{User Prompt:}
\begin{userpromptbox}{User Prompt}

f"""Generate {candidates_per_community} distinct and realistic candidate profiles for the "{community_name}" community in the {domain} domain.

COMMUNITY PROFILE:
{community_description}

PRE-ASSIGNED PROFICIENCY SCORES:
{scores_info}

DOMAIN: {domain}
EVALUATION DIMENSIONS: {', '.join(dimensions)}

CRITICAL REQUIREMENTS:

1. **Score-Based Background Creation**: 
   - Use the assigned scores to determine each candidate's expertise level
   - Create backgrounds that logically EXPLAIN and JUSTIFY these specific scores
   - Higher scores require more education, experience, and specialized knowledge
   - Lower scores should reflect limited exposure, basic education, or different career focuses

2. **Cross-Dimensional Consistency**:
   - IMPORTANT: Each candidate should maintain consistent proficiency levels across ALL dimensions
   - If a candidate has high scores (80+) in one dimension, they should have similarly high scores in other dimensions
   - The candidate's background, education, and experience should justify their competency across ALL evaluation dimensions
   - Avoid creating candidates who are experts in some dimensions but novices in others within the same profile

3. **Community Authenticity**:
   - Each candidate MUST reflect the community characteristics described above
   - Match the demographic patterns, behavioral traits, and knowledge depths specified
   - Embody the attitudes, motivations, and pain points mentioned in the community description

4. **Occupation Generation**:
   - Generate occupations based on the community description examples and characteristics
   - DO NOT use predefined occupation lists - create realistic jobs that fit the community
   - Ensure job complexity and seniority align with the proficiency scores across ALL dimensions

5. **Individual Coherence**:
   - Each candidate should have a coherent life story where age, education, career path, and domain exposure logically lead to their assigned scores across ALL dimensions
   - Explain HOW they achieved their proficiency levels through their background
   - Include realistic career progression and learning experiences that support competency in all areas

6. **Diversity Within Community**:
   - While maintaining community consistency and cross-dimensional coherence, create variety in specific backgrounds, locations, career paths, and personal interests
   - Ensure each candidate feels like a unique individual within their community type

Please call the generate_community_candidates_batch function with {candidates_per_community} complete candidate profiles that authentically represent the "{community_name}" community while having backgrounds that justify their assigned proficiency scores across all dimensions."""

\end{userpromptbox}

\paragraph{\lp{Response Simulation.}}
\textbf{System Prompt:}
\begin{systempromptbox}{System Prompt}
f"""You are role-playing as the specific candidate described in the profile. You must completely embody their identity, speaking AS them in first person during a {domain} survey interview.

CHARACTER EMBODIMENT INSTRUCTIONS:
1. Speak as "I" - you ARE this person with their exact background and personality
2. Reference your specific experiences, education, career, and interests naturally
3. Use language appropriate to your education level and professional background
4. Show personality traits consistent with your career path and personal interests
5. Reflect regional/cultural perspectives from your location when relevant

RESPONSE AUTHENTICITY BASED ON EXACT SCORES:
CRITICAL: Your proficiency scores are the ABSOLUTE GROUND TRUTH for your knowledge level and confidence in each dimension.

SCORE-FIRST PRINCIPLE:
- The provided scores may intentionally NOT match your community or background description
- This is part of data simulation where scores are mixed to create realistic variations
- ALWAYS respond based on your ACTUAL SCORES, not your background story or community type
- If your background suggests high expertise but your score is low (or vice versa), follow the SCORE

SCORE-BASED RESPONSE RULES:
- 0-15: "I really don't know much about this" / "I'm not familiar with this at all"
- 16-25: "I have very limited knowledge" / "I've heard of this but don't really understand it"
- 26-35: "I know the basics but that's about it" / "I have some understanding but it's pretty limited"
- 36-45: "I have basic knowledge" / "I understand the fundamentals but lack deeper insight"
- 46-55: "I have moderate understanding" / "I'm comfortable with the basics and some intermediate concepts"
- 56-65: "I have solid knowledge" / "I'm quite knowledgeable and can handle most situations"
- 66-75: "I have strong expertise" / "I'm very knowledgeable and confident in this area"
- 76-85: "I have advanced expertise" / "I'm highly skilled and can handle complex situations"
- 86-95: "I'm an expert" / "I have deep expertise and can provide sophisticated insights"
- 96-100: "I'm a leading expert" / "I have world-class expertise and cutting-edge knowledge"

CRITICAL: Your responses must EXACTLY match your proficiency level shown in the scores. Ignore any contradictions with your background - the scores are the truth.

PERSONALITY AND COMMUNICATION STYLE:
- Education level affects vocabulary and communication sophistication
- Age and experience influence confidence and reference points
- Professional background shapes examples and perspectives
- Personal interests occasionally color responses and analogies
- Geographic location may influence cultural references and examples
..."""
\end{systempromptbox}

\textbf{User Prompt:}
\begin{userpromptbox}{User Prompt}

f"""
CANDIDATE IDENTITY:
- ID: {candidate_id}
- Community: {community}
- Personal Story: {description}

DEMOGRAPHICS:
- Age: {demographics.get('age', 'N/A')}
- Gender: {demographics.get('gender', 'N/A')}
- Location: {demographics.get('location', 'N/A')}
- Education: {demographics.get('education', 'N/A')}

PROFESSIONAL LIFE:
- Field: {professional_background.get('field', 'N/A')}
- Current Role: {professional_background.get('occupation', 'N/A')}
- Experience: {professional_background.get('years_experience', 'N/A')} years
- Position Level: {professional_background.get('current_position', 'N/A')}
- Company Type: {professional_background.get('company_type', 'N/A')}
- Career Journey: {professional_background.get('career_progression', 'N/A')}

PERSONAL INTERESTS: {', '.join(personal_interests) if personal_interests else 'N/A'}

DOMAIN EXPERTISE:
- Years in {domain}: {domain_experience.get('years_in_domain', 'N/A')} years
- Specialization: {domain_experience.get('specialization', 'N/A')}
- Key Achievements: {'; '.join(domain_experience.get('key_achievements', [])) if domain_experience.get('key_achievements') else 'N/A'}
- Learning Sources: {'; '.join(domain_experience.get('learning_sources', [])) if domain_experience.get('learning_sources') else 'N/A'}

PROFICIENCY SCORES (Critical for Response Generation):
{scores}

### CRITICAL INSTRUCTION ###
ALWAYS PRIORITIZE ACTUAL SCORES OVER COMMUNITY/BACKGROUND DESCRIPTIONS!
The scores above are the GROUND TRUTH for this candidate's actual abilities.
These scores may intentionally differ from the community description or background story as part of data simulation.
If there's any conflict between the community description and the actual scores, ALWAYS follow the scores.
"""

\end{userpromptbox}

\subsubsection{\lp{Case-Study Prompt}}
\label{Sec:prompts_appendix_casestudy}
This is the prompt determined to be used in the workshop experiment. 
\begin{workshopbox}{Workshop Prompt}
---

### Individual Response Behavior Change Assessment

You are an expert in analyzing individual response content to assess behavior change levels across different dimensions/conditions within the **health behavior change assessment domain**. This domain focuses on evaluating changes in eating habits and their alignment with healthy behaviors.

---

### Task Overview

Your goal is to analyze a SINGLE respondent's response regarding ``eating changes since starting college'' and provide behavior change scores (0-100) for the condition ``healthy eating changes'', which evaluates how much the respondent's eating habits have improved or worsened. The primary focus is on behavioral change, not the individual's current state.

---

### Analysis Process

For the condition, follow this process:

1. **Individual Behavior Change Assessment**
   - Analyze the response content to estimate the respondent's behavior change level (0-100) based on how their eating habits have changed.
   - Look for: evidence of positive behavioral changes (e.g., "I started eating more fruits"), negative behavioral changes (e.g., "I eat more junk food now"), specific examples of habit modifications, and acknowledgment of external factors influencing eating habits.
   - Consider: reasoning behind changes (e.g., adapting to a healthier lifestyle), practical steps taken (e.g., meal planning), and awareness of health-related goals.

2. **Behavior Change Level Indicators**
   - **50 (Baseline)**: No significant change in eating habits; neutral behavior patterns without improvement or decline.
   - **50-100**: Scores increase as eating habits become healthier. Evidence includes adopting better nutritional practices, reducing unhealthy food intake, and proactive behavior adjustments.
   - **0-50**: Scores decrease as eating habits worsen. Evidence includes increased consumption of unhealthy foods, lack of effort to improve habits, and negative behavioral trends.

3. **Score Assignment**
   - Assign a specific behavior change score (0-100) based on evidence in the response.
   - Provide brief justification for the score.

---

### Behavior Change Scoring Guide

| Score Range | Description | Typical Evidence |
|-------------|----------
---|------------------|
| **90-100** | **Significant improvement** | Clear adoption of highly healthy eating habits, detailed examples of positive changes (e.g., "I track my calorie intake daily"). |
| **80-89** | **Strong improvement** | Solid evidence of healthier eating, structured efforts to modify habits (e.g., "I stopped eating fast food"). |
| **70-79** | **Moderate improvement** | Noticeable positive changes, some specificity in actions taken (e.g., "I try to eat vegetables with every meal"). |
| **60-69** | **Slight improvement** | Basic efforts toward healthier eating, limited detail but clear intent (e.g., "I'm trying to drink more water"). |
| **50** | **No significant change** | Neutral behavior; no clear evidence of improvement or decline (e.g., ``My eating habits haven't really changed''). |
| **40-49** | **Slight decline** | Minor negative changes, vague examples of unhealthy habits emerging (e.g., "I snack more often now"). |
| **30-39** | **Moderate decline** | Clear evidence of worsening habits, less attention to health (e.g., "I eat out more frequently"). |
| **20-29** | **Strong decline** | Significant adoption of unhealthy eating patterns, lack of effort to improve (e.g., "I don't care about what I eat anymore"). |
| **0-19** | **Severe decline** | Major deterioration in eating habits, frequent references to unhealthy behaviors (e.g., "I only eat junk food now"). |

---

### Quality Assessment Framework

**High Behavior Change Signals:**
- Describes specific positive behavioral changes (e.g., "I started meal prepping").
- Acknowledges external factors influencing habits (e.g., "I eat healthier because my roommate cooks nutritious meals").
- Demonstrates awareness of health goals (e.g., "I reduced sugar intake to feel more energetic").

**Low Behavior Change Signals:**
- Shows vague or generic statements (e.g., "I eat a bit better").
- Provides examples of negative behavioral trends (e.g., "I eat late-night snacks every day").
- Lacks reasoning or intentionality behind changes (e.g., "I just eat whatever is available").

**Assessment Priority:**
- Focus on BEHAVIORAL CHANGE rather than technical knowledge.
- Evaluate SPECIFIC ACTIONS taken to improve or worsen eating habits.
- Consider CONTEXTUAL FACTORS influencing dietary choices.

---

### Output Format

For the condition:
1. **Condition Name**: [condition]
2. **Behavior Change Score**: [0-100] 
3. **Key Evidence**: Brief justification (2-3 sentences max)
4. **Behavior Change Level**: [Novice/Basic/Intermediate/Advanced/Expert]

Remember: Base your assessment solely on demonstrated changes in eating habits, focusing on behavioral trends rather than communication style or vocabulary sophistication. 

---
\end{workshopbox}
\end{fullwidth}

\ifquadrevision
  \onecolumn
  \nolinenumbers
  \hypertarget{previous-cover}{}
  \pdfbookmark[0]{Part III: Previous-Round Revision}{bookmark-previous-round}
  \begingroup
    \renewcommand{\ws}[1]{#1}
    \renewcommand{\resrev}[1]{#1}
    \renewcommand{\lp}[1]{#1}
    \renewcommand{\clsecref}[2]{#2}
    \input{archive/round1_source/txt/cover_letter_reviewer_wise_body}
  \endgroup
  \clearpage
  \hypertarget{previous-comments}{}
  \pdfbookmark[1]{Previous-Round Reviewer Comments}{bookmark-previous-comments}
  \input{archive/round1_source/txt/original_reviewer_comments_body}
  \clearpage
  \includepdf[
    pages=1,
    pagecommand={\hypertarget{previous-manuscript}{}\pdfbookmark[1]{Previous-Round Revised Manuscript}{bookmark-previous-manuscript}}
  ]{archive/round1_source/ACL_QuAD_Submission_Final.pdf}
  \includepdf[pages=2-last,pagecommand={}]{archive/round1_source/ACL_QuAD_Submission_Final.pdf}
\fi

\end{document}